\documentclass{article}

\usepackage{microtype}
\usepackage{graphicx}
\usepackage{booktabs} 

\usepackage{hyperref}
\usepackage{chngcntr}

\usepackage[accepted]{icml2019}

\def\psfancypar#1#2{\begingroup\def\par{\endgraf\endgroup\lineskiplimit=0pt}
               \setbox2=\hbox{\large\sc #2}
               \newdimen\tmpht \tmpht \ht2 \advance\tmpht by \baselineskip
               \font\hhuge=Times-Bold at \tmpht
               \setbox1=\hbox{{\hhuge #1}}
               \count7=\tmpht \count8=\ht1
               \divide\count8 by 1000 \divide\count7 by \count8 
               \tmpht=.001\tmpht\multiply\tmpht by \count7 
               \font\hhuge=Times-Bold at \tmpht
               \setbox1=\hbox{{\hhuge #1}}
               \noindent
                \hangindent1.05\wd1
               \hangafter=-2 {\hskip-\hangindent
               \lower1\ht1\hbox{\raise1.0\ht2\copy1}%
                \kern-0\wd1}\copy2\lineskiplimit=-1000pt}

\newcommand{\E}{\mbox{{\rm E}}}

\def\boxit#1{\vbox{\hrule\hbox{\vrule\kern3pt
        \vbox{\kern3pt#1\kern3pt}\kern3pt\vrule}\hrule}}

\def\reals{ { {\rm  I \kern-0.15em R }  } }
\def\complex{ {\,{{\rm C} \kern-0.50em \raise0.20ex {  |}}\, }}

\def\Rbf{{\bf R}}

\def\be{\vskip .3cm \begin{equation}}
\def\ee{\end{equation} \vskip .4cm \noindent}
\newcommand{\R}{\mbox{$\hat {\bf R}_{N}$}}

\def\Rxx{\Rbf_{\ssstyle X\kern-.1em X}}

\let\ssstyle=\scriptscriptstyle

\def\Kout{\setbox1=\hbox{\Huge\bf K}\hbox to
1.05\wd1{\hspace{.05\wd1}
}}
\def\Sout{\setbox1=\hbox{\Huge\bf S}\hbox to 1.05\wd1{\hspace{.05\wd1}
}}

  \ifx\LabelFigloaded\MYundefined\relax
  \else
   \fi

  \def\LabelFigloaded{\relax}

  \chardef\LabelFigCatAt\the\catcode`\@
  \catcode`\@=11

 \let\LabelFigwlog@ld\wlog
 \def\wlog#1{\relax}

 \ifx\\\MYundefined@
    \let\\\relax
 \fi

  \def\ms@g{\immediate\write16}

 \def\N@wif{\csname newif\endcsname }
 \def\Temp@ {\N@wif\ifIN@}
 \ifx\INN@\MYundefined@
    \else \let\Temp@\relax
 \fi
 \Temp@

  \def\IN@{\expandafter\INN@\expandafter}
  \long\def\INN@0#1@#2@{\long\def\NI@##1#1##2##3\ENDNI@
    {\ifx\m@rker##2\IN@false\else\IN@true\fi}%
     \expandafter\NI@#2@@#1\m@rker\ENDNI@}
  \def\m@rker{\m@@rker}
 
  \newtoks\Initialtoks@  \newtoks\Terminaltoks@
  \def\SPLIT@{\expandafter\SPLITT@\expandafter}
  \def\SPLITT@0#1@#2@{\def\TTILPS@##1#1##2@{%
     \Initialtoks@{##1}\Terminaltoks@{##2}}\expandafter\TTILPS@#2@}

 \def\Shifted@@#1#2#3{\setbox0=\hbox{#3}%
   \raise -\dp0\vbox {\kern-#2%
       \hbox {\kern#1\unhbox0\kern-#1}%
           \kern#2}}

 \newcount\gridcount
 \newbox\auxGridbox@ \newbox\hGridbox@ \newbox\vGridbox@
 \newbox\Labelbox@ \newbox\auxLabelbox@
 \newbox\Coordinatebox@
 \newtoks\Labeltoks@
 \newdimen\Wdd@ \newdimen\Htt@
 \newdimen\Wddd@ \newdimen\Httt@
 
 \def\Wr@{\immediate\write16}

 \newdimen\GL@wd
 \def\GridLineWidth#1{\GL@wd=#1}

 \def\gobble#1{}
 \def\EdgeErr@{\Wr@{}%
      \Wr@{\string\Edges\space argument
      1, 10, 100 or 1000 please\string!}%
      }

 \newcount\Edgect@

 \def\Sweepup#1\endSweepup{}

 \def\SetEdges@{%
    \edef\Zr@@s{\expandafter\gobble\number\Edgect@\empty}%
        \count255=0\Zr@@s\relax
        \ifnum\count255=\z@\else\EdgeErr@\show\tailtest\fi
        \count255=1\Zr@@s\relax
        \ifnum\count255=\Edgect@\relax\else\EdgeErr@\show\leadtest\fi
    \EdgGl@b\edef\Zr@s{\expandafter\gobble\Zr@@s\empty}
    \ifnum\Edgect@>\@ne\relax\EdgGl@b\let\L@Dc\empty
        \else\EdgGl@b\edef\L@Dc{\string.}\fi
    \ifnum\Edgect@>\@ne\relax
        \EdgGl@b\edef\Edgescale@##1{\divide##1 by \Edgect@}%
        \else\EdgGl@b\edef\Edgescale@##1{}\fi
    }

 \def\Edges#1{\Edgect@=#1\relax
     \let\EdgGl@b\global \SetEdges@}

 \Edges{1}

 \def\hhrule{\hrule height \GL@wd\vskip-.\GL@wd}

 \def\hRule@{%
   \advance\gridcount -2%
   \vfil\hhrule\vfil
   \llap{\smash{\raise -2.5pt
     \hbox{\L@Dc\number\gridcount\Zr@s\kern2pt}}}%
   \hhrule
   }

\def\vvrule{\vrule width \GL@wd \kern-\GL@wd}

 \def\vRule@{\advance\gridcount 2%
   \hfil\vvrule\hfil
   \setbox\auxGridbox@=\vbox to 0pt
      {\vskip \Htt@\vskip 2pt
        \hbox to 0pt{\hss\L@Dc\number\gridcount\Zr@s\hss}\vss}%
      \wd\auxGridbox@=0pt \box\auxGridbox@
   \vvrule
   }

 \def\PlaceGrid@@{\gridcount=10 
  \setbox\hGridbox@=\hbox{%
        \hbox{%
             \hskip-.4pt\vrule
             \vbox to \Htt@{%
               \offinterlineskip\parindent=\z@\relax
               \hbox to \Wdd@{\hfil}
               \hRule@\hRule@\hRule@\hRule@
               \vfil\hhrule\vfil}%
             \vrule\hskip-.4pt}
    }%
  \gridcount=0%
  \setbox\vGridbox@=\hbox{%
      \vbox{\offinterlineskip\parindent=0pt\hsize=0pt
         \vskip-.4pt\hrule%
         \hbox to \Wdd@{%
                 \vtop to \Htt@{\vfil}%
                 \vRule@\vRule@\vRule@\vRule@
                 \hfil\vvrule\hfil}%
         \hrule\vskip-.4pt}}%
  \wd\hGridbox@=0pt\ht\hGridbox@=0pt
  \wd\vGridbox@=0pt\ht\vGridbox@=0pt
  \hbox{\box\hGridbox@\box\vGridbox@}%
  }

 \def\LabelsGlobal{\def\LabGl@b{\global}}
 \def\LabelsLocal{\def\LabGl@b{}}
 \LabelsGlobal 

 \def\SetLabels#1\endSetLabels{%
   \LabGl@b\Labeltoks@={#1()\\}%
   }

 \LabGl@b\Labeltoks@={()\\}

 \def\ShowGrid{\LabGl@b\let\PlaceGrid@\PlaceGrid@@}
 \def\HideGrid{\LabGl@b\let\PlaceGrid@\relax}
 \def\Grids{\ShowGrid\LabGl@b\let\GridSwitch@\ShowGrid}
 \def\noGrids{\HideGrid\LabGl@b\let\GridSwitch@\HideGrid}

 \noGrids

 \def\bAdjust@@{%
     \setbox\auxLabelbox@=\hbox{\raise \dp\auxLabelbox@
            \box\auxLabelbox@}}
 \def\bAdjust@{\let\vAdjust@\bAdjust@@}

 \def\eAdjust@@{\dimen0=-.5\ht\auxLabelbox@
     \advance\dimen0 by .5\dp\auxLabelbox@
     \setbox\auxLabelbox@=
            \hbox{\raise\dimen0\box\auxLabelbox@}}
 \def\eAdjust@{\let\vAdjust@\eAdjust@@}

 \def\tAdjust@@{%
     \setbox\auxLabelbox@=\hbox{\raise-\ht\auxLabelbox@
            \box\auxLabelbox@}}
 \def\tAdjust@{\let\vAdjust@\tAdjust@@}

 \let\vAdjust@\relax

 \def\lAdjust@{\let\hAdjust@\rlap}
 \def\rAdjust@{\let\hAdjust@\llap}

 \let\hAdjust@\relax\let\vAdjust@\relax

 \def\FetchLabel@#1(#2)#3\\{%
     \IN@0#2@@\ifIN@
        \setbox0=\hbox{\ignorespaces#1#3\unskip}%
        \ifdim\wd0>0pt
           \ms@g{}%
           \ms@g{ !!! Bad label(s)? !!!}%
           \message{ #1(#2)#3}%
        \fi
        \def\LabelMole@##1\endFetchLabel@{%
            \IN@0()\\@##1@%
            \ifIN@\def\Temp@{\FetchLabel@##1\endFetchLabel@}%
            \else\def\Temp@{}%
            \fi
            \Temp@
           }%
     \else
       \ignorespaces#1\unskip
       \setbox\auxLabelbox@=%
         \hbox to 0pt{\hss\ignorespaces\hAdjust@
          {\ignorespaces#3\unskip}\hss}%
       \vAdjust@
       \let\hAdjust@\relax\let\vAdjust@\relax
       \AugmentLabelBox@@{#2}%
       \ht\Labelbox@=0pt\dp\Labelbox@=0pt
       \let\LabelMole@\FetchLabel@%
     \fi\LabelMole@}

 \newtoks\XYSep@ 
 \def\SetXYSeparator#1{%
     \IN@0#1@@\ifIN@\XYSep@{*}%
     \else
     \XYSep@{#1}%
     \fi
     }

 \SetXYSeparator*

 \def\AugmentLabelBox@@#1{%
     \IN@0\the\XYSep@ @#1@\ifIN@
       \SPLIT@0\the\XYSep@ @#1@%
       \setbox\Labelbox@=\hbox to 0pt{%
         \unhbox\Labelbox@
         \Shifted@@{\the\Initialtoks@\Wddd@}%
         {\the\Terminaltoks@\Httt@}%
         {\box\auxLabelbox@}}%
     \else
         \ms@g{}%
         \ms@g{ !!! Bad insertion point. !!!}%
         \message{ (#1\ this point was rejected.)}%
     \fi
    }

 \def\FetchOption@#1[#2]#3\endFetchOption@{%
    \def\temp{#1}
    \ifx\temp\empty
       \Edgect@=#2\relax
       \let\EdgGl@b\relax
       \SetEdges@
       \Cleaner@#3%
    \fi}

 \def\Cleaner@#1[@]{\Labeltoks@{#1}}
     
 \def\PlaceLabels@@{\mathsurround=0pt
     \def\Cr@{\\}%
     \let\L\lAdjust@\let\R\rAdjust@
     \let\B\bAdjust@\let\E\eAdjust@\let\T\tAdjust@
     \expandafter\FetchOption@\the\Labeltoks@[@]\endFetchOption@
     \Wddd@=\Wdd@ \Edgescale@\Wddd@ 
     \Httt@=\Htt@ \Edgescale@\Httt@
     \expandafter\FetchLabel@\the\Labeltoks@\endFetchLabel@
     \box\Labelbox@
     }%

 \let \PlaceLabels@\PlaceLabels@@

 \def\AffixLabels#1{\setbox\Coordinatebox@=\hbox{#1}%
      \Wdd@=\wd\Coordinatebox@ \Htt@=\ht\Coordinatebox@
      \advance\Htt@ \dp\Coordinatebox@
      \hbox{\copy\Coordinatebox@\kern-\Wdd@ 
           \Shifted@@{0pt}{-\dp\Coordinatebox@}%
           {\PlaceLabels@\PlaceGrid@}%
           \kern\Wdd@}%
      \GridSwitch@ 
      \LabGl@b\Labeltoks@{()\\}%
      }
 
   \let\wlog\LabelFigwlog@ld   
   \catcode`\@=\LabelFigCatAt  

                                By

              Raymond S\'eroul <A18645@FRCCSC21.BITNET>
                                and 
              Laurent Siebenmann <lcs@topo.math.u-psud.fr>
    
              VERSIONS: July 1991, Oct 1991, Jan 1992, July 1992

INTRODUCTION

      This labelling package is intended for TeX users who
rely on non-TeX sources for for their graphics inserts.  It
provides means for adding TeX labels to such inserts with a
minimum of fuss. 

       For most labels, TeX users have in the past found it
reasonably convenient to rely on non-TeX sources. Typical
occasions when an inescapable need for TeX labels seemed to
arise are

 (a) when the graphics program lacks certain exotic or complex
mathematical symbols

 (b) when the very highest typographical quality is wanted for the
labels

 (c) when labels included with the graphics fail to print, 
 and you cannot figure out why (cf. boxedeps.doc).  The labels
 provided by labelfig.tex are 100

       Since this package first appeared, many users, who in the
past scarcely dreamed of using TeX labels, have come to use
nothing but.  So it is now appropriate to add

Intoxication Warning:  TeX labels may be addictive and expensive. 

     If you have a fast preview you may disagree, and even find
that this package provides an agreeable paste-up environment; see
extra applications at end.

     Note to publishers: It is possible and convenient to ultimately
export the TeX labels produced by labelfig.tex to become an integral
part of the EPS file. This is often desired by a publisher who typically
uses an "upmarket" graphics or page layout program, with which the
staff is skilled in perfecting figures.  See Appendix I for
a recipe.

     The authors are grateful to Patrick Ion of Math Reviews for
helpful comments and encouragement.

BASIC INSTRUCTIONS

    After reading in the macro file using

preview or proof your figure with a coordinate grid printed on
top, by typing the following:

    \ShowGrid  
    \AffixLabels{<the graphics insertion>}

Here <the graphics insertion> is what you would type to insert
the graphics object alone without the grid.  This must provide
for the space around it. For example <the graphics insertion>
might well be \BoxedEPSF{MyFigure scaled 700} using the
boxedeps.tex macro package (from same source); this provides a
TeX box containing the encapsulated PostScript insert specified by
the file MyFigure. \AffixLabels{...} provides the grid (supposing
\ShowGrid is present) and later, once you have specified labels
using the grid, it will "tack on" the labels.

     The grid is a sort of (usually elongated) checkerboard of
ten rows and ten columns and its (internal) partitions are by
default numbered  .1, ... ,.9  both horizontally (X-coordinate
running left to right) and vertically (Y-coordinate running bottom
to top).  Thus the points enclosed by the grid correspond to the
points of the unit square in the cartesian "X-Y" plane, the lower
left corner corresponding to the origin (0,0).  By extrapolation,
the full page corresponds to a larger rectangle in the plane.

     These coordinates serve to position labels as follows.
Before the \AffixLabels{...} command type label specifications:

  \SetLabels
   (<X-coordinate>*<Y-coordinate>) <first label> \\
   .
   .
   .
   (<X-coordinate>*<Y-coordinate>)  <last label> \\
  \endSetLabels

Each row specifies one label and is terminated by \\.  In each
row, the position indicator comes first; it is written as a
standard cartesian point except that the X- and Y- coordinates
are separated by * rather than a comma because TeX allows a
comma as decimal point. There are no dimension units to specify
as the unit is the grid itself.

     By default, this cartesian point specifies where the middle
of the baseline of the label will be located.  However if you precede
the point by \L [or \R] the left [or right] edge of the baseline will
be located there. Similarly you may also precede the point by \T, \E,
or \B to vertically align the top equator or bottom of the label box
at the specified point.  This gives nine standard positions of
the label with respect to the insertion point --- corresponding to
the eight principle points of the compas and the center

                     \L\T     \T      \R\T

                     \L\E     \E      \R\E

                     \L\B     \B      \R\B

But this neglects the default "baseline" level of TeX,
giving potentially three more positions

                     \L    <no tag>   \R

For text, the baseline level is often the preferred. Its relation to
the others is variable. It will often coincide with the bottom level,
as happens for "X".  But it is often distinct, as for "g", in which
case you have in all 12 distinct positions rather than 9.

     It is convenient to think of this specification of label
position as attaching the label by a thumb-tack to the coordinate
grid. There are up to twelve positions of the thumb-tack on the
label, while the position of the thumb-tack on the coordinate grid is
arbitrary.  Normally, one choses the position of the thumb-tack on
the label to be the one that is the closest to the item being
labeled.  There are good reasons for this "rule of thumb":

   (a)  It facilitates correct positioning at first try.

   (b)  If the scale of the figure must be altered after labels
have been affixed, the labels have a good chance of remaining well
positioned.

   (c)  The visible grid need not extend beyond the "bounding box"
for the figure, because the best preferred position is always
(at least almost) within the bounding box .

The second reason is particularly important. Indeed it often
happens that scale has to be altered after labelling begins, in
order to either provide space for the labels, or to adjust
proportions between the labels and the figure.  (The size of labels
is unaffected by scaling.)

     Here is an artificial but self-contained test which uses
TeX rules to make a graphics object.

TEST

    Do not skip this!

 \def\FrameIt#1{\hbox{\vrule$\vcenter {\hrule\kern3pt%
             \hbox {\kern3pt #1\kern3pt}%
               \kern3pt\hrule}$\relax\vrule}}

 \def\Caption#1#2{\FrameIt{%
       \vtop {\hsize=#1\relax \parindent=0pt
         \leftskip=0pt \rightskip=0pt plus15pt
         \parfillskip=0pt
         \lineskip=1pt\baselineskip=0pt
         #2}}}

 \def\FirstQuadrant{\hbox to 100pt{\vrule\vbox to 100pt{%
        \hbox to 100pt{\hfil}\vfil\hrule}\hss}}

  \SetLabels
    \R(.5*.2) $\zeta\,\cdot$\\
    (.9*-.10) $\xi$\\
    \R(-.03*.9) $\eta$\\
    \T(.5*.9) \Caption{70pt}{%
          \it The norm of
          $g(\xi+i\eta)$ is indicated on
          contours of this invisible surface.}\\
  \endSetLabels

  \AffixLabels{\FirstQuadrant}

  \end

  Note that the coordinates to use for labels are indicated on the
edges of the grid (when visible) corresponding to the conventional
x- and y- axes of the Cartesian plane. By default the grid is
1-by-1. However, by the command \Edges{100}, you can change this
to 100-by-100 and many users find this alternative most
convenient. Place the command \Edges{...} in your style file (or
header) since its effect is is global. Other possible edge values
are 10 and 1000.

  If you use the command \Edges{...} at all, do so with care.  For
if you accidentally delete an \Edges{...} command your labels will
abruptly be badly misplaced and may logically but mysteriously
generate "dimension too big" errors under TeX and "off page" errors
under your driver.  

  You can dictate the edgescale for an individual figure by giving
the scale in brackets immediately after \SetLabels.  Thus, to
import into an article using say \Edge{100} a figure labelled using
another edgescale, say the original 1-by-1 default, you can use
\SetLabels[1]...\endSetLabels.

GETTING IT DOWN PAT

     Complicated labeling deserves the same respect as
complicated mathematics.  Do not expect it to come out perfect the
first time!  What is needed in either case is a mechanism to
repeatedly typeset troublesome pieces.

     One mechanism is always available.  One does complicated
labelling in a separate "test" file involving just the figure being
labelled;  a texpert will know how to \dump TeX's current state as
a temporary format that restarts rapidly at each retry.  Usually,
one then pastes the completed labelled figure back into the main
TeX file, but, of course, one can also \input it as an auxiliary
file.

     If you do not have a TeXpert at handy, here is a first
approximation to an efficient setup. By deletions reduce a copy
of your article to just a few lines before and after the figure.
Now label the figure, and finally, copy and paste the labelled
figure to the original article. Then copy the next figure to label
into this testbed and repeat. The TeXpert can improve the  speed
at which TeX starts up, by compiling a format specifically for
your article; just one caution: best NOT include in the format
ephemeral details of setup like \Set<mydriver>ArtSpecials (from
boxedeps.tex because this reads  figure dimensions which you may
change during your work session.

     An improved mechanism to repeatedly typeset troublesome
pieces is now available on the Macintosh; it is called LinoTeX;
see the same ftp sources.  It could be set up on many types
of computer.

     Before using labelfig.tex to attach labels to a graphics
object inserted using boxedeps.tex or BoxedArt.tex, make it a
firm rule to carefully adjust the bounding box using the trimming
commands of these packages, and also at least tentatively scale
and position the object. Beware of changing the grid inadvertently
after the labels have been positioned.  For example, correcting
the bounding box of a PostScript graphics object can foul up the
labels by changing the coordinate grid to which the labels are
attached. This is particularly true for the trimming  commands of
boxedeps.tex and BoxedArt.tex. However, as noted already, change
of scale is much less disruptive, and modest adjustments should be
well tolerated.

     Sometimes the labels protrude so far from the bounding box
of a figure that the figure has to be repositioned.  Best do this
by ad hoc spacing, say using \hglue and \vglue; altering the
bounding box would create a vicious circle.

     Remember that you are responsible for preventing labels
from overlapping. You are responsible for all label typography
including size and style. A label is really just about anything
that can be put in a TeX box. Note that spaces at the beginning
and end of labels will normally be suppressed; if you really want
them you must protect them with TeX braces.

     This package temporarily sets the \mathsurround parameter
of TeX to zero  while the labels are being affixed. This is done
because nonzero \mathsurround space would influence the position
of left and right aligned labels; then, when a texpert or printer
modifies mathsurround, diagram labeling might be disastrously
altered. There is a small price to pay involving labels that are
formatted as caption boxes including mathematics: you  may want or
need to specify an explicit mathsurround space within the caption
box; it will not influence anything outside.

     Those hostile to the use of * as separator between
the X and Y coordinates of label insertion points, are free to
impose another using \SetXYSeparator{<the new separator>}.  
Americans may prefer "," to "*" since they never use a 
comma as a decimal point; on the other hand, * may be more visible.

APPENDIX (I)  MERGING labelfig.tex LABELS INTO AN EPSF GRAPHICS OBJECT.

     As promised in the introduction, here is a recipe useful for
publishers. It works at least on Macintosh and at least for vectorized
graphics and Adobe type1 fonts.  (There is surely a similar recipe for
PCs under MSWindows.)

 (a)  Use boxedeps.tex utility to integrate the figure given by the eps
file, "x.eps" say, with a visible frame around it.  See
\ShowDisplacementBoxes command in boxedeps.tex.  To get precise results
automatically it is important to use the \Trim... commands of
boxedeps.tex making the "DisplacementBox" neatly fit the figure.

 (b)  Use the TeX printer driver and LaserWriter (versions >= 8.1.1) to
export to an EPSF the DVI page containing the integrated, labelled
figure. You now have an EPS file  "xx.eps"  that contains too much, and at
the wrong scale, and at wrong position.

 (c)  Convert the EPSF to an Adode Illustrator format EPSF using
the shareware utility called epsConvert by Sam Weiss
1993-- (currently $25).

 (d)  In Illustrator (or a compatible program), group the labels and the
"DisplacementBox"; copy them to the clipboard and paste them into "x.ps".
This step requires that all the label fonts be "visible to the Macintosh.

 (e)  Translate and scale the pasted group consisting of the labels plus
the "DisplacementBox" so as to make the "DisplacementBox" the bounding
box of (labelless) figure represented by "x.eps".  At this point the
labels will be correctly placed on the figure "x.eps".

 (f)  Ungroup and delete the "DisplacementBox".  The result is the
desired single EPS file, "x+.eps" say, It contains the original figure
plus its labels.  

     Using grouping and ungrouping appropriately in "x+.eps", a
publisher's staff can very efficiently improve label positions etc.

APPENDIX II)  SOME EXOTIC APPLICATIONS

     The grid of labelfig.tex is analogous to a light-table in
classical page makeup with wax or latex glue.  In principle, you
can use it to compose any page from its indivisible parts.  This
even has some of the artisanal charm of classical paste-up
provided you have a fast screen preview to make the process
"interactive".

     In practice labelfig.tex is a tool for nonstandard jobs.
Here are a few going beyond the labelling already discussed.

(I)  GRAPHICS INTEGRATION.

     This is accomplished by treating the imported graphics
objects as labels.  The underlying graphics object is then
typically an empty  \vbox to <dimension>{\vfill} in a TeX
\midinsert...\endinsert construction.  A label line
might be of the form

   (.1*.1) \\

The exact form of the special command varies from driver to
driver.  However, in the case of encapsulated PostScript graphics
(EPSF norm), by relying on boxedeps.tex, one can have the
following standard syntax (independant of driver  (see
boxedeps.doc for details.
  
  (.1*.1) \BoxedEPSF{MyFigure scaled <scale in mils>}\\

This may be slow since it requires TeX to read the PostScript
file to read bounding box using many complex macros.  So you
may want to try

  (.1*.1) \EPSFSpecial{MyFigure}{<scale in mils>}\\

which is fast and driver independant, but it squashes the
bounding box, normally to its lower left corner.

     Similarly for graphics of the Macintosh PICT norm ---
using BoxedArt.tex (same sources) in place of boxedeps.tex.

     This approach to integration is to be recommended when
one is assembling a composite graphics object.

 (II)  COMMUTATIVE DIAGRAM ENHANCEMENT

     Commutative diagrams or arrays of mathematical objects
connected by arrows of various sorts are common in mathematics.
The mathematical objects require the use of TeX.  Recently TeX
acquired a good collection of arrows of all slopes --- that of
LamSTeX --- plus pwerful macros to build the diagrams.

     However, even the LamSTeX collection is often
inadequate; it lacks for example double shafted arrows, dotted
arrows and curved arrows. Fortunately it is possible to produce
such arrows on an individual basis using sophisticated graphics
programs such as Illustrator and AldusFreehand (both serving
the EPSF norm) or using Metafont (with its public domain norm).
Since the creation of each new arrow is a work of love, you
probably want to limit the number of arrows by using LamSTeX
for most arrows. The 40K commutative diagram module of LamSTeX
has been adapted to work with AmSTeX and a copy may be posted
with LabelFig and related files. Unfortunately no one has yet
offered a version that works with Plain TeX or LaTeX.

       Suffice it here to say that when the exotic arrow has
been somehow imported into TeX, labelfig.tex treats it as a
label that one affixes to the commutative diagram.  Two other
steps will be treated in separate notes, namely the matter of
extracting the dimension specifications for the arrow and the
construction of the arrow --- for these steps are far from
unique and often depend intimately on your computer environment. 
Notes for the Macintosh-Textures-Illustrator combination are
found in the file ExoticArrows.doc.

 (III) NESTING 

Ingenuity pays off in exploiting labelfig.tex. One can
mix graphics and typography quite freely.  labelfig.tex is good
for freeform or overlapping arrangements, while boxedeps.tex (or
BoxedArt.tex) is best for regimented non-overlapping
arrangements --- and the two can be combined.

     The default behavior of labelfig.tex is not ideal 
for nesting objects, because to prevent trouble for beginners
the register for labels is globally cleared when \AffixLabels
concludes.  But there are switches available

      \LabelsGlobal      \LabelsLocal

which change this.  To understand this, extend the above test 
by something like:

 \LabelsLocal

 \SetLabels
    (.5*.5) AAA\\
 \endSetLabels

 {
 \SetLabels
    (.5*.5) ZZZ\\
 \endSetLabels
   \AffixLabels{\FirstQuadrant}
 }

   \AffixLabels{\FirstQuadrant}

     There are however potential pitfalls.  Neither
labelfig.tex nor boxedeps.tex has been tested under extreme
conditions. Problems may occur if their procedures are
indiscriminately nested. For boxedeps.tex (not labelfig.tex)
there is a precise cause for worry, namely many of its
variables are "global", which means that TeX braces will not
provide the protection one might expect.

COMMAND SUMMARY FOR labelfig.tex

  Here [...] means optional (one or zero)
       [...]* means any number of such constructs

  \SetLabels
    [[<P>](<X><Sep><Y>) <label> \\]*
  \endSetLabels
  \ShowGrid  
  \AffixLabels{<the figure>}

   --- <P> is tack position, one of eleven or empty
              order irrelevant

                   \L\T      \T      \R\T

                   \L\E      \E      \R\E

                     \L               \R

                   \L\B      \B      \R\B

   --- (<X><Sep><Y>) insertion point;
  <Sep> is separator, = * by default;
  \SetXYSeparator{<Sep>} changes it.
   <X> and <Y> are real numbers

  --- <label> a label to attach 

  --- <the figure> the figure to label 

  \GlobalLabels (default)     
  \LocalLabels  setting for nested constructs.

 \Grids makes ALL grids appear; \HideGrid then makes just next disappear.
 \noGrids returns to default.  The commands are always global.

 \GridLineWidth{<dimension>} adjusts width of grid lines. Default is very
small, to give "hairline" effect. If your grid lines are missing try
setting \GridLineWidth{1pt}.

 \Edges#1 globally changes the edge size of all grids to the numerical 
value #1, which must be 1, 10, 100, or 1000.  The default is 1.

VERSION HISTORY.
 --- Jan 1993: \Edges#1 and [??] option after \SetLabels
 --- July 1992: \Grids, \noGrids, \HideGrid;
       Gridlines become hairlines; \GridLineWidth{<dimension>}.
 --- Oct 1991, Jan 1992: \SetXYSeparator{<Sep>},  \LabelsGlobal,
       \LabelsLocal.
 --- July 1991: first release

Address for bugs and other feedback:

        Raymond S\'eroul
        IREM and Lab. de Typographie Informatise
        Univ. Rene Descartes
        Strasbourg

    Tel 33-88-41-63-45
    Email:  A18645@FRCCSC21.BITNET

        Laurent Siebenmann
        Mathematique, Bat. 425,
        Univ de Paris-Sud,
        91405-Orsay,
        France

    Tel 33-1-6941-7949; 
    Email: lcs@topo.math.u-psud.fr  

\usepackage{amsmath,amssymb,amsfonts,latexsym,verbatim,color,epsfig,psfrag}
\usepackage{times,multirow,multicol, array}
\usepackage{algorithm,algorithmic}
\usepackage{setspace}
\usepackage{float}
\usepackage{caption}
\usepackage{subcaption}
\usepackage{adjustbox}

\newcolumntype{L}[1]{>{\raggedright\let\newline\\\arraybackslash\hspace{0pt}}m{#1}}
\newcolumntype{C}[1]{>{\centering\let\newline\\\arraybackslash\hspace{0pt}}m{#1}}
\newcolumntype{R}[1]{>{\raggedleft\let\newline\\\arraybackslash\hspace{0pt}}m{#1}}

\icmltitlerunning{Dimension-Wise Importance Sampling Weight Clipping for Sample Efficient Reinforcement Learning}

\begin{document}
	
	\twocolumn[
	\icmltitle{Dimension-Wise Importance Sampling Weight Clipping  for Sample-Efficient Reinforcement Learning}
	
	
	
	
	\begin{icmlauthorlist}
		\icmlauthor{Seungyul Han}{kaist}
		\icmlauthor{Youngchul Sung}{kaist}
	\end{icmlauthorlist}
	
	\icmlaffiliation{kaist}{School of Electrical Engineering, KAIST, Daejeon, South Korea}
	\icmlcorrespondingauthor{Youngchul Sung}{ycsung@kaist.ac.kr}
	
	\icmlkeywords{Machine Learning, ICML}
	
	\vskip 0.3in
	]
	
	
	
	\printAffiliationsAndNotice{}  

	\begin{abstract}
		In importance sampling (IS)-based reinforcement learning algorithms such as Proximal Policy Optimization (PPO), IS weights are typically clipped to avoid large variance in learning. However, policy update from clipped statistics induces large bias in tasks with high action dimensions, and bias from clipping makes it difficult to reuse old samples with large IS weights.
		In this paper, we consider PPO, a representative on-policy algorithm, and propose its improvement by  dimension-wise IS weight clipping which separately clips the IS weight of each action dimension to avoid large bias and adaptively controls the IS weight to bound policy update from the current policy.  This new technique enables efficient learning for high action-dimensional tasks and reusing of old samples like in off-policy learning to increase the sample efficiency.
		Numerical results show that the proposed new algorithm outperforms  PPO and other RL algorithms in various Open AI Gym tasks.
	\end{abstract}

	\section{Introduction}
	\label{sec:intro}

	Many deep reinforcement learning (RL) algorithms  aim to maximize the discounted return for a given environment by a parameterized policy \cite{mnih2013playing,mnih2015human, lillicrap2015continuous, schulman2015trust}.
	For learning continuous action tasks, which is the focus of this paper,   many algorithms are based on policy gradient (PG), which    updates the policy parameter to the  direction of gradient decent of an objective function \cite{sutton2000policy}.
	PG methods can be classified into  on-policy PG and off-policy PG. On-policy PG updates the target policy from the samples generated by the target policy only \cite{schulman2015trust,peters2005natural}, whereas off-policy PG updates the target policy from samples generated by the behavior policy which can be different from the target policy. Off-policy PG methods has the advantage that they can reuse old sample batches and set the behavior policy to better explore state and action spaces, and thus many state-of-the-art algorithms are based on off-policy PG.
	Current off-policy PG methods can be divided into several types. Value-based  PG methods update the policies  to maximize the Q-value statistics estimated by TD errors \cite{lillicrap2015continuous,fujimoto2018addressing,haarnoja2017rein} and IS-based PG methods calibrate the statistics differences between the target policy and the behavior policy \cite{degris2012off, wang2016sample, levine2013guided}. Due to the advantages of off-policy learning, reusing off-policy data in on-policy learning is recently considered \cite{gu2016q,gu2017interpolated,nachum2017trust,han2017amber}.

	In this paper, we consider PPO \cite{schulman2017proximal}, which is a widely-considered stable and fast-converging on-policy algorithm, and propose its improvement by adopting several new features. PPO uses a clipped objective function to bound policy update  from the current policy for stable learning. However, due to its structure, the gradient of clipped samples completely vanishes and this causes sample inefficiency in high action-dimensional tasks. In order to solve this problem, we propose {\em  dimension-wise importance sampling weight clipping (DISC)}, which  clips the IS weight of each action dimension instead of clipping the overall likelihood ratio as in PPO. With DISC, the gradient of samples does not completely vanish for most samples since there  exist unclipped dimensions, and this increases the sample efficiency.  In order to further enhance the sample efficiency, we propose reusing old samples generated at the previous iterations like in off-policy learning. PPO uses Generalized Advantage Estimator (GAE) \cite{schulman2015high} to estimate the advantage of the on-policy $n$-step trajectory, but GAE has difficulty in estimating from the old trajectories generated at the previous iterations. To circumvent this difficulty,  we consider  V-trace \cite{espeholt2018impala} that estimates target values from off-policy trajectories and combine GAE with V-trace to estimate the advantage in DISC. We evaluate our algorithm on various OpenAI GYM tasks \cite{brockman2016openai}  and numerical results show that DISC outperforms the PPO baseline and other state-of-the-art continuous RL algorithms.

	\vspace{0.5em} {\em Notations:}  ${\mathbb{E}}[\cdot]$ denotes expectation, $\textrm{sgn}(\cdot)$ is the sign function of $(\cdot)$, and $x \sim P$  means that random sample $x$ is generated from distribution $P$. $\hat{X}$ denotes the empirical estimation of random variable $X$. $\mathcal{N}(\mathbf{u},\mathbf{\Sigma})$ represents  Gaussian distribution with mean vector $\mathbf{u}$ and covariance matrix $\mathbf{\Sigma}$.

	\section{Related Works}
	\label{sec:related}

	\textbf{IS-based PG:} Calibration based on IS weight has been considered in various off-policy PG methods for unbiased objective function estimation since  the target policy and the behavior policy are different in off-policy RL \cite{meuleau2000off, shelton2001importance, peshkin2002learning, jie2010connection}. Large IS weights cause large variances and hence  truncation of IS weights is proposed  to prevent large IS weights \cite{wawrzynski2009real}. However, truncation induces large bias. The off-policy actor-critic algorithm \cite{degris2012off} reduces variance by applying per-decision IS (not the whole trajectory) in actor-critic. The actor-critic with experience replay (ACER) algorithm further applies bias correction in addition to truncated IS of actor-critic to use off-policy data with low variance \cite{wang2016sample}.

	\textbf{Value estimation:} One can estimate the Q-function with 1-step TD error which can be used for both on-policy and off-policy. In addition, there exist many $n$-step value estimation methods to improve accuracy.
	For on-policy PG, TD($\lambda$) \cite{sutton1988learning}, Least-Squares TD($\lambda$) \cite{bradtke1996linear}, GAE \cite{schulman2015high} are commonly used. For off-policy PG, a $n$-step trajectory is generated from the behavior policy, so the path statistic should be calibrated to evaluate the target policy.
	Thus,  IS weight is applied in off-policy evaluation: \cite{precup2000eligibility} suggested using per-decision IS weight, and  Retrace ($\lambda$) \cite{munos2016safe} and V-trace \cite{espeholt2018impala} evaluate the policy by using the truncated trace for low-variance target value estimation.

	\textbf{Extension of on-policy PG using off-policy data:}  In on-policy learning, the target policy and the behavior policy are the same and hence samples generated by the target policy can be used only at the current iteration  \cite{schulman2015trust}. Recently, in order to improve the sample efficiency of on-policy learning, using off-policy data to update on-policy gradient is proposed \cite{gu2016q, gu2017interpolated}. In particular, Trust-PCL \cite{nachum2017trust} applies path consistency learning to use off-policy data while maintaining the stability of trust region policy optimization.

	\textbf{Extension of PPO:} There exist several improvements of PPO, e.g., distributed optimization for enhancing the performance of large scale RL \cite{heess2017emergence}, reusing old sample batches with small IS weights  \cite{han2017amber}, and reducing variance by using action-dependent stein control variate \cite{liu2017sample}.

	\section{Set Up and Background}
	\label{sec:background}

	We assume a Makrov decision process (MDP) environment $(\mathcal{S},\mathcal{A},P,\gamma,R)$, where $\mathcal{S}$ is the state space, $\mathcal{A}$ is the action space, $P$ is the transition probability, $\gamma$ is the discount factor, and $R$ is the reward function. The agent has  policy $\pi : \mathcal{S}\rightarrow \mathcal{A}$ that selects action $a_t$ for given state $s_t$,
	and interacts with the environment to generate  an episode $(s_t,a_t,R(s_t,a_t))_{t=0}^{T-1}$ of length $T$.  The agent optimizes its policy to maximize the discounted return  of the episode $\sum_{t=0}^{T-1}\gamma^t R(s_t,a_t)$, and  a parameterized policy $\pi_\theta$ is assumed in this paper.

	The basic idea of PG is as follows. The objective function of discounted return is defined as a function of the policy parameter $\theta$: $J(\theta) = \mathbb{E}_{\tau_0 \sim \pi_\theta}[\sum_{t=0}^{T-1}\gamma^t r_t]$, where $r_t=R(s_t,a_t)$, and $\tau_t = (s_t,a_t,\cdots,s_{T-1},a_{T-1})$ is the sample trajectory from time $t$.  The policy gradient theorem \cite{sutton2000policy} gives the gradient of $J(\theta)$ with respect to $\theta$ as
	\[ 
	\nabla_\theta J(\theta) = \mathbb{E}_{s_t\sim \eta_{\pi_\theta}, a_t \sim \pi_\theta}[A_{\pi_\theta}(s_t,a_t) \nabla_\theta \log\pi_\theta (a_t|s_t) ],
	\] 
	where $\eta_{\pi_\theta}$ is the discounted state visitation frequency, and $A_{\pi_\theta}(s_t,a_t) = Q_{\pi_\theta}(s_t,a_t)-V_{\pi_\theta}(s_t)$ is the advantage function with the state-action value function $Q_{\pi_\theta}(s_t,a_t) = \mathbb{E}_{\tau_t \sim \pi_\theta}[\sum_{t=i}^{T-1}\gamma^{t-i} r_t]$ and the state value function $V_{\pi_\theta}(s_t) = \mathbb{E}_{a_t,\tau_{t+1} \sim \pi_\theta}[\sum_{t=i}^{T-1}\gamma^{t-i} r_t]$. Then, the policy parameter $\theta$ can be  optimized by stochastic gradient descent to the direction $\nabla_\theta J(\theta)$ to maximize the return.
	In the case that the behavior policy $\mu$ that generates samples is different from the target policy $\pi_\theta$,  the objective function can be calibrated by using importance sampling weight as \cite{degris2012off}
	\[
	\nabla_\theta J_{\textrm{off}}(\theta) = \mathbb{E}_{s_t\sim \eta_{\pi_\theta}, a_t \sim \mu}[\rho_t A_{\pi_\theta}(s_t,a_t) \nabla_\theta \log\pi_\theta (a_t|s_t)],
	\]
	where $\rho_t = \frac{\pi_\theta(a_t|s_t)}{\mu (a_t|s_t)}$ is the IS weight between $\pi_\theta$ and $\mu$.

	PPO  is based on the minorization and maximization (MM) technique for the original objective function $J(\theta)$ and  is  a widely-considered on-policy PG method that updates the policy $\pi_\theta$ multiple  times by sampling mini-batches from the current sample batch.
	In the beginning of the $i$-th iteration, the current sample batch $B_i = \{(s_{i,0},a_{i,0},r_{i,0}),\cdots,(s_{i,N-1},a_{i,N-1},r_{i,N-1})\}$ of length $N$ is generated by $\pi_{\theta_{i}}$. Then, it updates the target policy $\pi_\theta$ by sampling mini-batches from $B_i$ for several epochs. Since the target policy $\pi_\theta$ for policy update is different from the policy $\pi_{\theta_i}$ that generated $B_i$,  PPO calibrates the statistics difference between the current target policy $\pi_\theta$ and the policy $\pi_{\theta_i}$ that generated $B_i$ based on the IS weight $\rho_t = \frac{\pi_\theta(a_t|s_t)}{\mu (a_t|s_t)}$ with $\mu = \pi_{\theta_i}$.
	Furthermore, PPO clips the IS weight to bound the amount of policy update for stable learning \cite{schulman2015trust, wang2016sample}. Thus,  the  objective function of PPO is given by
	\begin{align}
	\hat{J}_{PPO}(\theta) &= \frac{1}{M} \sum_{m=0}^{M-1} \min \{ \rho_m \hat{A}_m, \textrm{clip}_\epsilon(\rho_m) \hat{A}_m\},\label{eq:ppo} \\
	&= \frac{1}{M} \sum_{m=0}^{M-1} \min \{\kappa_m \rho_m , \kappa_m \textrm{clip}_\epsilon(\rho_m) \} \kappa_m\hat{A}_m, \nonumber
	\end{align}
	where  $\rho_t=\frac{\pi_\theta(a_t|s_t)}{\pi_{\theta_i}(a_t|s_t)}$, $\hat{A}_t$ is an estimate of $A_{\pi_{\theta_i}}(s_t,a_t)$, $\textrm{clip}_\epsilon(\cdot):=\textrm{clip}(\cdot,1-\epsilon,1+\epsilon)$ is the clip function, $\kappa_m=\mathrm{sgn}(\hat{A}_m)$, and the $M$ samples of each mini-batch are randomly sampled from $B_i$.
	For empirical estimation of $A_{\pi_{\theta_i}}$,  PPO uses GAE \cite{schulman2015high}.   GAE estimates the advantage as $\hat{A}_t = \sum_{l=t}^{T-1} (\gamma\lambda)^{l-t} \delta_{l}$, where $\delta_t = r_t + \gamma V_{w_i}(s_{t+1}) - V_{w_i}(s_t)$ is the temporal difference (TD) residual, $V_w$ is the parameterized value function that approximates $V_{\pi_{\theta}}$, and $w_i$ is the value parameter at the $i$-th iteration. The value parameter $w$ is updated to minimize the mean square error (MSE) between $V_w$ and the target value $\hat{V}_t = \hat{A}_t + V_{w_i}(s_t)$, given by
	$\hat{J}_{V}(w)= \frac{1}{M}(V_w(s_m)-\hat{V}_m)^2$.
	GAE can be  computed recursively as $\hat{A}_t = \delta_t + (\gamma\lambda) \hat{A}_{t+1}$ in a similar way to that  TD($\lambda$) backup performs value estimation \cite{sutton1998reinforcement}.

	\section{Dimension-Wise Clipping}
	\label{sec:dimclip}

\begin{figure}
		\centering
		\includegraphics[width=0.23\textwidth]{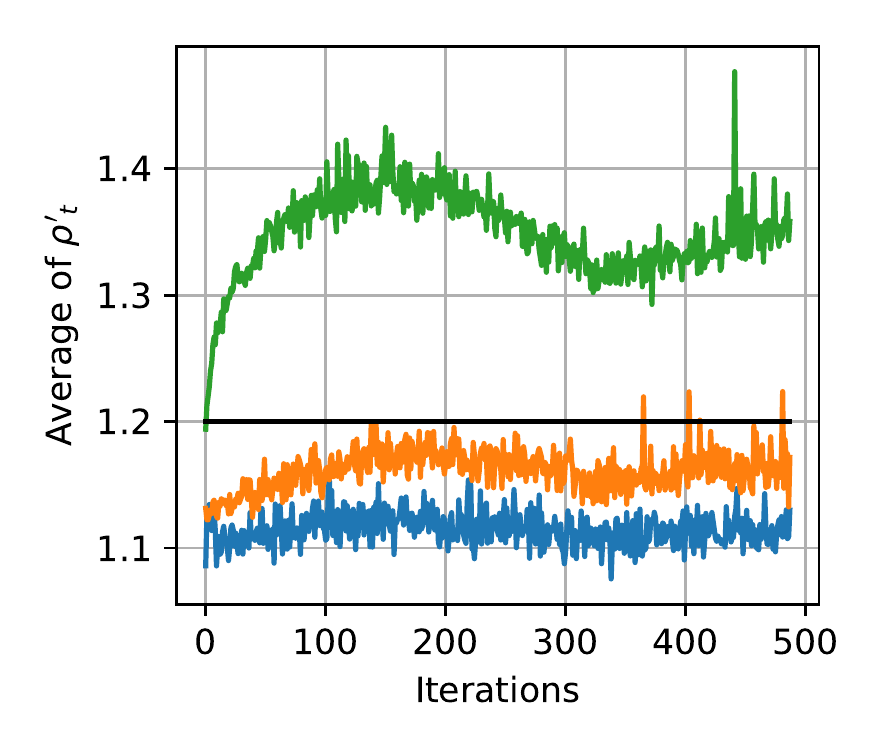}
		\includegraphics[width=0.23\textwidth]{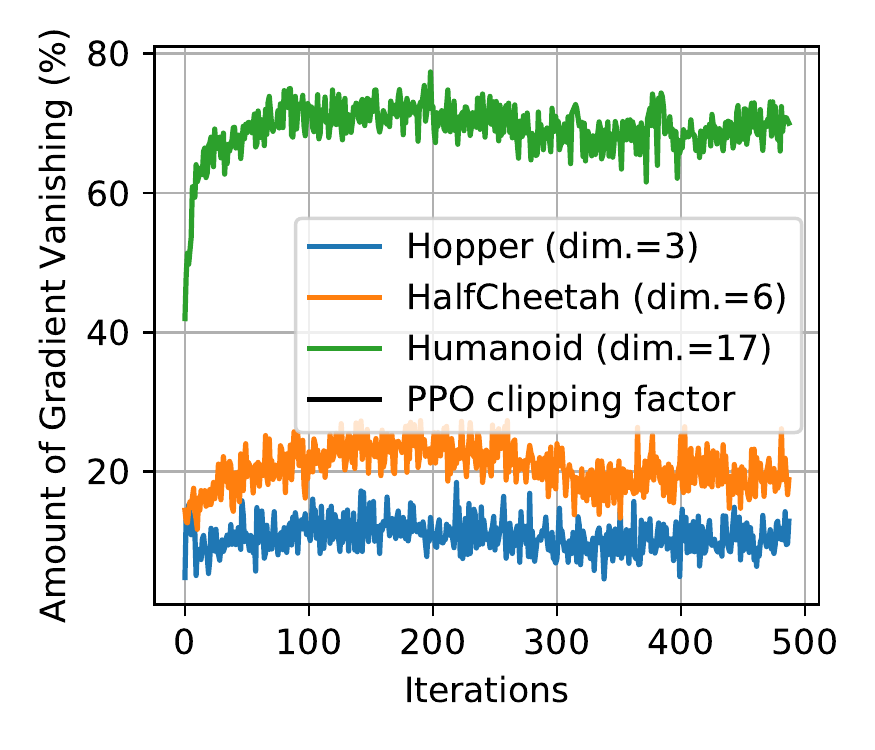}
		\caption{(left) $\rho'_t$ and (right) the amount of gradient vanishing}
		\label{fig:vgrppo}
	\end{figure}

	Note that in the PPO objective function \eqref{eq:ppo}, $\rho_t$ is a function of the optimization variable $\theta$ but $\hat{A}_t$ is fixed for the given behavior policy $\pi_{\theta_i}$ which generated the current sample batch $B_i$. Hence, on average, maximization of the cost \eqref{eq:ppo} with respect to $\theta$ increases
	$\rho_t$  if $\hat{A}_t>0$, and decreases $\rho_t$ if $\hat{A}_t < 0$. For stable update,  PPO  limits the amount of the policy update by clipping the objective function unlike TRPO restricts the policy update by using a KL divergence constraint.
	This clipping prevents $\rho_t$ from becoming  too large or too small, and  results in stable update for many environments.
	The clipped objective function of PPO $\hat{J}_t =\min\{\rho_t\hat{A}_t,\textrm{clip}_\epsilon(\rho_t)\hat{A}_t)\}$ reduces to a constant $\hat{J}_t = (1+\epsilon)\hat{A}_t$ when $\hat{A}_t>0$ and  $\rho_t>1+\epsilon$, and to a constant $\hat{J}_t = (1-\epsilon)\hat{A}_t$ when $\hat{A}_t<0$ and $\rho_t<1-\epsilon$. Therefore, the gradient vanishes for such  samples.
	This zero-gradient can cause a serious problem especially in high action-dimensional tasks.
	In order to assess the amount of gradient vanishing,  we consider the average shifted deviation of the IS weight from one $\rho'_t:=|1-\rho_t|+1$\footnote{To measure how much IS weight is far from $1$, we consider $\rho'_t:=|1-\rho_t|+1$.} and the corresponding amount of gradient vanishing, which are shown in
	Fig. \ref{fig:vgrppo}. Here, the average $\rho'_t$ at the $i$-th iteration
	is defined as the average of IS weight deviation $\rho'_t$ of each mini-batch averaged again over all drawn mini-batches at the last epoch and the the amount of gradient vanishing is defined as the ratio of the samples with clipped IS weights to the total samples at the last epoch. It is seen that for the low action-dimensional tasks such as Hopper and HalfCheetah whose action dimensions are 3 and 6, respectively,  average $\rho'_t$ is  below the clipping factor of PPO (the black line) and the amount of gradient vanishing is not significant ($15\sim 20 \%$). However, for the high action-dimensional tasks such as Humanoid whose action dimension is 17, average $\rho'_t$  exceeds the clipping line (the reason will be explained shortly) and the corresponding amount of gradient vanishing is too large (above $60\%$). Hence, clipping the loss function directly and the corresponding zero gradient samples in PPO reduce sample efficiency for high action-dimensional tasks.

	\subsection{Dimension-Wise Clipping}
	\label{subsec:dimclip}

	For learning continuous action tasks, many RL algorithms use independent Gaussian distribution $\pi_{\theta}(\cdot|s_t) = \mathcal{N}(\mathbf{u},\sigma^2\mathbf{I})$ as the stochastic policy, where $\mathbf{\mu} = (u_0,u_1,\cdots,u_{D-1})$ is the mean vector, $D$ is the action dimension, and $\mathbf{I}$ is an identity matrix. Then, the policy can be factorized into the action dimensions as $\pi_\theta (a_t|s_t) = \prod_{d=0}^{D-1} \pi_{\theta,d} (a_{t,d}|s_t)$, where $\pi_{\theta,d} (\cdot|s_t) \sim \mathcal{N}(\mu_d,\sigma^2)$ and $a_{t,d}$ is the $d$-th element of $a_t$.
	Suppose that the amount of policy distribution change by update is similar for each dimension. Then, the variation of $\pi_\theta$  grows exponentially as $D$ increases. This yields  large IS weights for high action-dimensional tasks, as seen in the case of Humanoid in Fig. \ref{fig:vgrppo}.
	In order to prevent large IS weights in high action-dimensional tasks while maintaining the advantage of IS weight clipping, we propose  a new objective function for policy optimization, which  clips the IS weight of each dimension separately. That is,
	using the fact that the IS weight can be factorized as
	\[
	\rho_t = \frac{\pi_{\theta}(a_t|s_t)}{\pi_{\theta_i}(a_t|s_t)} = \prod_{d=0}^{D-1}\frac{\pi_{\theta,d}(a_{t,d}|s_t)}{\pi_{\theta_i,d}(a_{t,d}|s_t)}=:\prod_{d=0}^{D-1}\rho_{t,d},
	\]
	we clip the IS weight $\rho_{t,d}$ of each dimension as $\textrm{clip}_\epsilon(\rho_{t,d})$. In addition, in order to prevent the IS weight from being too far from $1$, we include an additional loss for IS weight control:
	$J_{IS} = \frac{1}{2M}\sum_{m=0}^{M-1}(\log(\rho_m))^2$.
	Thus, our objective function for DISC is given by
	{\small
		\begin{align}
		& \hat{J}_{DISC}(\theta) \nonumber\\
		&=\frac{1}{M}\sum_{m=0}^{M-1}\left[\prod_{d=0}^{D-1} \min\{\kappa_m \rho_{m,d},\kappa_m\textrm{clip}_\epsilon(\rho_{m,d})\}\right] \kappa_m\hat{A}_m \nonumber\\
		&~~~~- \alpha_{IS}J_{IS}, \label{eq:disc}
		\end{align}
	}where $\kappa_m = \textrm{sgn}(\hat{A}_m)$, and the weighting factor $\alpha_{IS}$ is adaptively determined in a similar way of adapting the weighting factor for the KL divergence in PPO \cite{schulman2017proximal}:
	\begin{equation}\label{eq:IS}
	\left\{
	\begin{array}{ll}
	&\alpha_{IS}\leftarrow\alpha_{IS}/2~~~~~\textrm{if } J_{IS} < J_{targ}/1.5\\
	&\alpha_{IS}\leftarrow\alpha_{IS}\times 2~~\textrm{if } J_{IS} > J_{targ}\times 1.5
	\end{array}
	\right.
	\end{equation}

\begin{figure}
		\centering
		\includegraphics[width=0.23\textwidth]{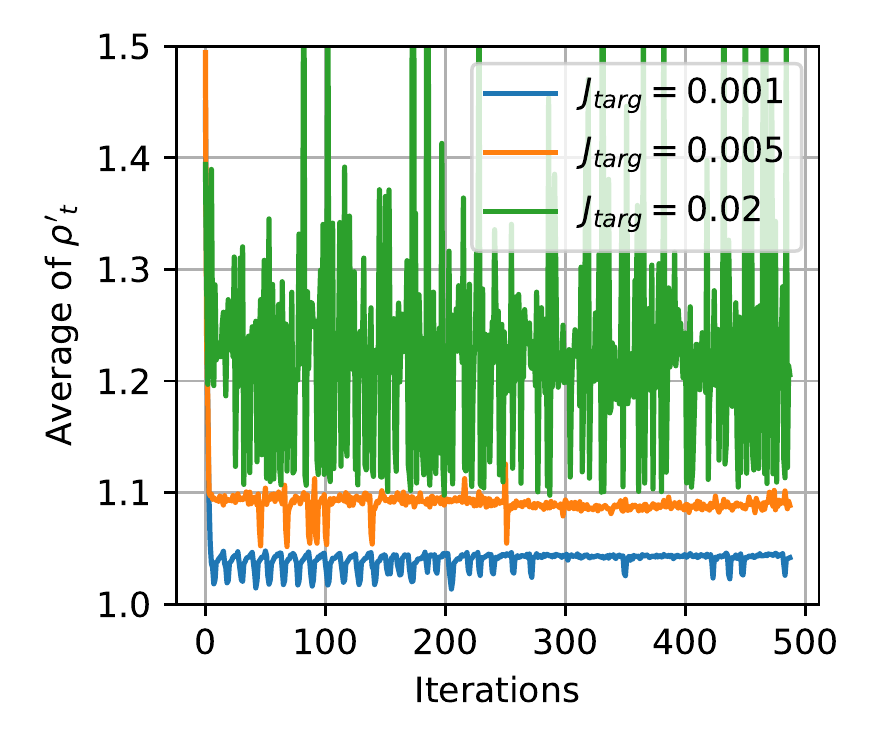}
		\includegraphics[width=0.23\textwidth]{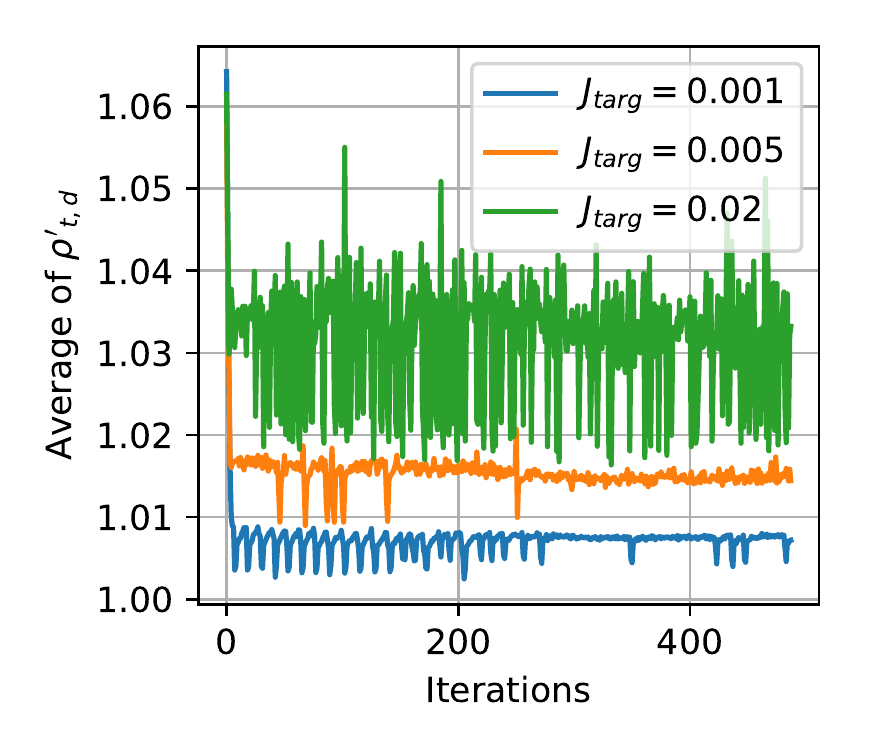}
		\caption{(left) average $\rho'_t$  and (right) average $\rho'_{t,d}$ (averaged over all samples and all dimensions) of DISC for Humanoid}
		\label{fig:ISdisc}
	\end{figure}

	DISC solves the vanishing gradient problem of PPO by dimension-wise clipping. Even though $\min\{\kappa_m\rho_{m,d}, \kappa_m\textrm{clip}_\epsilon(\rho_{m,d})\}$ becomes constant from clipping for some $d$, there exist other dimensions that are not clipped for most samples. Thus, we can update the policy by the gradient from the unclipped objective function unless $\rho_{t,d}$ are clipped for all $d$.
	Fig. \ref{fig:ISdisc} shows average $\rho'_t$ and average $\rho'_{t,d}:=|1-\rho_{t,d}|+1$ of DISC for $J_{targ} = 0.0001,~0.0002,~0.0005$  for the Humanoid task, and the corresponding amount of gradient vanishing is almost zero.
	Note that average overall $\rho'_t$ of DISC is much smaller as compared to the Humanoid curve in Fig. 1(left) especially for small $J_{targ}<0.005$. Furthermore,   average $\rho'_{t,d}$ for each dimension is much less than $\rho'_t$, as expected. Hence, dimension-wise IS clipping efficiently reduces gradient vanishing as well as increases sample efficiency.

	\subsection{Reusing Old Sample Batches}
	\label{subsec:reusing}

	In order to further enhance sample efficiency, we reuse old sample batches. The experience replay buffer $\mathbf{R}$ stores old sample batches $B_{i-1},\cdots,B_{i-L+1}$ in addition to the current sample batch $B_i$, where $L$ is the maximum number of old sample batches. If $L=1$, then the algorithm is on-policy, and otherwise ($L>1$), it is off-policy. We sample mini-batches multiple times in the whole buffer $\mathbf{R}$. However, if the IS weights of samples in the old batches deviate too much from $1$, then the weights  will be clipped and it is sample inefficient. Thus, we follow \cite{han2017amber}, which considers reusing old sample batches  only if the IS weights of samples in the old batches does not deviate too much from $1$. That is,  if for a given old batch
	\begin{equation}\label{eq:ambercond}
	\frac{1}{N}\sum_{t=0}^{N-1}\rho'_t < 1+\epsilon_b,
	\end{equation}
	where $\epsilon_b$ is a parameter determining batch inclusion and $N$ is the number os samples in the batch, then the old batch is reused. However, in our DISC, we slightly modify  the inclusion criterion and reuse old sample batches satisfying the condition
	\begin{equation}\label{eq:batchcond}
	\frac{1}{ND}\sum_{t=0}^{N-1}\sum_{d=0}^{D-1}\rho'_{t,d} < 1+\epsilon_b.
	\end{equation}
	Here, we reuse old sample batches only for computation of the first term in the right-hand side (RHS) of \eqref{eq:disc} not for the second term $J_{IS}$. $J_{IS}$ is computed by using the on-policy batch $B_i$ to implement  policy update near the current policy.
	
	\begin{figure}
		\centering
		\includegraphics[width=0.23\textwidth]{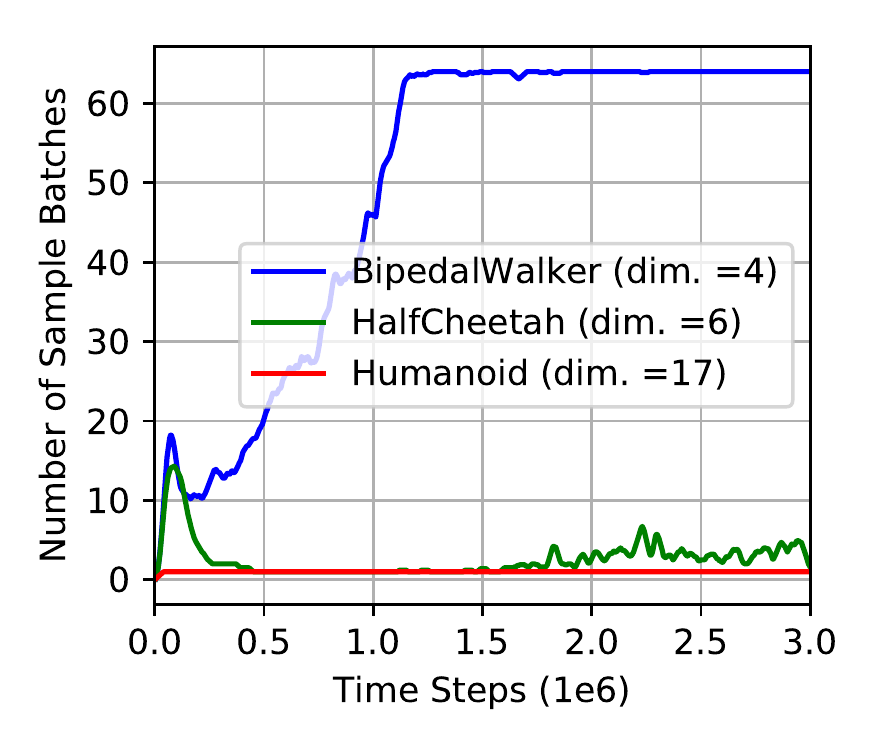}
		\includegraphics[width=0.23\textwidth]{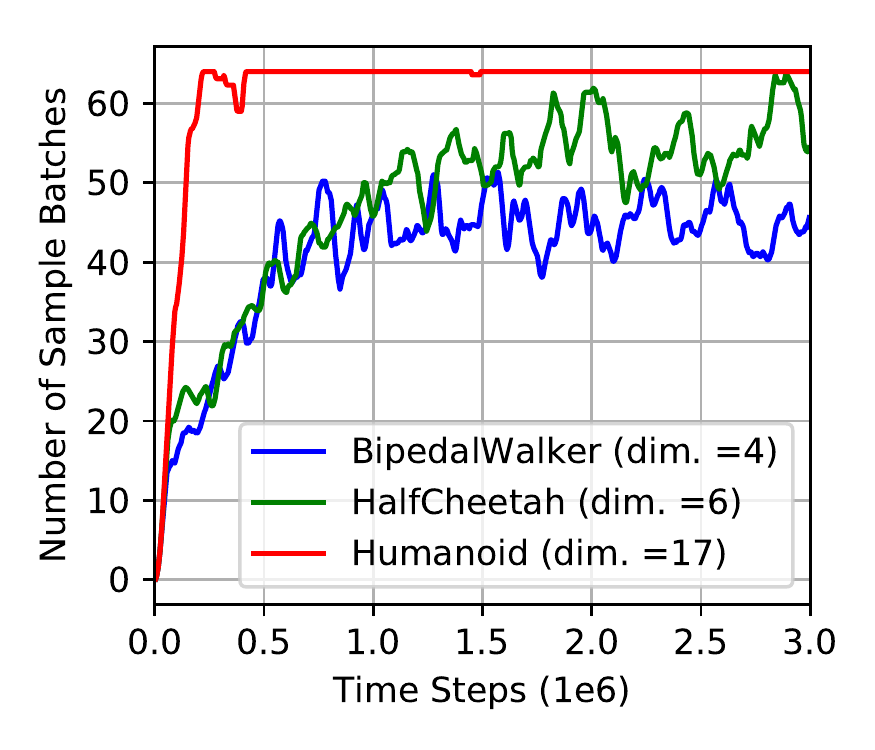}
		\caption{The number of reused sample batches: (left)  PPO-AMBER \cite{han2017amber}  and  (right) DISC}
		\label{fig:numbatch}
	\end{figure}
	
	Fig. \ref{fig:numbatch} shows the number of reused sample batches of PPO-AMBER \cite{han2017amber} and DISC for BidepalWalker, HalfCheetah, and Humanoid whose action dimensions are $4$, $6$, and $17$, respectively.
	PPO-AMBER selects more sample batches based on  \eqref{eq:ambercond} for
	lower action-dimensional tasks to avoid too much clipping from reusing samples, and it is ineffective for high action-dimensional tasks such as Humanoid, as seen in Fig. \ref{fig:numbatch}. Thus, it cannot use off-policy sample batches for high action-dimensional tasks.
	On the other hand, DISC can exploit more old sample batches for higher action-dimensional tasks since it determines old batch inclusion based on $\rho'_{t,d}$ which is not too big even in Humanoid.
	Old batch inclusion of DISC  matches  the purpose of experience replay of using  more samples for better exploration in high action-dimensional tasks.

	\subsection{GAE with V-trace}
	\label{subsec:gae-v}

	PPO computes the GAE $\hat{A}_t$ and the corresponding target value $\hat{V}_t$ to evaluate the current policy by using the on-policy sample batch $B_i$ at the beginning of iteration once. Thus, at the policy evaluation step, the evaluated policy is the same as  the behavior policy, and PPO does not need IS calibration. However, if we use old sample batches generated from the previous policies $\pi_{\theta_{i-1}},\pi_{\theta_{i-2}}\cdots$ which are different from the current behavior policy $\pi_{\theta_i}$, we need to apply IS calibration even for evaluating the current policy from old samples for   low variance and safe learning, like in  Retrace($\lambda$) \cite{munos2016safe} or V-trace \cite{espeholt2018impala}. In particular,
	V-trace evaluates the multi-step target value from an off-policy trajectory as
	\begin{equation}
	\hat{V}_t = V_{w_i}(s_t) + \sum_{l=t}^{T-1}  \left(\prod_{i=t}^{l}c_i\right) \gamma^{l-t}\delta_l,
	\end{equation}
	where $c_t = \lambda\min(1.0,\rho_t)$ is the truncated IS weight \cite{espeholt2018impala}. It is proven that the V-trace operator is a contraction mapping \cite{espeholt2018impala}. Note that V-trace becomes the GAE target value in the on-policy case since $\rho_t=1$. Thus, in order to estimate the advantages of the previous trajectories $B_{i-1},B_{i-2},\cdots$ in DISC, we combine GAE with V-trace (GAE-V),  given by
	\begin{equation}
	\hat{A}_t=\sum_{l=t}^{T-1}  \left(\prod_{i=t+1}^{l}\min(1.0,\rho_i)\right) (\gamma\lambda)^{l-t}\delta_l,
	\end{equation}
	and its target value $\hat{V}_t = \min(1.0,\rho_t)\hat{A}_t + V_{w_i}(s_t)$ is just the V-trace target value.  Combining the new features, the final DISC algorithm is summarized in Algorithm 1.
	
	\begin{algorithm}
		\begin{algorithmic}
			\STATE{Initialize parameters $\alpha_{IS}\leftarrow1$,  $\theta$, $w$, }
			\FOR{each iteration}
			\STATE{$\theta_{i}\leftarrow \theta,~w_{i}\leftarrow w$}
			\STATE{Sample on-policy trajectory $B_i$ of size $N$ from $\pi_{\theta_i}$}
			\STATE{Store $B_i$ in replay buffer $\mathbf{R}$}
			
			\STATE{Compute GAE-V target $\hat{A}_t,\hat{V}_t$ by using all off-policy trajectories $B_i,\cdots,B_{i-L+1}$ in $\mathbf{R}$}
			\FOR{each epoch}
			\FOR{each gradient step}
			\STATE{Sample mini-batch of size $M$ from the sample batches  satisfying \eqref{eq:batchcond} in $\mathbf{R}$}
			\STATE{Compute the DISC objective function $\hat{J}_{DISC}(\theta)$}
			\STATE{Compute the value loss $\hat{J}_V(w)$}
			\STATE{$\theta\leftarrow \theta+\beta\nabla_{\theta}\hat{J}_{DISC}(\theta)$}
			\STATE{$w\leftarrow w+\beta\nabla_w\hat{J}_V(w)$}
			\ENDFOR
			\ENDFOR
			\STATE{Update $\alpha_{IS}$ as \eqref{eq:IS}}
			\ENDFOR
		\end{algorithmic}
		\caption{DISC}
		\label{alg:DISC}
	\end{algorithm}

	\section{Experiments}
	\label{sec:experimens}
	
	In this section, we provide numerical results and ablation study on our DISC algorithm. The source code for DISC  is available at \href{http://github.com/seungyulhan/disc/}{http://github.com/seungyulhan/disc/}.

	\subsection{Simulation Setup}

	Detailed description of the hyper-parameters of PPO, PPO-AMBER and DISC is provided in Table A.1 in Appendix.
	We basically followed the hyper-parameter setup of PPO in the Open AI baseline \cite{baselines}. In PPO, the learning rate $\beta$ anneals from $0.0003$ to $0$, but we set a minimum learning rate as $0.0001$ to prevent convergence to local optimum.
	We set the mini-batch size $M = (\textrm{the number of sample batches}/\textrm{gradient steps per epoch})$, where the number of sample batches is computed based on \eqref{eq:ambercond} for PPO-AMBER and \eqref{eq:batchcond} for DISC at each iteration. Note that we unify the hyper-parameter setup for PPO, PPO-AMBER and DISC, and only consider the single actor case for all continuous tasks for fair comparison.  The hyper-parameters of all other algorithms is explained in Section A in Appendix.

	We used deterministic evaluation based on  $10$ episodes generated by the corresponding deterministic policy (i.e., the mean network $\mathbf{u}$ of the Gaussian target policy $\pi_{\theta_i}$) at the time of evaluation, i.e., at the end of each iteration.
	We averaged the performance over $5$ seeds  and used the moving average window of $10$ iterations for figures.  The solid line and shaded part in each episode reward figure represent the average episode reward sum and its standard deviation for all seeds, respectively. In  the  max average return tables, the bold value  is the best max average return among the compared algorithms and $\pm$ means its standard deviation.

	\subsection{Comparison with the PPO Baseline}

	First, we  compared DISC with two PPO baselines: PPO whose loss is described as $\eqref{eq:ppo}$ \cite{schulman2017proximal} and PPO-AMBER \cite{han2017amber} which simply uses old sample batches as \eqref{eq:ambercond} and uses GAE for advantage estimation. For PPO-AMBER, we consider two cases with different batch inclusion factors: $\epsilon_b=0.1,~0.2$. We evaluated the three algorithms
	on Mujoco simulator \cite{todorov2012mujoco} and Box2d engine \cite{catto2011box2d} in OpenAI GYM continuous control tasks \cite{brockman2016openai} which are  challenging environments with continuous action spaces, as described in Fig. \ref{fig:mujoco}. The dimensions of state and action spaces of each task are given in Table A.2 in Appendix.
	
	Fig. \ref{fig:perfbase} shows the average episodic reward sum of DISC and the two baseline algorithms, and Table \ref{table:marbase} shows the corresponding maximum average return. Note that DISC significantly outperforms PPO and PPO-AMBER for all continuous action tasks. Note that PPO works in low-dimensional tasks such as Hopper and BipedalWalker whose action dimensions are $3$ and $4$, respectively, but it does not learn for high action-dimensional tasks such as Ant and  Humanoid whose action dimensions are $8$ and $17$, respectively. This is because the gradient of most samples vanishes for these high action-dimensional tasks, as explained in Section \ref{subsec:dimclip}. Furthermore, this vanishing gradient problem  makes it difficult to reuse old samples and hence using experience replay for PPO-AMBER does not work in high action-dimensional tasks. On the other hand, DISC effectively solves the vanishing gradient problem, as described in Section \ref{subsec:reusing}. Hence, the performance gain of DISC over PPO is significant in these high dimensional difficult tasks. PPO with large batch or distributed system also works well in Humanoid task \cite{heess2017emergence}, but it requires hyperparameter adjustment for each environment. Here, we did not adjust the hyperparameters of PPO, PPO-AMBER and DISC for each task.

	\begin{figure}
		\centering
		\includegraphics[width=0.48\textwidth]{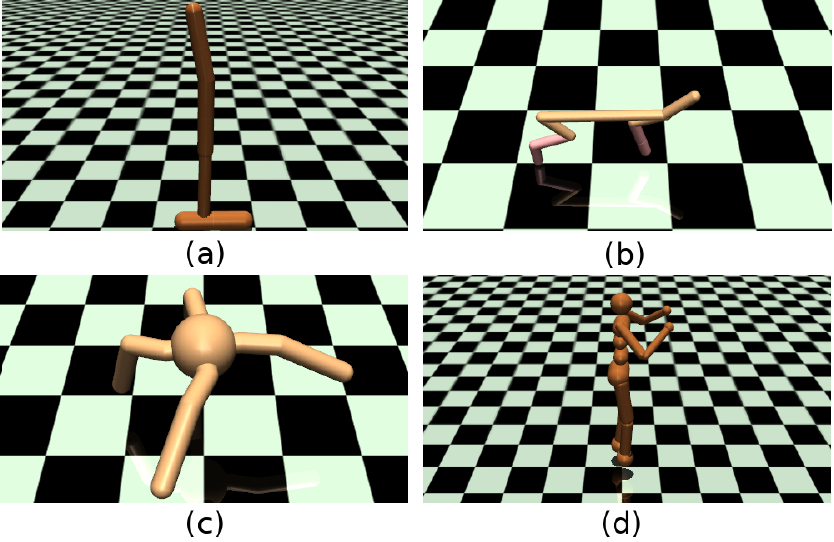}
		\caption{Examples of Open AI GYM continuous control tasks: (a) Hopper-v1 (b) HalfCheetah-v1 (c) Ant-v1 (d) Humanoid-v1}
		\label{fig:mujoco}
	\end{figure}
	
	\begin{figure*}
		\centering
		\includegraphics[width=0.23\textwidth]{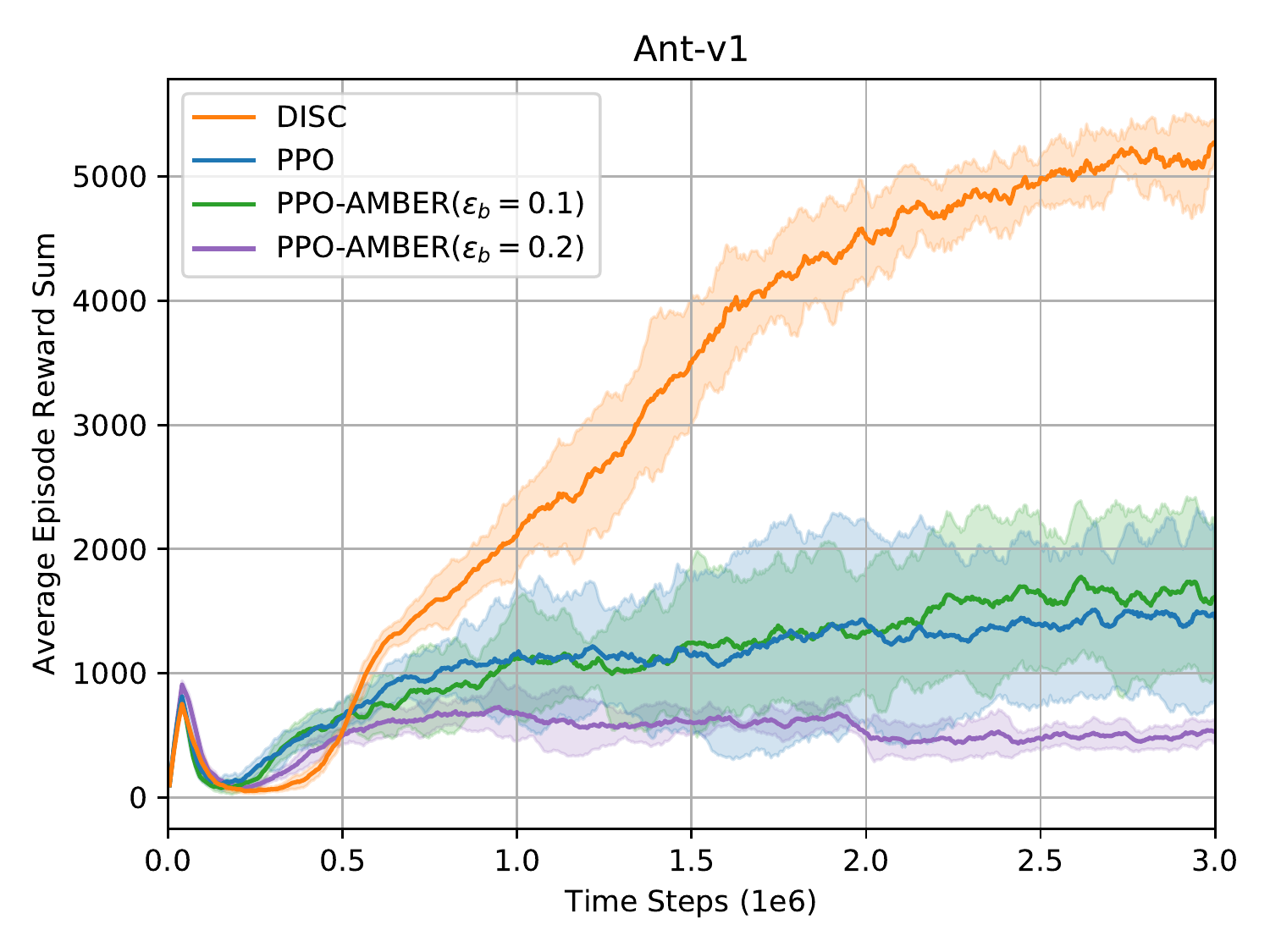}
		\includegraphics[width=0.23\textwidth]{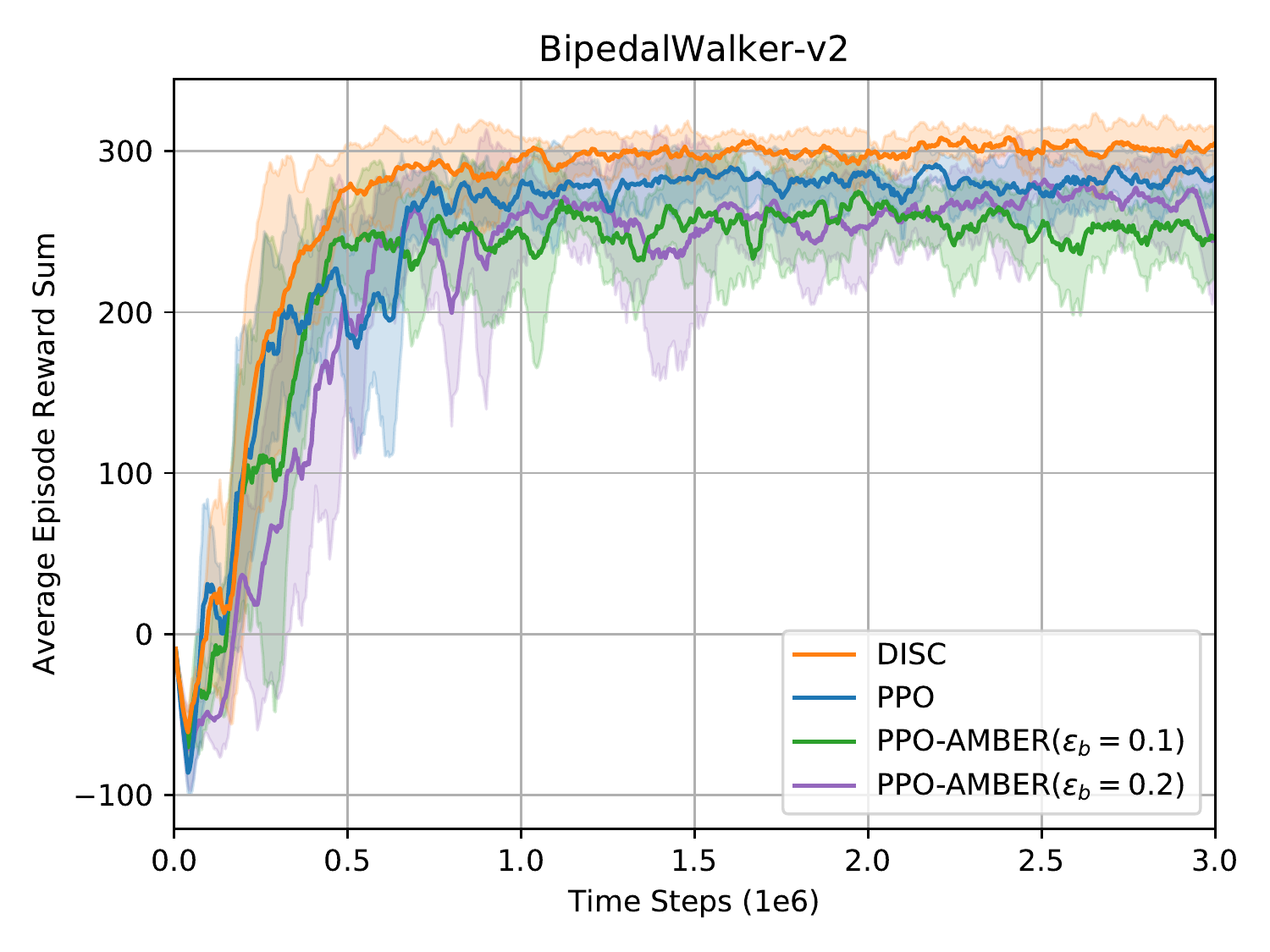}
		\includegraphics[width=0.23\textwidth]{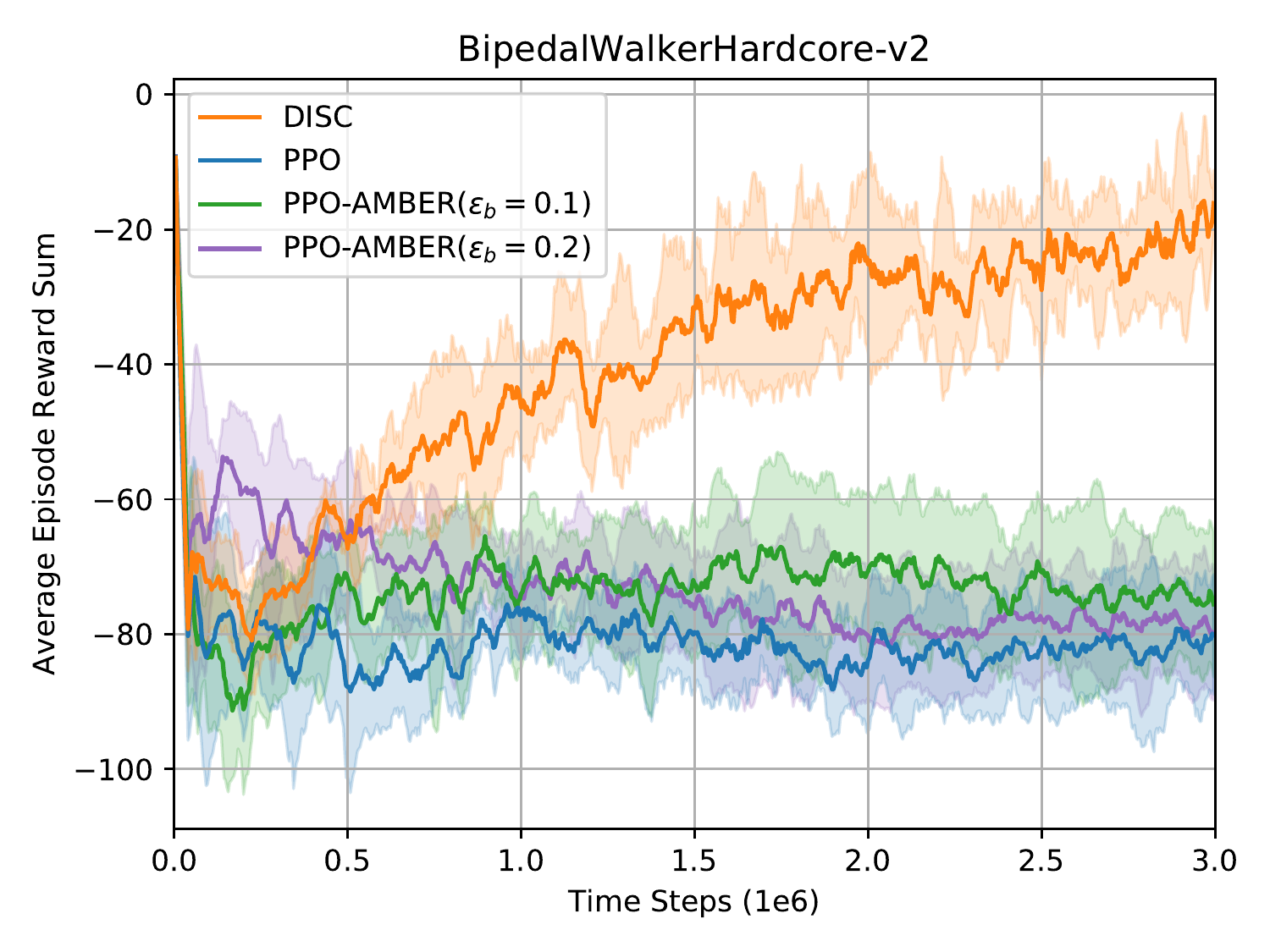}
		\includegraphics[width=0.23\textwidth]{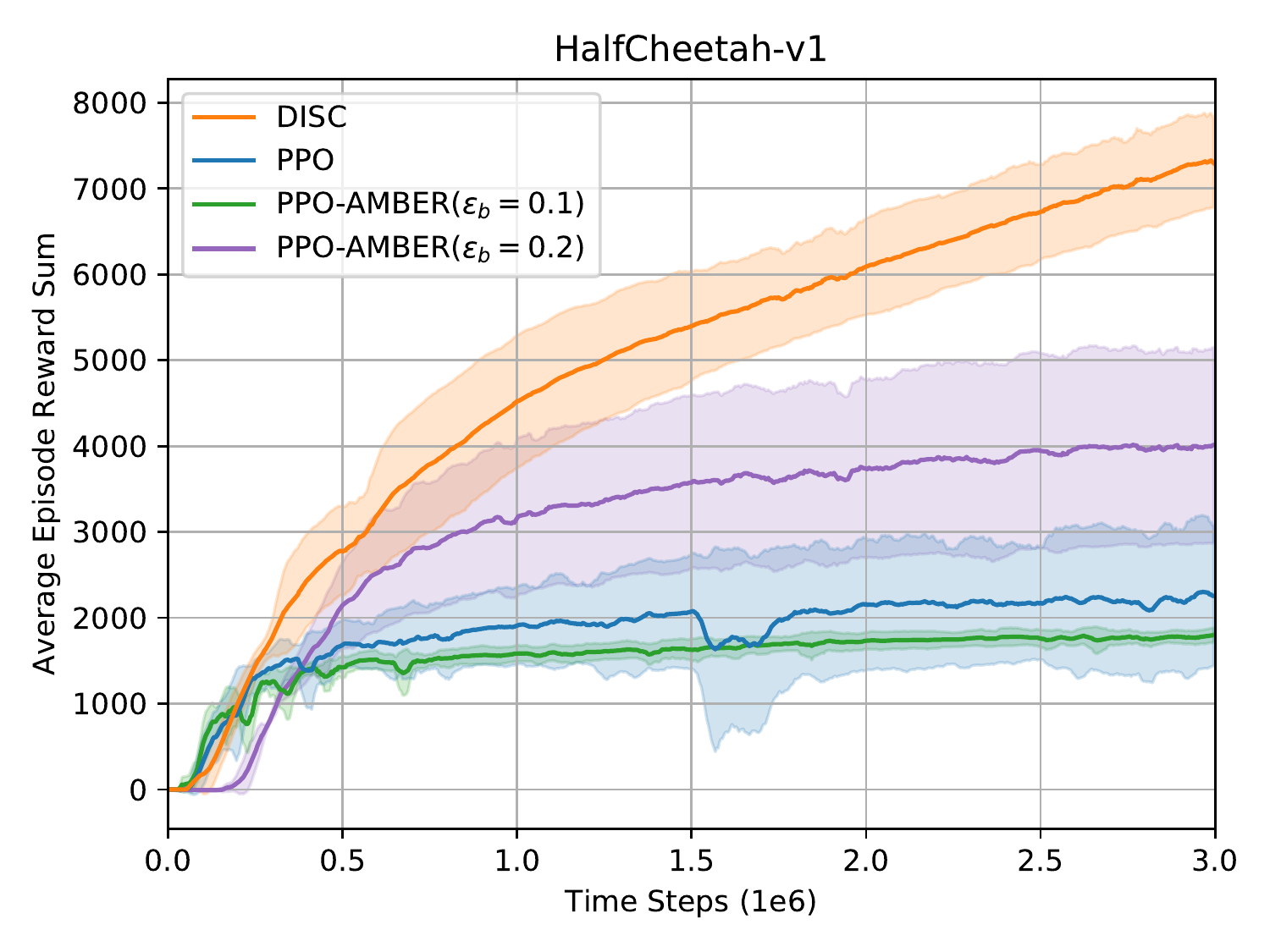}
		\includegraphics[width=0.23\textwidth]{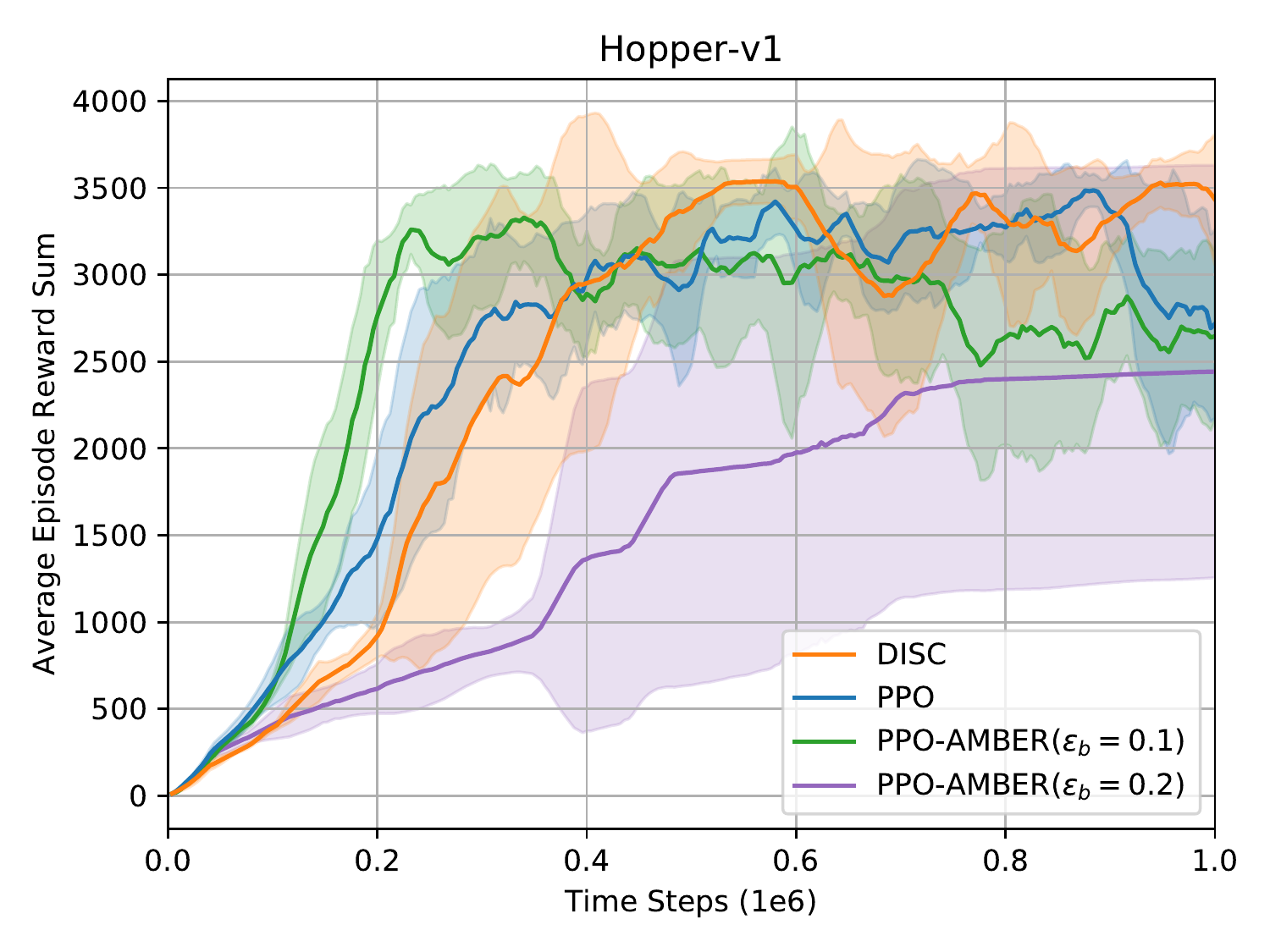}
		\includegraphics[width=0.23\textwidth]{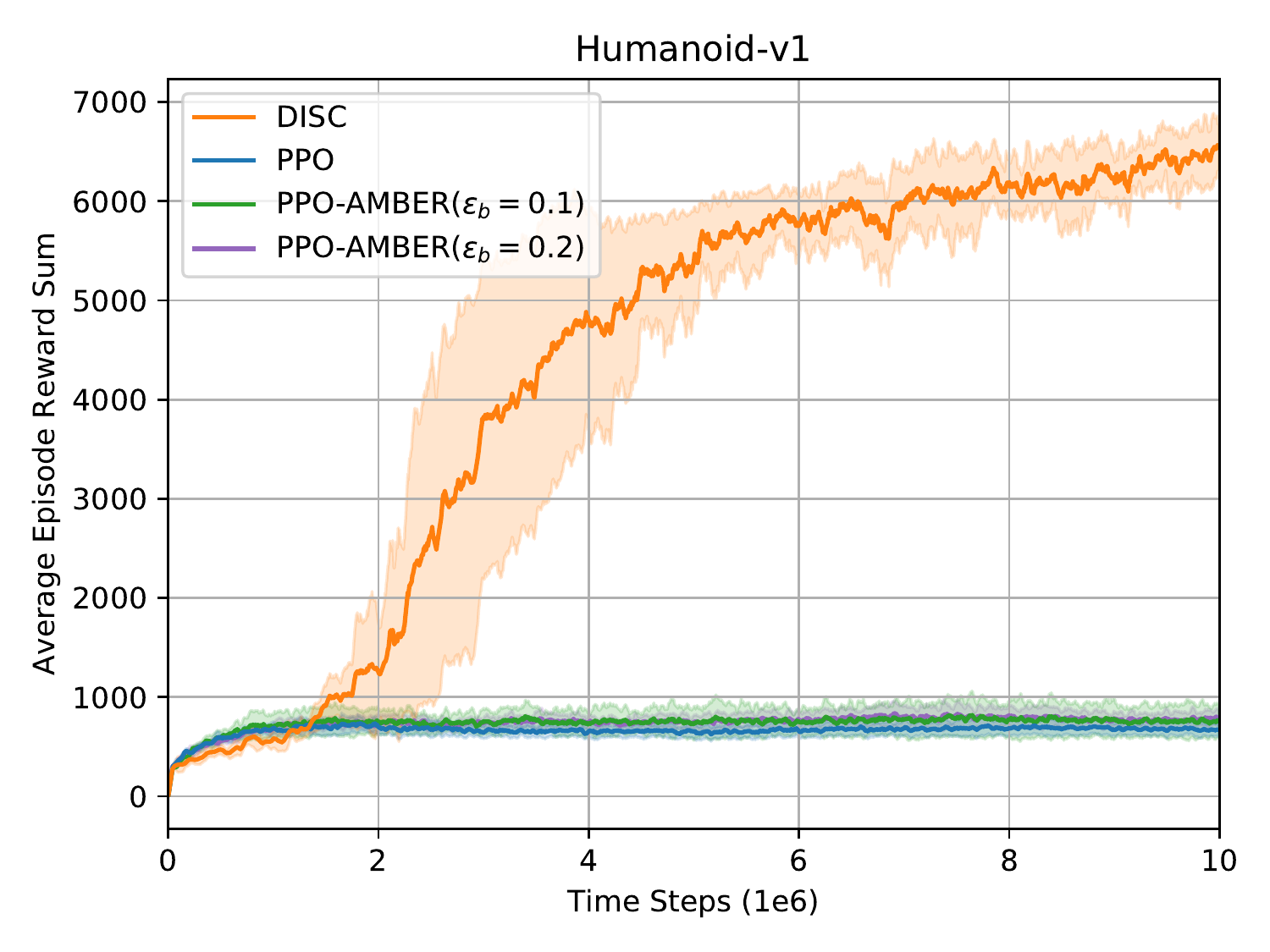}
		\includegraphics[width=0.23\textwidth]{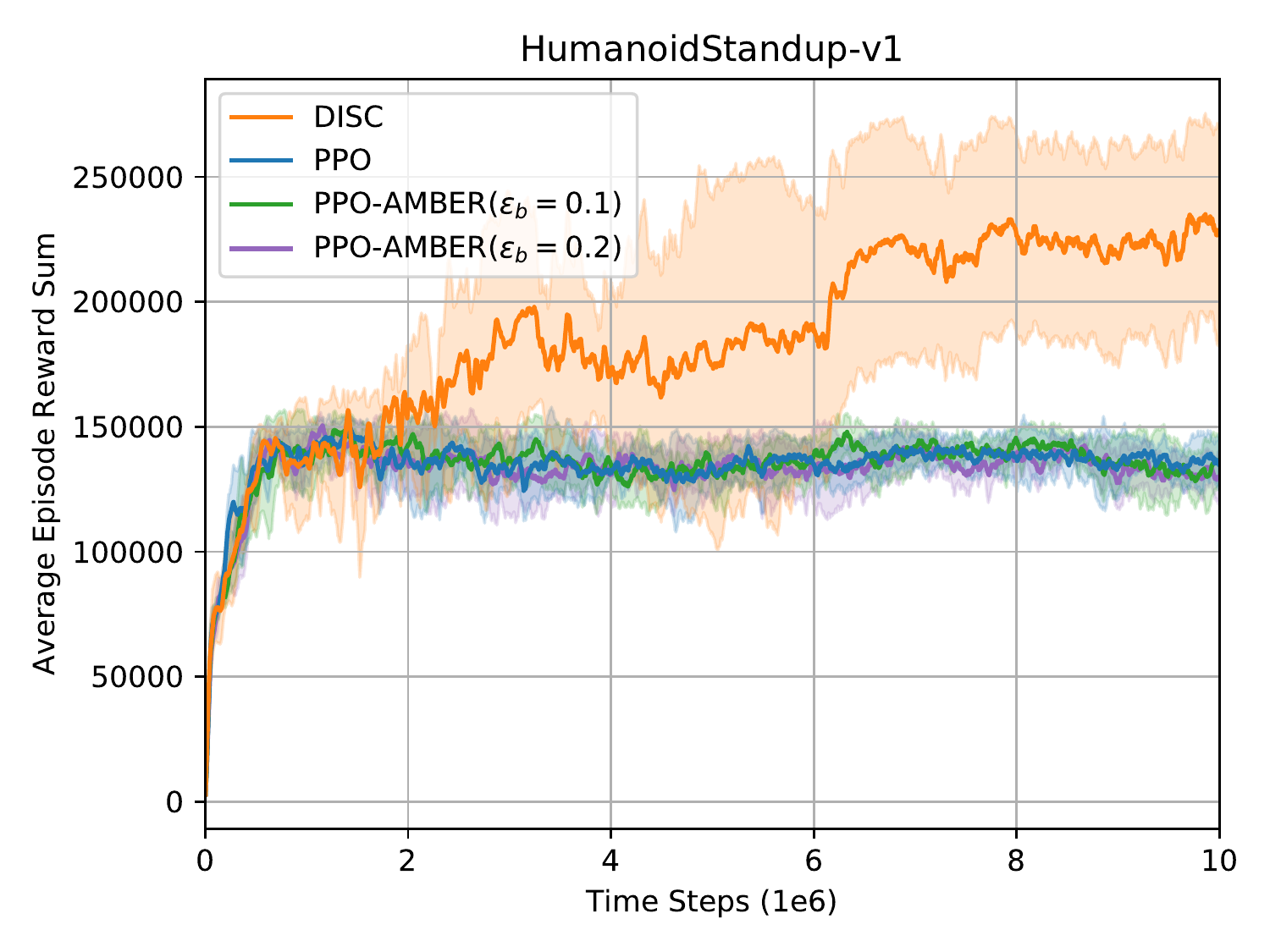}
		\includegraphics[width=0.23\textwidth]{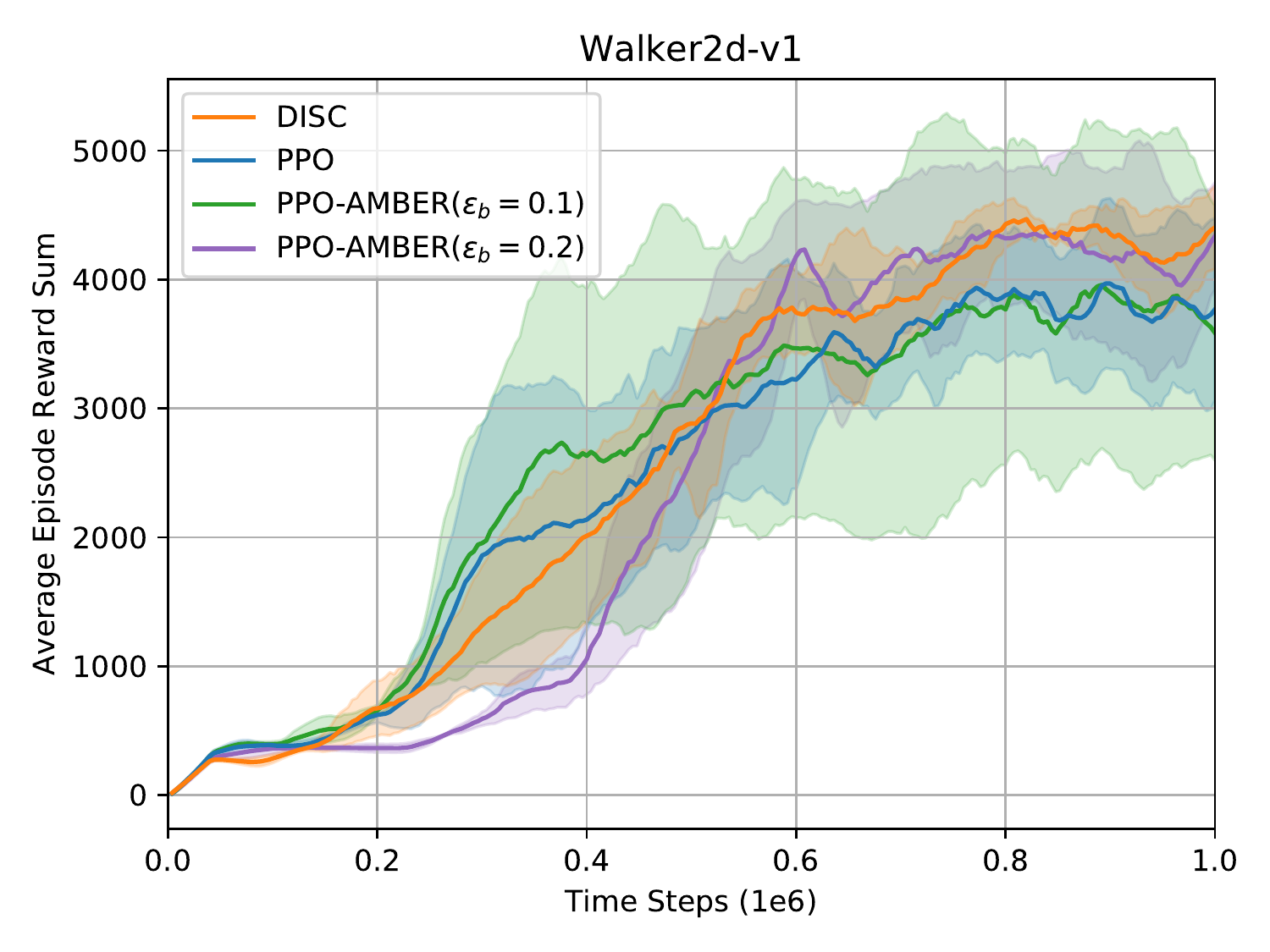}
		\caption{Performance comparison on Open AI GYM continuous control tasks}
		\label{fig:perfbase}
	\end{figure*}
	\begin{table*}[!h]
		\centering
		\begin{tabular}{l|C{12em}C{6em}C{6em}C{6em}}
			\hline
			& DISC & PPO & PPO-AMBER ($\epsilon_b=0.1$) & PPO-AMBER ($\epsilon_b=0.2$)\\
			\hline
			Ant-v1 & $\mathbf{5469.04}\pm\mathbf{283.62}$ & $1628.96$ & $1935.02$ & $831.09$ \\
			BipedalWalker-v2 & $\mathbf{314.29}\pm\mathbf{5.16}$ & $\mathbf{302.90}$ & $297.24$ & $294.20$  \\
			BipedalWalkerHardcore-v2 & $\mathbf{8.89}\pm\mathbf{25.81}$ & $-62.11$ & $-54.67$ & $-45.90$  \\
			HalfCheetah-v1 & $\mathbf{7413.89}\pm\mathbf{635.30}$ & $2342.75$ & $1814.84$ & $4073.58$ \\
			Hopper-v1 & $\mathbf{3570.40}\pm\mathbf{227.53}$  & $\mathbf{3571.22}$ & $\mathbf{3498.32}$ & $2445.57$ \\
			Huamnoid-v1 & $\mathbf{6705.12}\pm\mathbf{313.44}$ & $821.30$ & $917.44$ & $901.33$ \\
			HumanoidStandup-v1 & $\mathbf{246435.89}\pm\mathbf{40387.30}$ & $154048.51$ & $153212.20$ & $154163.24$ \\
			Walker2d-v1 & $\mathbf{4769.96}\pm\mathbf{142.25}$ & $4202.48$  & $4168.16$ & $\mathbf{4654.61}$  \\
			\hline
		\end{tabular}
		\caption{Max average return of DISC and baseline algorithms}
		\label{table:marbase}
	\end{table*}
	
	\subsection{Ablation Study}
	
\begin{figure*}
		\centering
		\begin{subfigure}[b]{0.24\textwidth}
			\includegraphics[width=\linewidth]{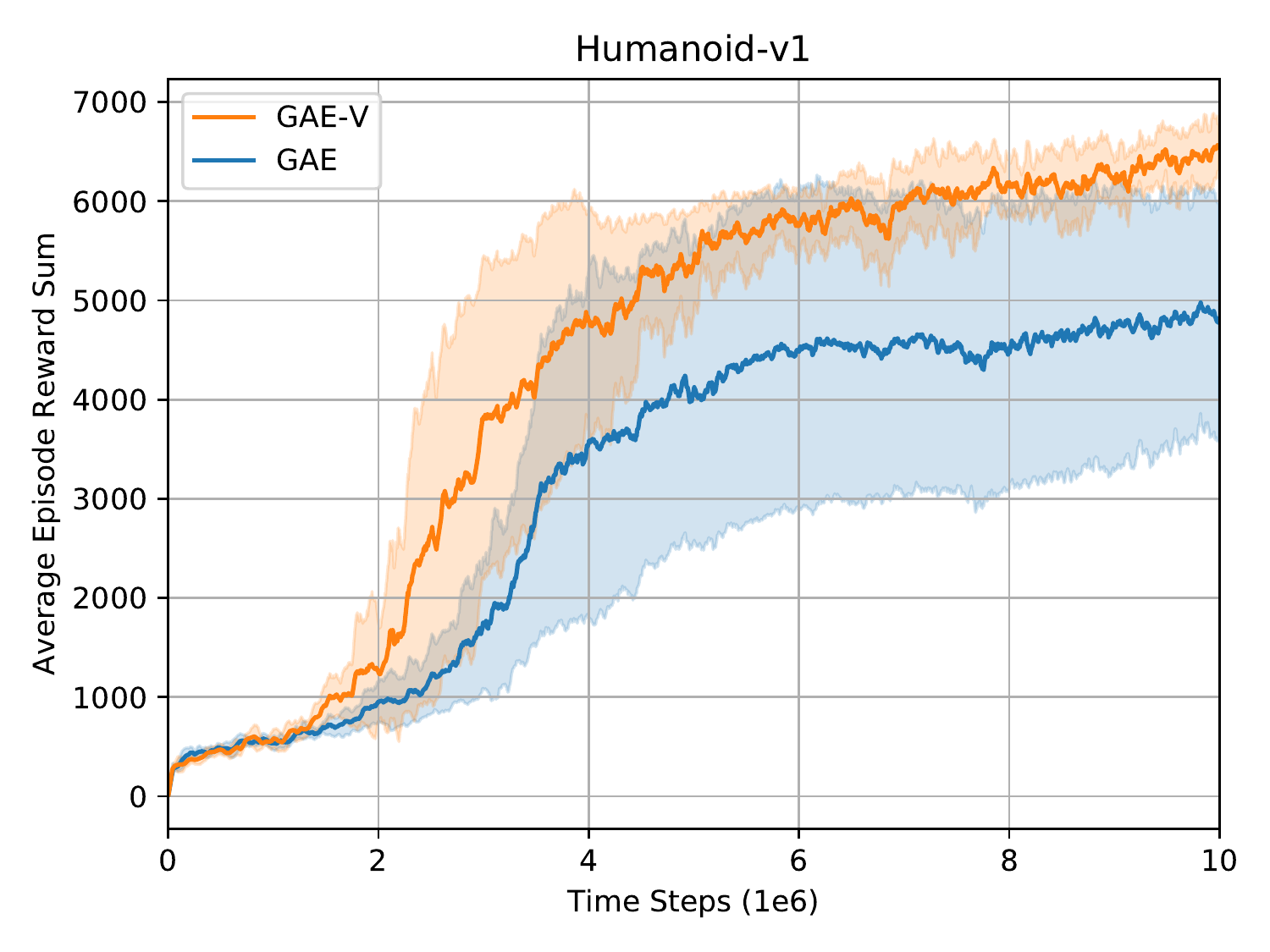}
			\caption{GAE-V vs. GAE}
			\label{fig:gaev}
		\end{subfigure}
		\begin{subfigure}[b]{0.24\textwidth}
			\centering
			\includegraphics[width=\linewidth]{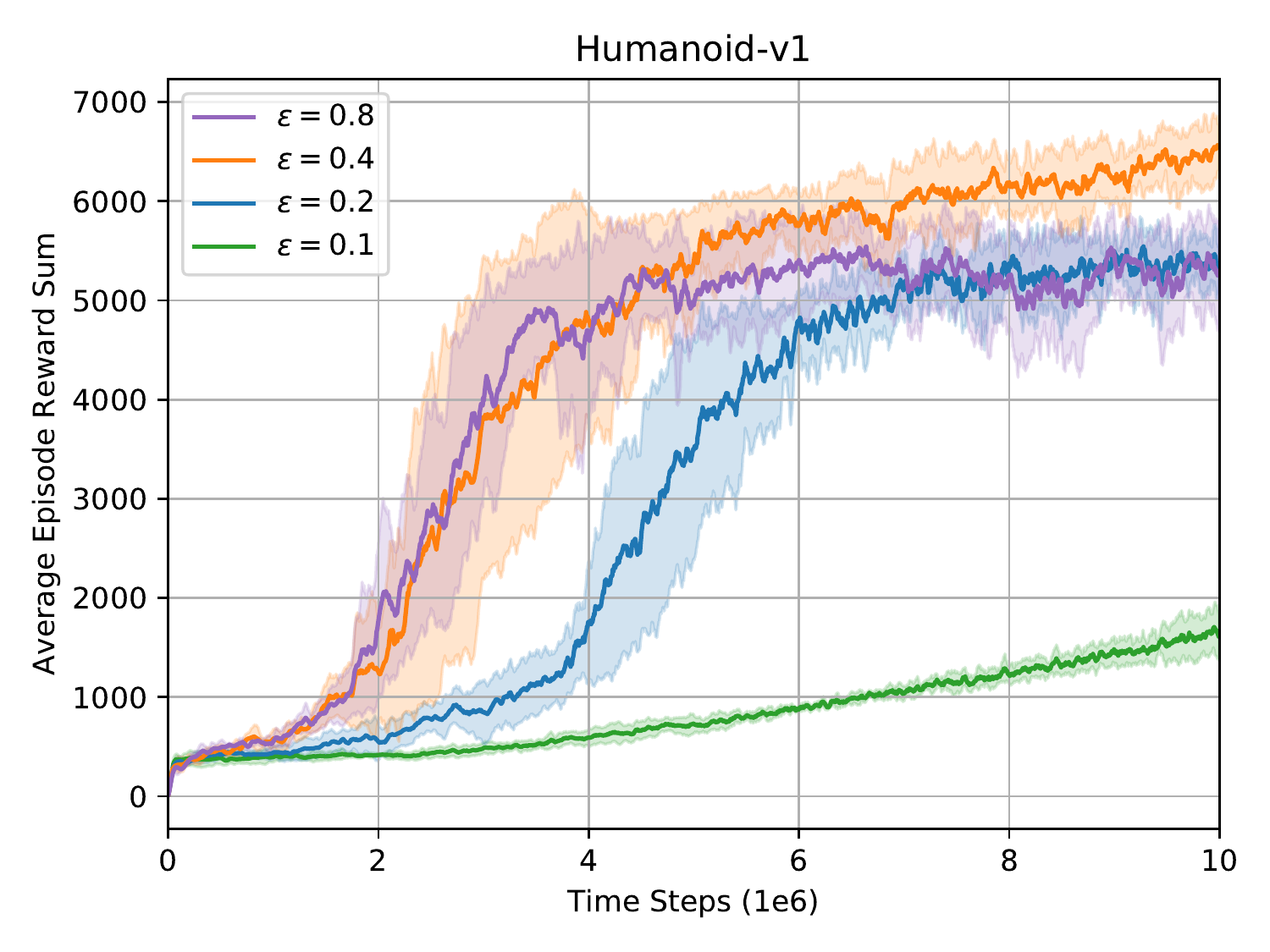}
			\caption{Clipping factor $\epsilon$}
			\label{fig:clip}
		\end{subfigure}
		\begin{subfigure}[b]{0.24\textwidth}
			\centering
			\includegraphics[width=\linewidth]{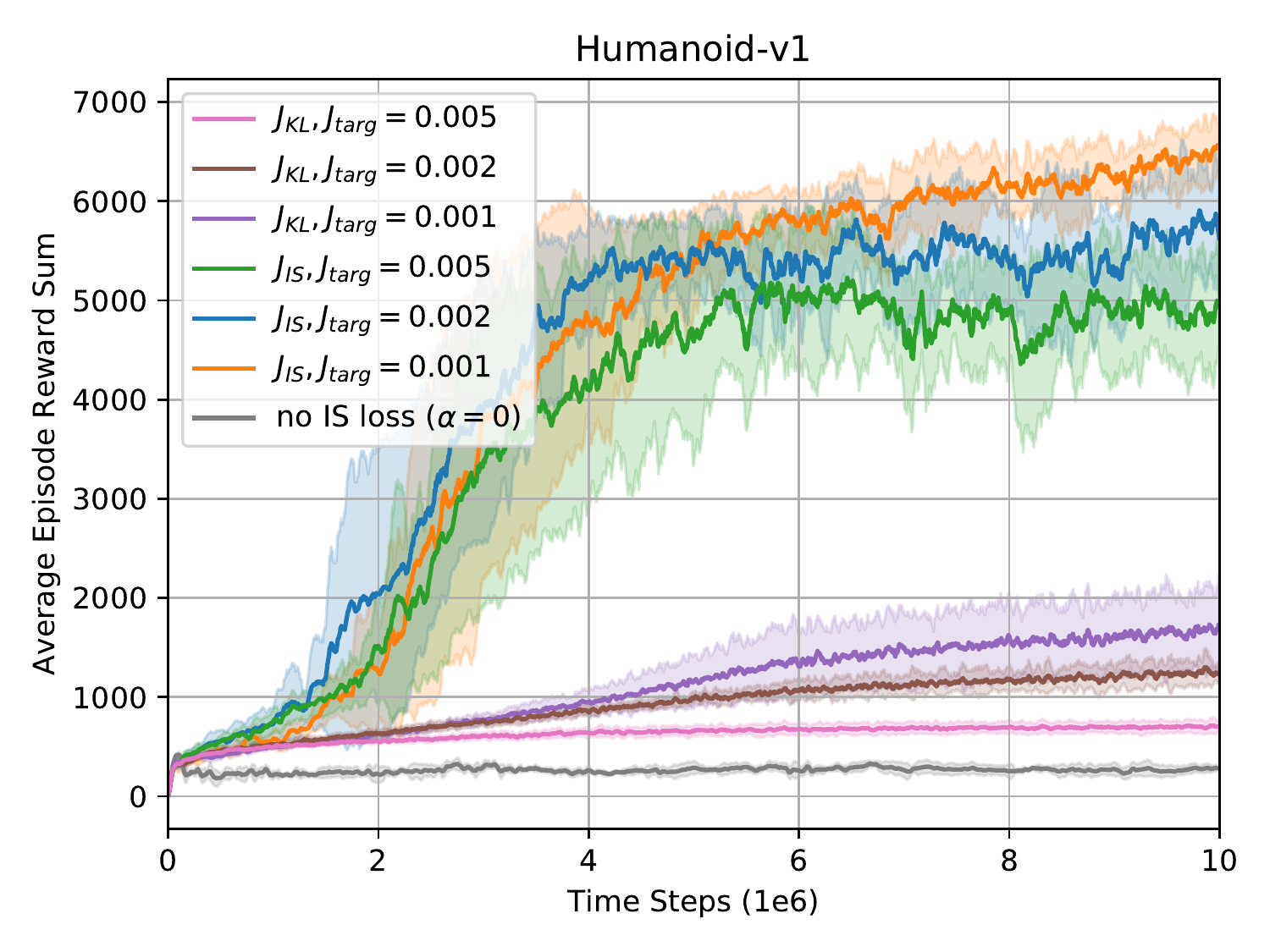}
			\caption{IS target parameter $J_{targ}$}
			\label{fig:IStarg}
		\end{subfigure}
		\begin{subfigure}[b]{0.24\textwidth}
			\centering
			\includegraphics[width=\linewidth]{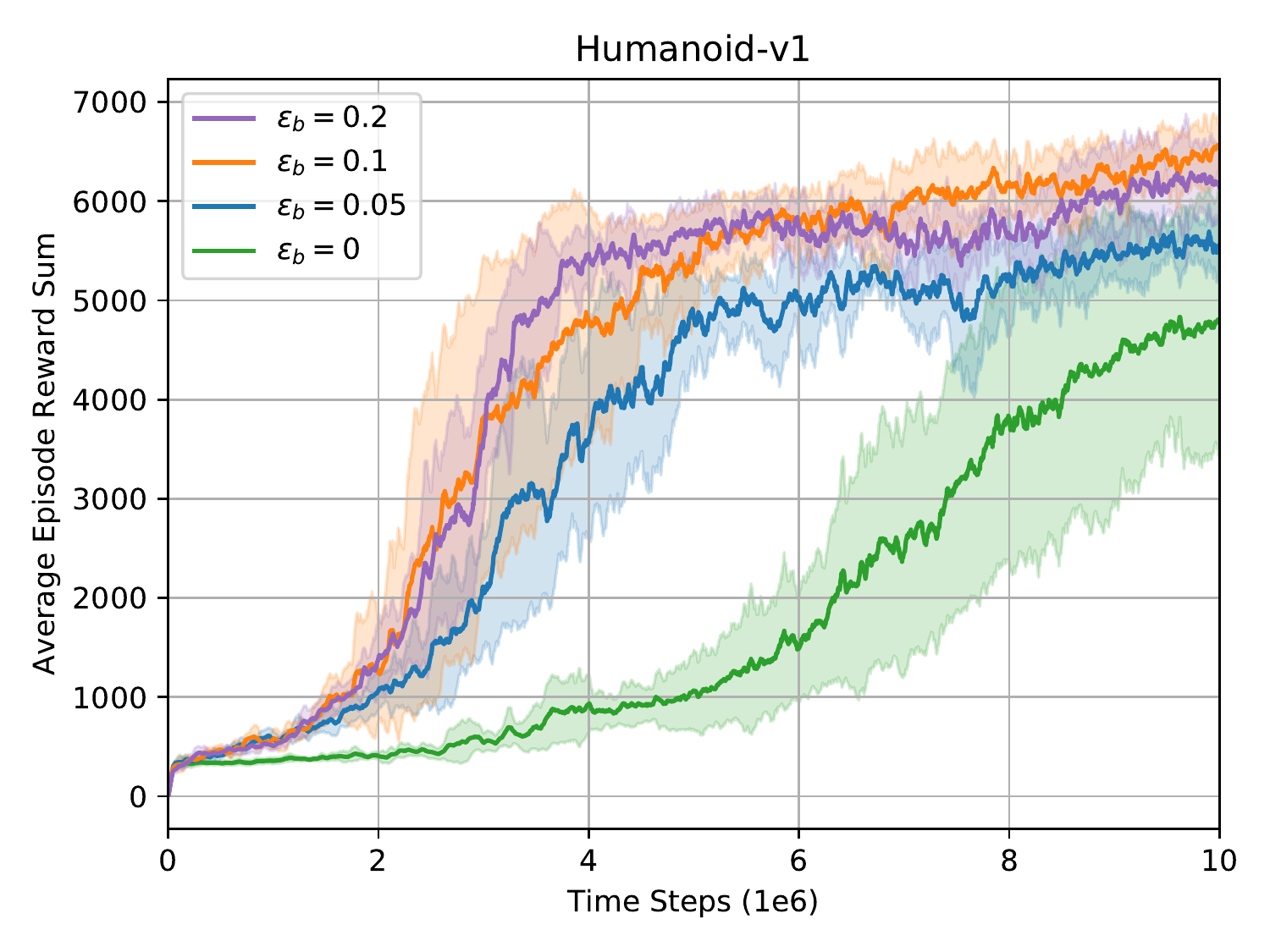}
			\caption{Batch inclusion parameter $\epsilon_b$}
			\label{fig:batchlim}
		\end{subfigure}
		
		\caption{Ablation study results on Humanoid-v1}
		\label{fig:ablation}
	\end{figure*}

	In this section, we provide in-depth ablation study on the important components and several hyper-parameters of DISC: GAE-V, the clipping factor $\epsilon$, the IS target factor $J_{targ}$ in \eqref{eq:IS}, and the batch inclusion constant $\epsilon_b$ in \eqref{eq:batchcond} based on Mujoco tasks. With this ablation study,  we investigate the impact of the key elements of DISC  and the sensitivity of the associated hyper-parameters.
	Fig.\ref{fig:ablation} shows the ablation study result on the Humanoid-v1 environment, and the ablation study results on other environments (Ant-v1, HalfCheetah-v1, Hopper-v1, and Walker2d-v1) are given in Appendix B.
	For the ablation study of each hyper-parameter, all other hyper-parameters are fixed as the values in Table A.1 in Appendix.
	The best value for each hyper-parameter is represented in orange color in the figures and is the hyper-parameter value in Table A.1 in Appendix.
	
	\textbf{GAE-V vs. GAE:} In DISC, we combine GAE with V-trace to incorporate the statistics difference of the previous sample trajectories. Fig. \ref{fig:gaev} shows the Humanoid performance of DISC with GAE and DISC with GAE-V.  It is seen that the impact of GAE-V over GAE for DISC is noticeable.
	For PPO, GAE-V reduces to GAE because it uses only the current trajectory ($\rho_t=1$).  However,  DISC uses old sample batches in the replay buffer and their statistics are different from that of the current policy. Therefore, GAE-V for advantage estimation is beneficial for the current policy evaluation from the old trajectories.

	\textbf{Clipping Factor $\epsilon$:} The clipping parameter $\epsilon$ controls clipping of the IS weight of each dimension $\rho_{t,d}$. If $\rho_{t,d}>1+\epsilon$ and $\hat{A}_t>0$ (or $\rho_{t,d}<1-\epsilon$ and $\hat{A}_t<0$), then the gradient of dimension $d$ becomes zero.   Fig. \ref{fig:clip} shows  the performance of DISC for several clipping factors: $\epsilon=0.1,~0.2,~0.4,~0.8$. When the clipping factor is too low, the number of clipped samples is too large, and this slows down learning. On the other hand, when the clipping factor is too large, it will update the policy too much with large IS weights and this causes unstable learning. Thus, DISC with a large clipping factor converges fast in the initial stage of learning, but  degrades at the final performance. Although the results for other tasks are not shown, simulations show  that the best clipping factor for DISC is around $0.4$, as seen in Fig. \ref{fig:clip}. Note that the performance of DISC is sensitive when the clipping factor is too small. This means that reducing the clipped samples is important to enhance the performance.

	\textbf{IS Loss Target Parameter $J_{targ}$:} In Fig. \ref{fig:ISdisc}, we see that the average IS weight deviation from one for each dimension $\rho'_{t,d}$ decreases as the IS target parameter $J_{targ}$ decreases.
	Note that $J_{IS}$  has a similar role as the KL divergence loss  $J_{KL}=D_{KL}(\pi_{\theta_i}||\pi_{\theta})$ in some PPO variants, which bounds  policy update from  the current policy point.
	In order to see the difference of $J_{IS}$ and $J_{KL}$, we ran DISC with $J_{IS}$ and DISC with $J_{KL}$ for several values of  $J_{targ}$, and the result is shown in Fig. \ref{fig:IStarg}. It is seen that  $J_{IS}$ is much more effective than $J_{KL}$ for Humanoid, which has high action dimensions. Although it is not shown here, we observe that  $J_{KL}$ works reasonably well for low action-dimensional environments such as HalfCheetah, or Hopper environment, as seen in Fig. B.3 in Appendix, whereas
	$J_{IS}$ works well in all environments.

\begin{figure}
		\centering
		\begin{subfigure}[b]{0.23\textwidth}
			\includegraphics[width=\linewidth]{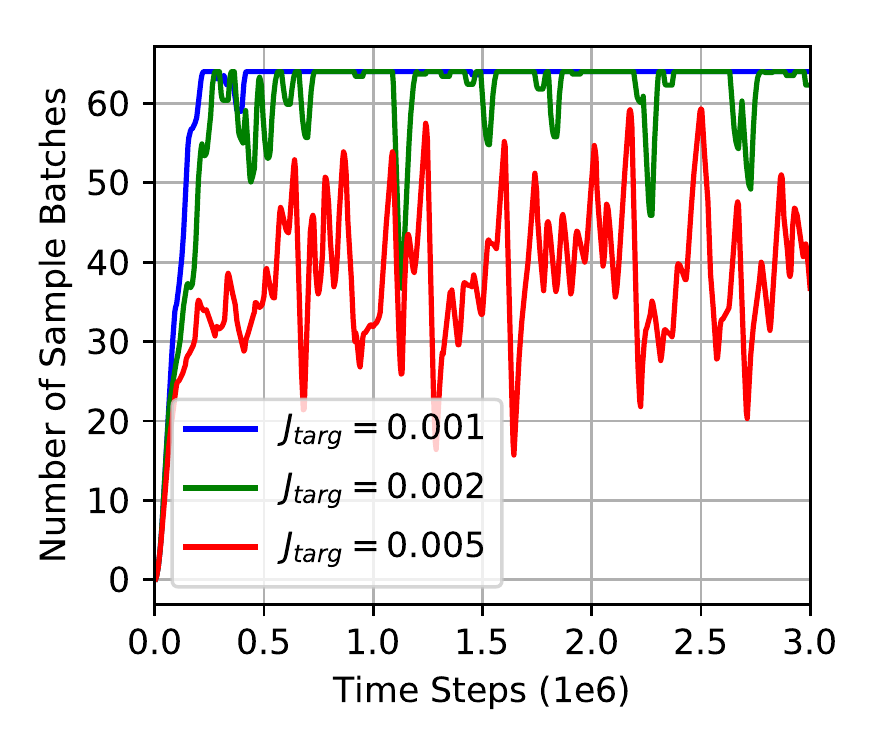}
			\caption{The num. of sample batches}
			\label{fig:numbatchIS}
		\end{subfigure}
		\begin{subfigure}[b]{0.23\textwidth}
			\includegraphics[width=\linewidth]{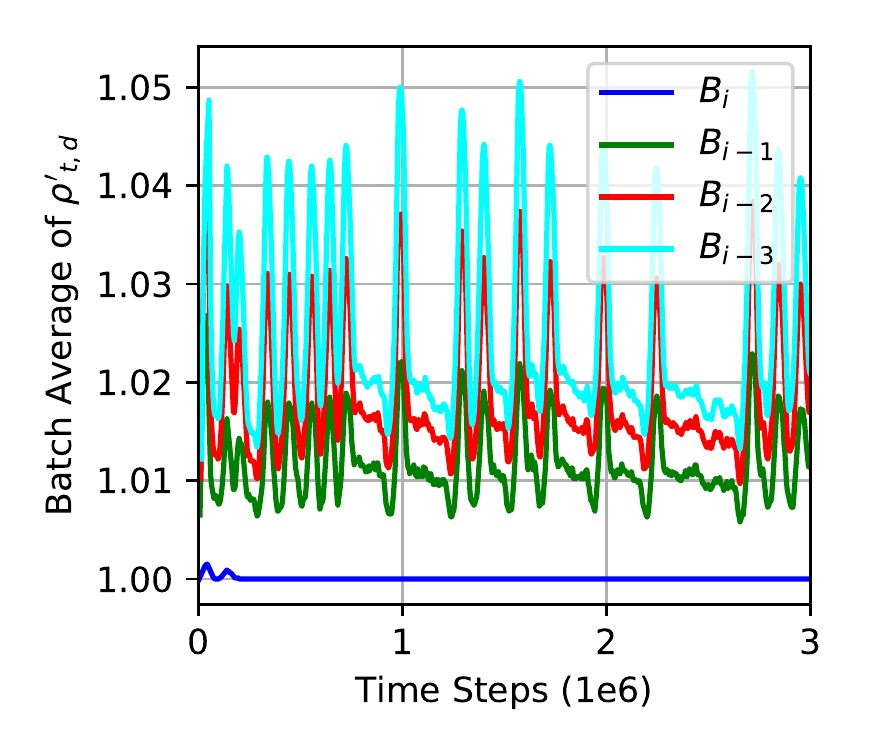}
			\caption{The average of $\rho'_{t,d}$}
			\label{fig:batchIS}
		\end{subfigure}
		\caption{(a) the number of used sample batches for different $J_{targ}$ and (b) the average of $\rho'_{t,d}$ for samples in the previous $4$ sample batches}
	\end{figure}

	Fig. \ref{fig:numbatchIS} shows the number of sample batches included for DISC with $J_{IS}$
	based on  $\eqref{eq:batchcond}$ for several values of $J_{targ}$ with a fixed $\epsilon_b$. As $J_{targ}$ increases,  the number of  included sample batches decreases and  the performance degrades, as seen in Fig \ref{fig:IStarg}.
	Large $J_{targ}$ means large  policy update step and hence the speed of learning is faster at the beginning of learning. However, it causes larger IS weights and it reduces the number of sample batches satisfying \eqref{eq:batchcond}.  As result, it reduces the sample efficiency of DISC and degrades the performance. Hence,  small $J_{targ}$ is preferred for DISC.
	
	\textbf{Batch Inclusion Parameter $\epsilon_b$:} In order to reduce bias for DISC,  we only include the old sample batches whose average $\rho'_{t,d}$ is less than $1+\epsilon_b$, as seen in \eqref{eq:batchcond}. Here, $\epsilon_b$ is chosen to be smaller  than the clipping factor $\epsilon$.

	Fig. \ref{fig:batchIS} shows that the average of $\rho'_{t,d}$ of the latest $4$ sample batches: $B_i,~B_{i-1},~B_{i-2},~B_{i-3}$ for each iteration $i$ for $\epsilon_b=0.1$, $\epsilon=0.4$, and $J_{targ}=0.001$. As expected, the average $\rho'_{t,d}$ is larger for older sample batches since the statistics of older samples deviate more from the current sample static.

	Fig. \ref{fig:batchlim} shows the performance of DISC for several values of the batch inclusion parameter:  $\epsilon_b=0,~0.05,~0.1,~0.2$, where $\epsilon_b=0$ means that we  use only the current sample batch $B_i$ for  policy update.
	It is seen that small $\epsilon_b<0.1$ degrades the performance because DISC does not reuse old sample batches much and this  reduces sample efficiency. Especially, DISC with $\epsilon_b=0$ learns much slowly and this shows the importance of reusing old sample batches.
	On the other hand, DISC with too large $\epsilon_b$ also degrades the performance because it uses too old sample batches that induce bias from clipping. So, there exists a soft spot for $\epsilon_b$. We observe that $\epsilon_b=0.1$ works well for most tasks.

	\textbf{Consistency :} From the above ablation studies, we suggest the best hyper-parameter setup for DISC in Table A.1 in Appendix.
	Fortunately, the  best parameter setup for DISC does not depend much on the environment. DISC has the consistent roughly best hyper-parameter setup for most environments, as seen in Fig. B.4 in Appendix, while
	some of the recent state-of-the-art RL algorithms require a different hyper-parameter setup for each environment.

	\begin{table*}[!h]
		\centering
		\begin{adjustbox}{width=1\textwidth}
			\small
			\begin{tabular}{L{5.1em}|C{5.1em}C{4.9em}C{4.9em}C{4.9em}C{4.9em}C{4.9em}C{4.9em}C{4.9em}C{4.9em}}
				\hline
				& DISC & DDPG & TRPO & ACKTR & Trust-PCL & SQL & TD3 & SAC \\
				\hline
				Ant & $\mathbf{5469.04}$ & $-6.87$ & $1562.98$ & $3015.22$ & $\mathbf{5482.45}$ & $2802.18$ & $\mathbf{5508.08}$ & $\mathbf{5671.21}$ \\
				HalfCheetah & $7413.89$ & $4020.33$ & $2394.03$ & $3678.57$ & $5597.58$ & $6673.42$ & $11244.30$ & $\mathbf{14817.63}$ \\
				Hopper & $\mathbf{3570.40}$ & $729.23$ & $2662.36$ & $3004.15$ & $3073.03$ & $2432.42$ & $2942.88$ & $3322.59$ \\
				Huamnoid & $\mathbf{6705.12}$ & $857.98$ & $1420.34$ & $4814.80$ & $138.46$ & $5010.72$ & $63.33$ & $\mathbf{6883.53}$ \\
				HumanoidS & $\mathbf{246435.89}$ & $142220.05$ & $147258.61$ & $109655.30$ & $79492.38$ & $138996.84$ & $58693.87$ & $139513.04$ \\
				Walker2d & $\mathbf{4769.96}$ & $810.93$ & $2468.22$ & $2350.81$ & $2226.43$ & $2592.78$ & $\mathbf{4633.84}$ & $3884.05$ \\
				\hline
			\end{tabular}
		\end{adjustbox}
		\caption{Max average return of DISC and other state-of-art continuous RL algorithms}
		\label{table:otherrl}
	\end{table*}

	\subsection{Comparison to Other Continuous RL Algorithms}

	In this subsection, we provide the performance comparison between  DISC and  other state-of-the-art  on-policy and off-policy RL algorithms. We compare DISC with the following state-of-the-art RL algorithms: Deep Deterministic Policy Gradient (DDPG) \cite{lillicrap2015continuous} which applies deep Q-learning to the deterministic policy gradient, Trust Region Policy Optimization (TRPO) \cite{schulman2015trust} which uses a KL divergence constraint to stably update the policy in the trust region, Actor Critic using Kronecker-factored Trust Region (ACKTR) \cite{wu2017scalable} which optimizes both actor and critic by using Kronecker-factored curvature to approximate natural gradient, Trust region with Path Consistency Learning (Trust-PCL) \cite{nachum2017trust} which uses off-policy data in trust region methods by using path consistency learning, Soft Q-Learning (SQL) \cite{haarnoja2017rein} which optimizes energy-based policy by using stein variational gradient descent, Twin Delayed Deep Deterministic Policy Gradient (TD3) \cite{fujimoto2018addressing} which reduces Q-function approximation error based on clipped double Q-learning for continuous control, and Soft Actor Critic (SAC) \cite{haarnoja2018soft} which  applies  energy-based soft policy iteration and evaluation to off-policy actor critic methods.

	TRPO and ACKTR are on-policy RL like PPO, and other algorithms are off-policy RL. For DDPG, TRPO, and ACKTR, we implemented the algorithms by following  the OpenAI baselines \cite{baselines}. For other algorithms, we use the author-provided implementations in Github. Table \ref{table:otherrl} shows the maximum average return of all the considered algorithms  on Mujoco tasks\footnote{We skipped  Box2D simulation since the parameter setup is not given for some RL algorithms. HumaniodS is the abbreviation of the HumanoidStandup environment.}, and the results show that DISC has the top-level performance in five tasks out of the six considered tasks. Especially, DISC achieves the highest performance in HumanoidStandup with 17 action dimensions so that other state-of-the-art RL algorithms cannot catch up. Indeed, it is seen that  DISC is a stable and good algorithm for high action-dimensional tasks. Detailed learning curves are given in Fig. C.1 in Appendix.

	\section{Conclusion}
	\label{sec:conclusion}
	
	In this paper, we have proposed a new continuous action control algorithm, DISC, by applying  dimension-wise IS weight clipping together with GAE-V for advantage estimation. The proposed DISC algorithm separately clips the IS weight of each action dimension to solve the vanishing gradient problem, and adaptively controls the IS weight to bound policy update from the current policy for stable learning. The proposed method enables efficient learning for high action-dimensional tasks and reusing of old samples like in off-policy learning to increase the sample efficiency.
	Numerical results show that the proposed new algorithm outperforms PPO and other RL algorithms in various Open AI Gym tasks, and the performance gain by DISC is significant for high action-dimensional tasks, which have been considered difficult tasks for conventional learning. Further works may include the bias/variance analysis of DISC,  proof of the convergence of the DISC objective function, and  analysis of the IS weight loss $J_{IS}$.
	
	\newpage
	\section*{Acknowledgements}
	
	This research was supported by Basic Science Research Program through the National Research Foundation of Korea (NRF) funded by the Ministry of Science, ICT \& Future Planning(NRF-2017R1E1A1A03070788).
	
	\nocite{langley00}
	
	\bibliography{referenceBibs}
	\bibliographystyle{icml2019}
	
	\newpage
	\onecolumn
	\appendix
	\counterwithin{table}{section}
	\counterwithin{figure}{section}
	\section{Hyperparameter Setting and Environment Description}
	The hyperparameters  for simple PPO, PPO-AMBER, and DISC are summarized in Table \ref{table:parameter}, and the dimensions of  state and action spaces  of Open AI GYM tasks are given in Table \ref{table:env}.
	
	\begin{table}[!h]
		\centering
		\begin{tabular}{l|C{9em}C{9em}C{9em}}
			\hline
			& PPO & PPO-AMBER & DISC \\
			\hline
			Clipping factor $(\epsilon)$ & 0.2 & 0.2 & 0.4 \\
			Horizon $(N)$ & 2048 & 2048 & 2048 \\
			Discount factor $(\gamma)$ & 0.99 & 0.99 & 0.99 \\
			TD parameter $(\lambda)$ & 0.95 & 0.95 & 0.95 \\
			Epoch & 10 & 10 & 10 \\
			Gradient steps per epoch & 32 & 32 & 32 \\
			Mini-batch size $(M)$ & 64 & \textrm{variable} & \textrm{variable} \\
			Optimizer & Adam & Adam & Adam \\
			Learning rate $(\beta)$ & \multicolumn{3}{c}{$\max(0.0001,~\textrm{Anneal}(0.0003,~0))$}\\
			Policy distribution & \multicolumn{3}{c}{Independent Gaussian distribution} \\
			Policy and value network & \multicolumn{3}{c}{Feed forward network with $2$ hidden layers of size $64$ and $\tanh(\cdot)$ activation}\\
			Batch inclusion parameter $(\epsilon_b)$ & $\cdot$ & 0.1, 0.2 & 0.1 \\
			Replay length $(L)$ & $\cdot$ & 64 & 64 \\
			IS target parameter $(J_{targ})$ & $\cdot$ & $\cdot$ & 0.0001 \\
			Initial IS weighting factor $(\alpha_{IS})$  & $\cdot$ & $\cdot$ & 1 \\
			\hline
		\end{tabular}
		\caption{Hyper-parameter setting of PPO, PPO-AMBER, and DISC}
		\label{table:parameter}
	\end{table}
	
	\begin{table}[!h]
		\centering
		\begin{tabular}{l|C{6em}C{6em}}
			\hline
			\textbf{Mujoco} & State dim. & Action dim.  \\
			\hline
			Ant-v1 & 111 & 8 \\
			HalfCheetah-v1 & 17 & 6 \\
			Hopper-v1 & 11 & 3  \\
			Humanoid-v1 & 376 & 17 \\
			HumanoidStandup-v1 & 376 & 17 \\
			Walker2d-v1 & 17 & 6 \\
			\hline
			\textbf{Box2d} & State dim. & Action dim.  \\
			\hline
			BipedalWalker-v2 & 24 & 4 \\
			BipedalWalkerHardcore-v2 & 24 & 4 \\
			\hline
		\end{tabular}
		\caption{The dimensions of state and action spaces of OpenAI GYM continuous control tasks}
		\label{table:env}
	\end{table}

	\newpage
	
	\section{Ablation Study on More Tasks}
	
	\begin{figure}[!h]
		\centering
		\includegraphics[width=0.24\textwidth]{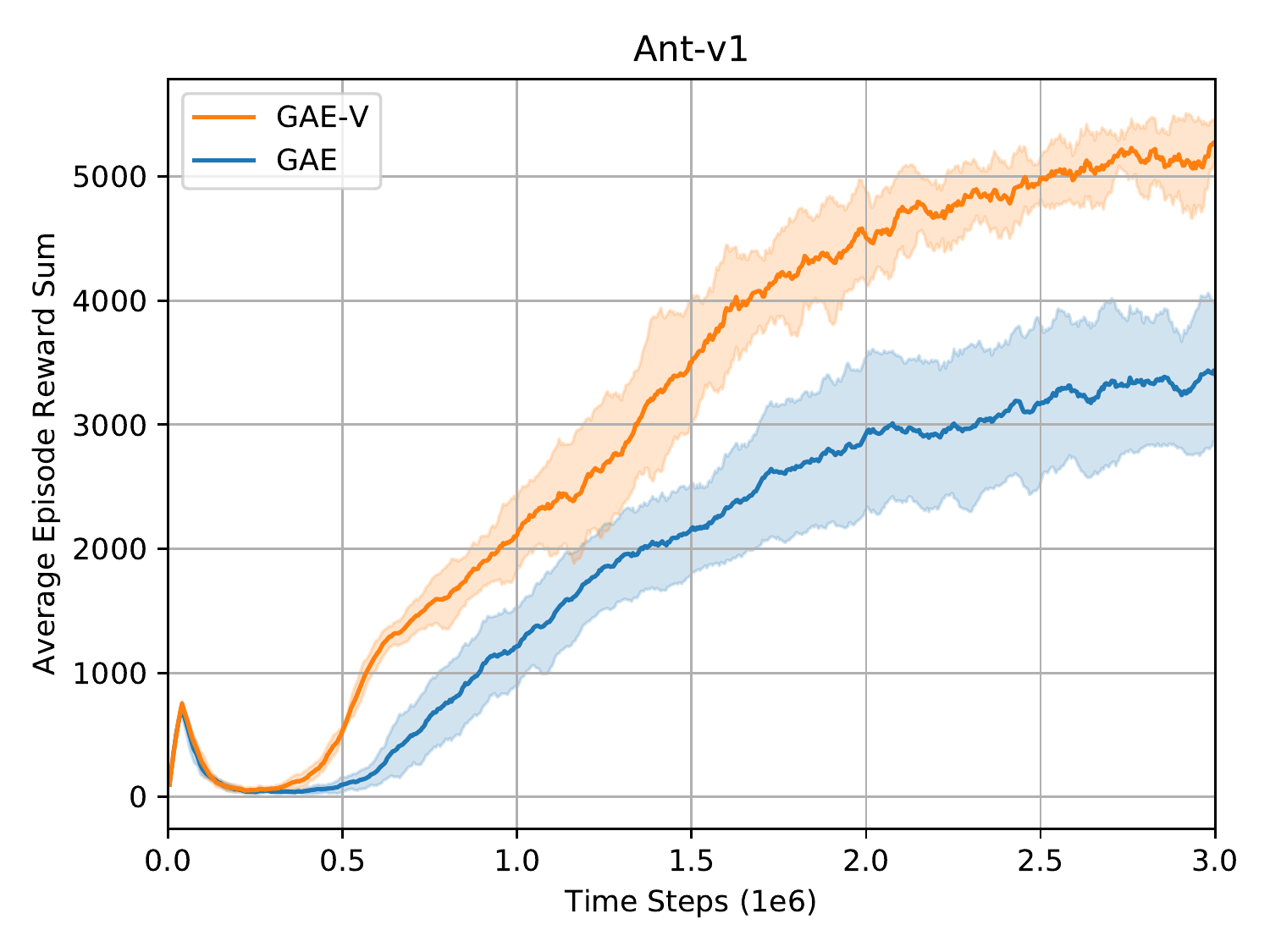}
		\includegraphics[width=0.24\textwidth]{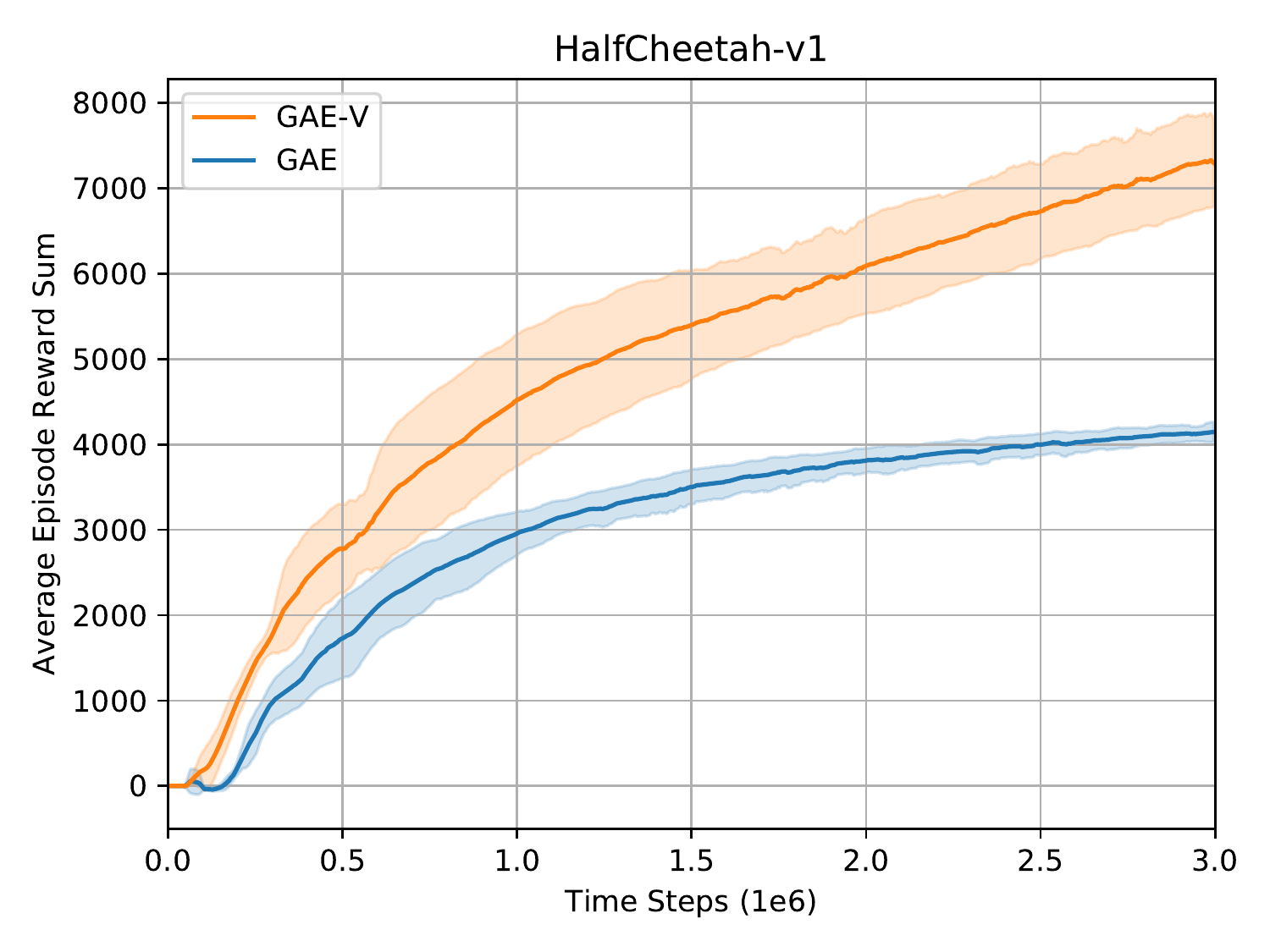}
		\includegraphics[width=0.24\textwidth]{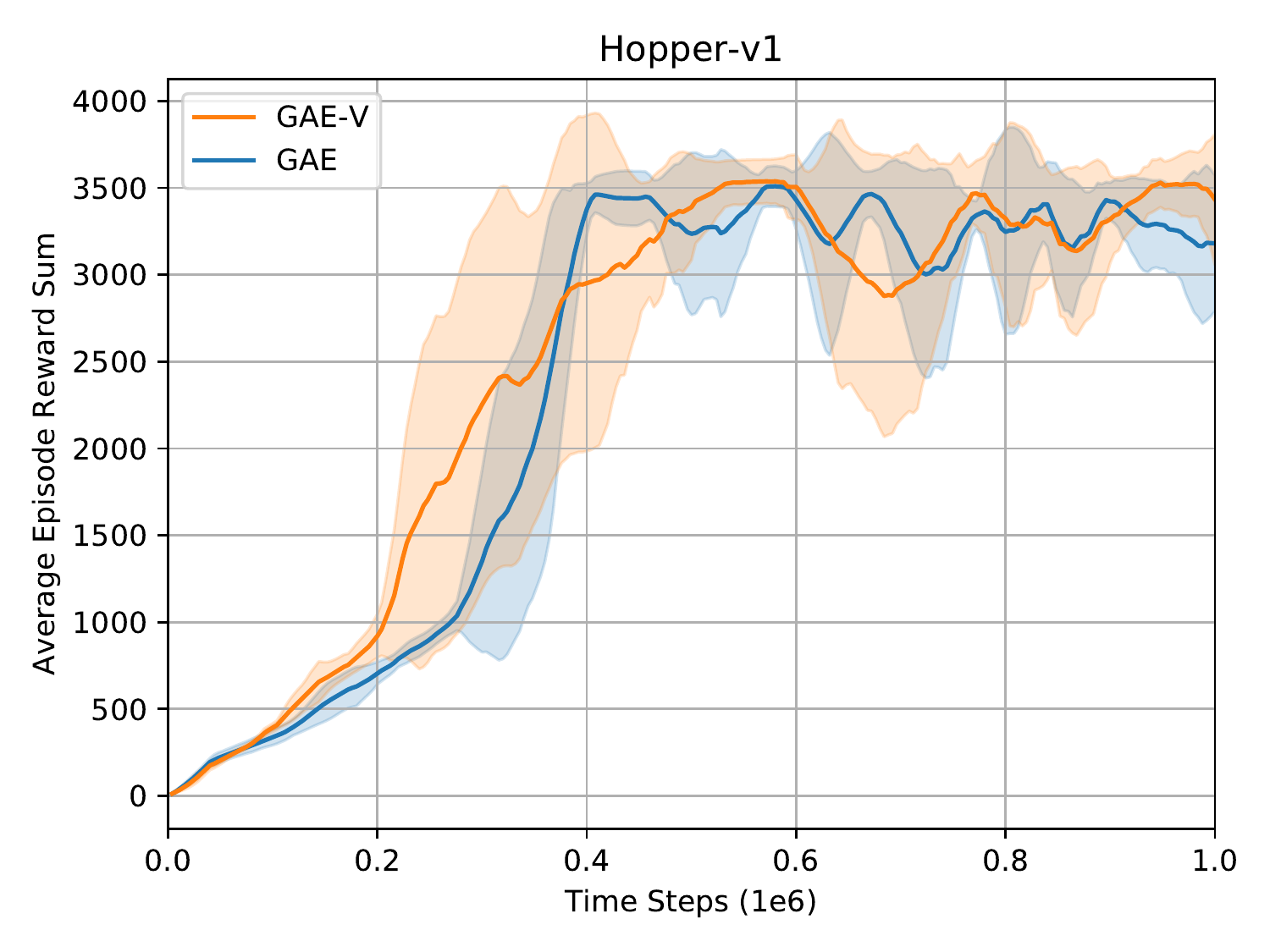}
		\includegraphics[width=0.24\textwidth]{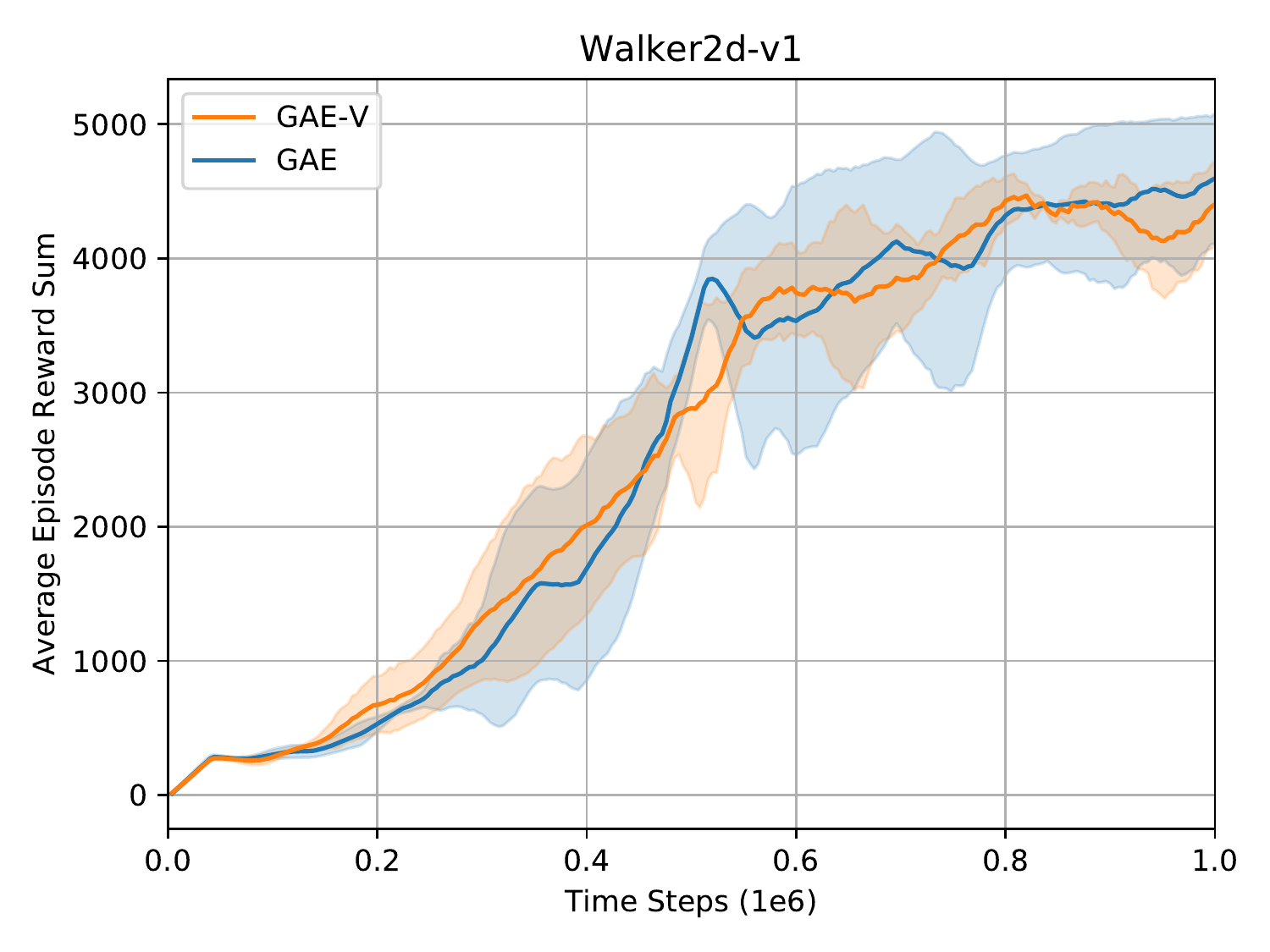}
		\caption{GAE-V versus GAE}
		\label{fig:ablgaev}
		\centering
		\includegraphics[width=0.24\textwidth]{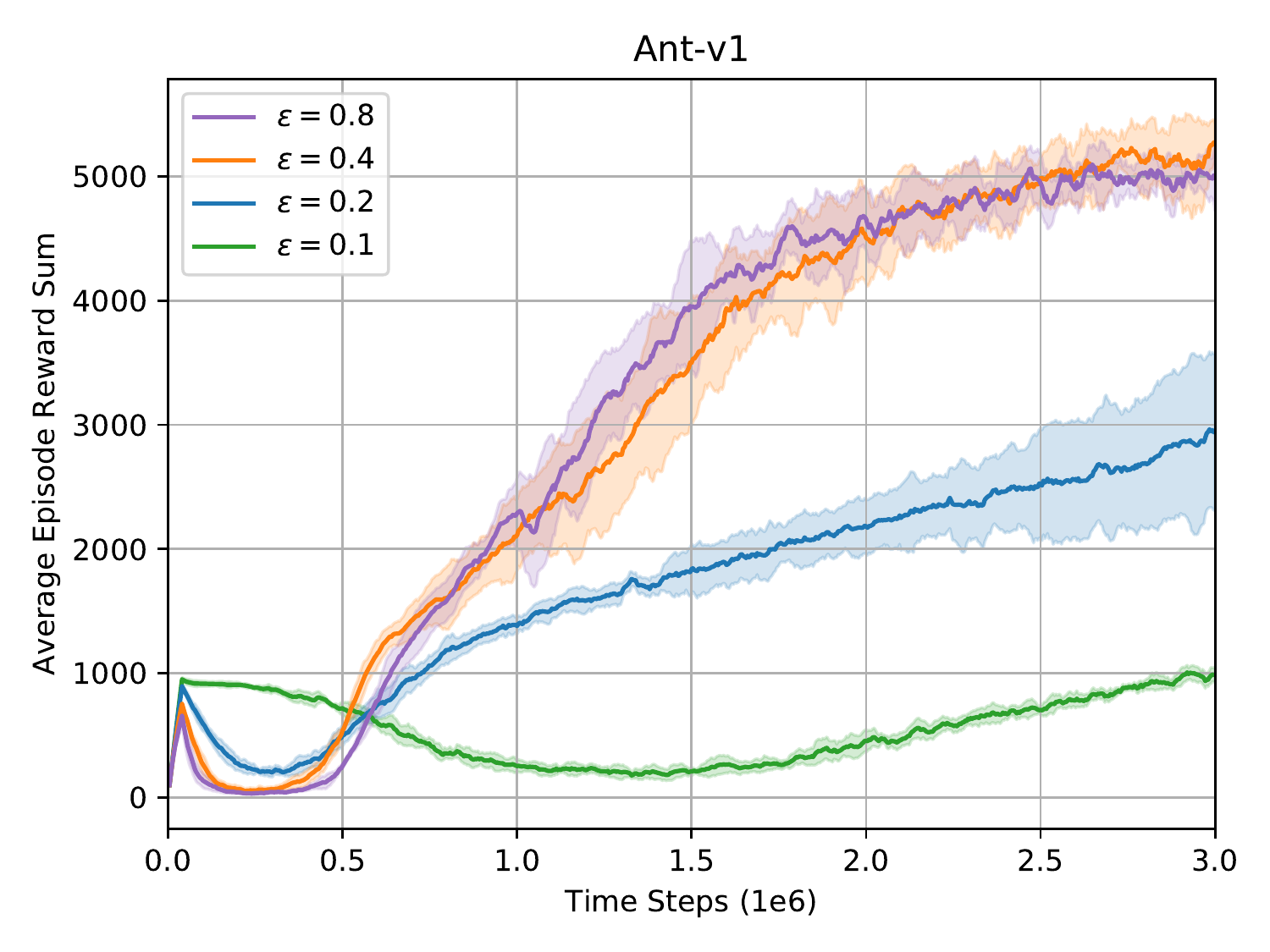}
		\includegraphics[width=0.24\textwidth]{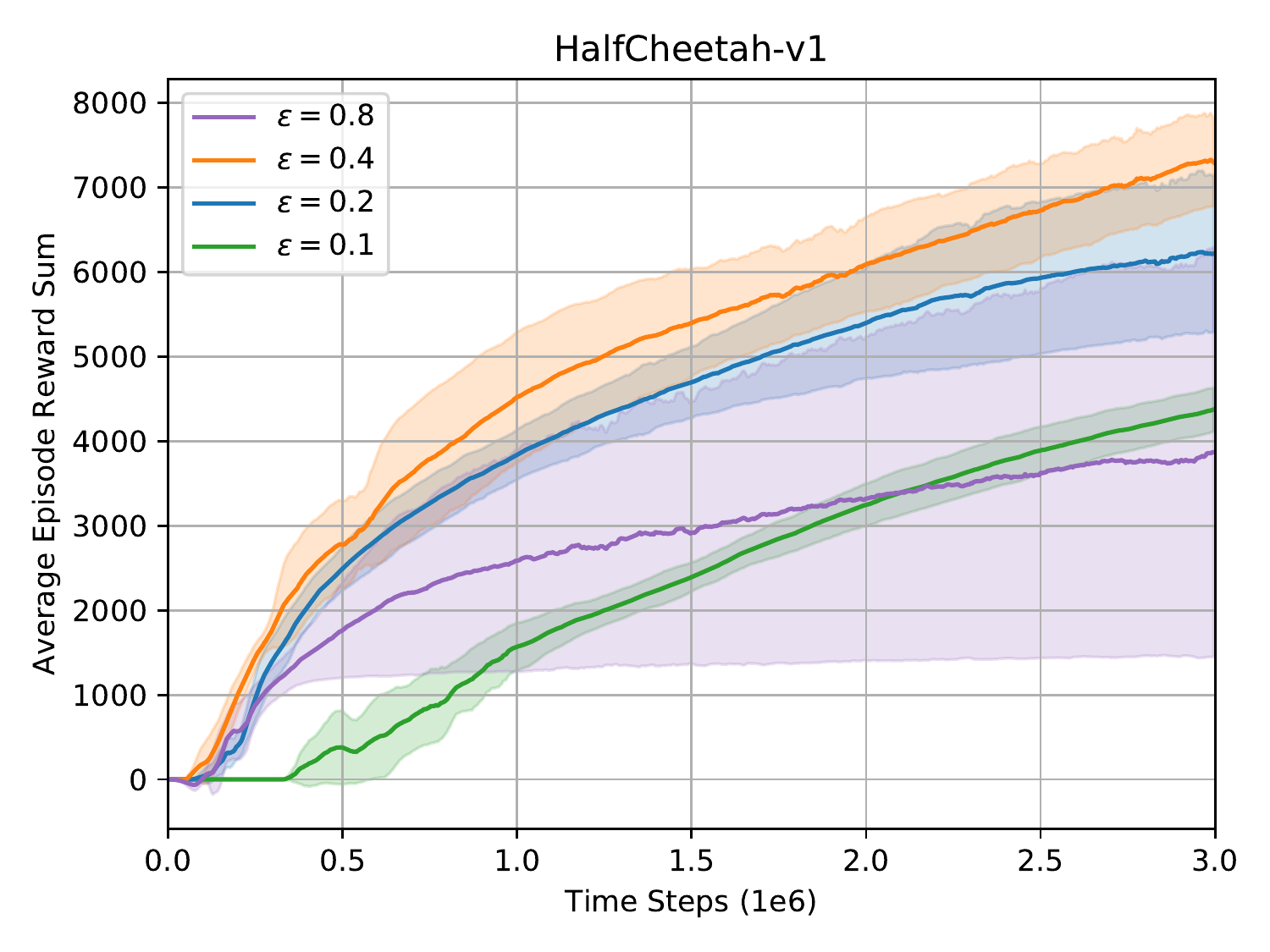}
		\includegraphics[width=0.24\textwidth]{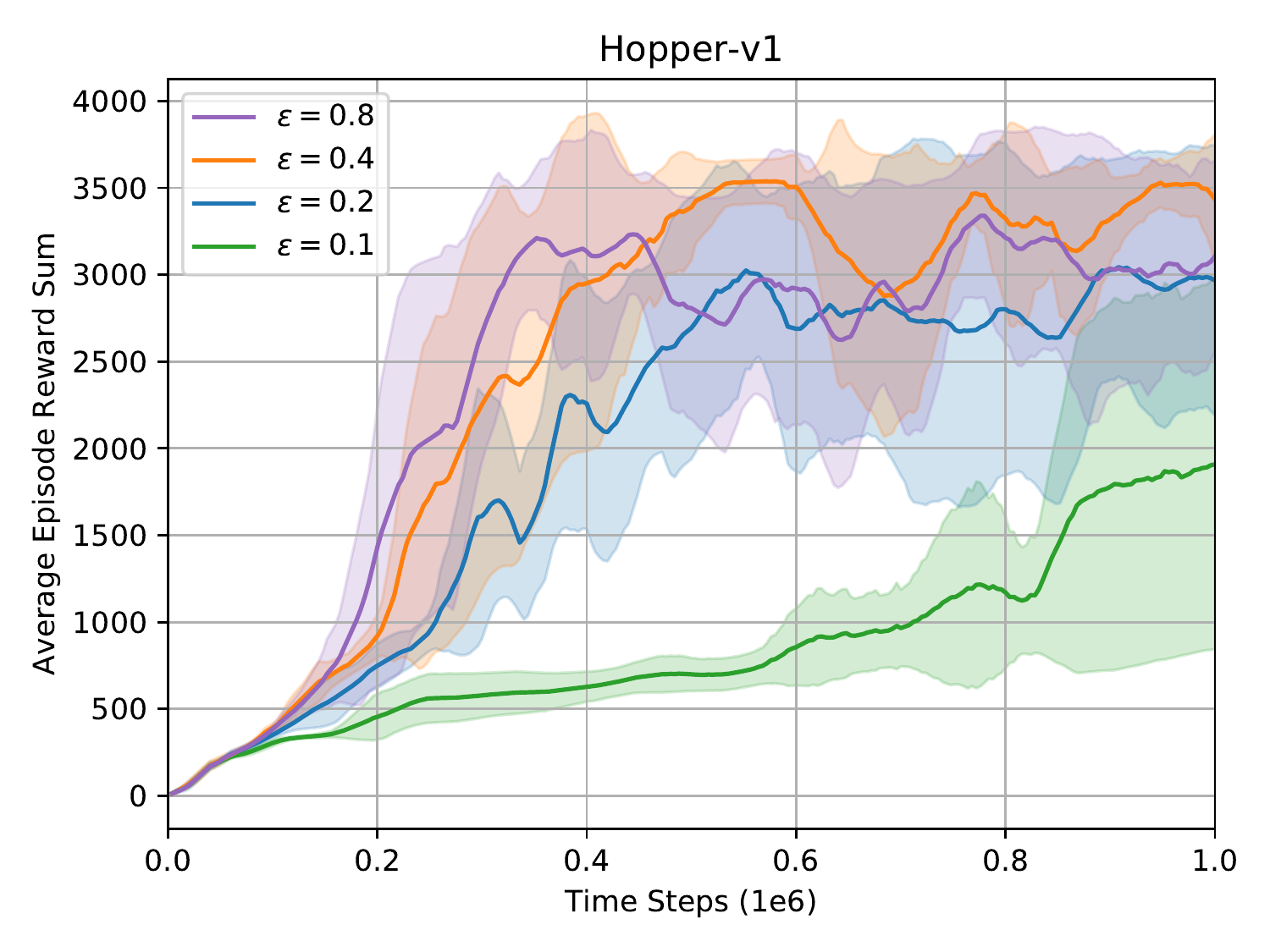}
		\includegraphics[width=0.24\textwidth]{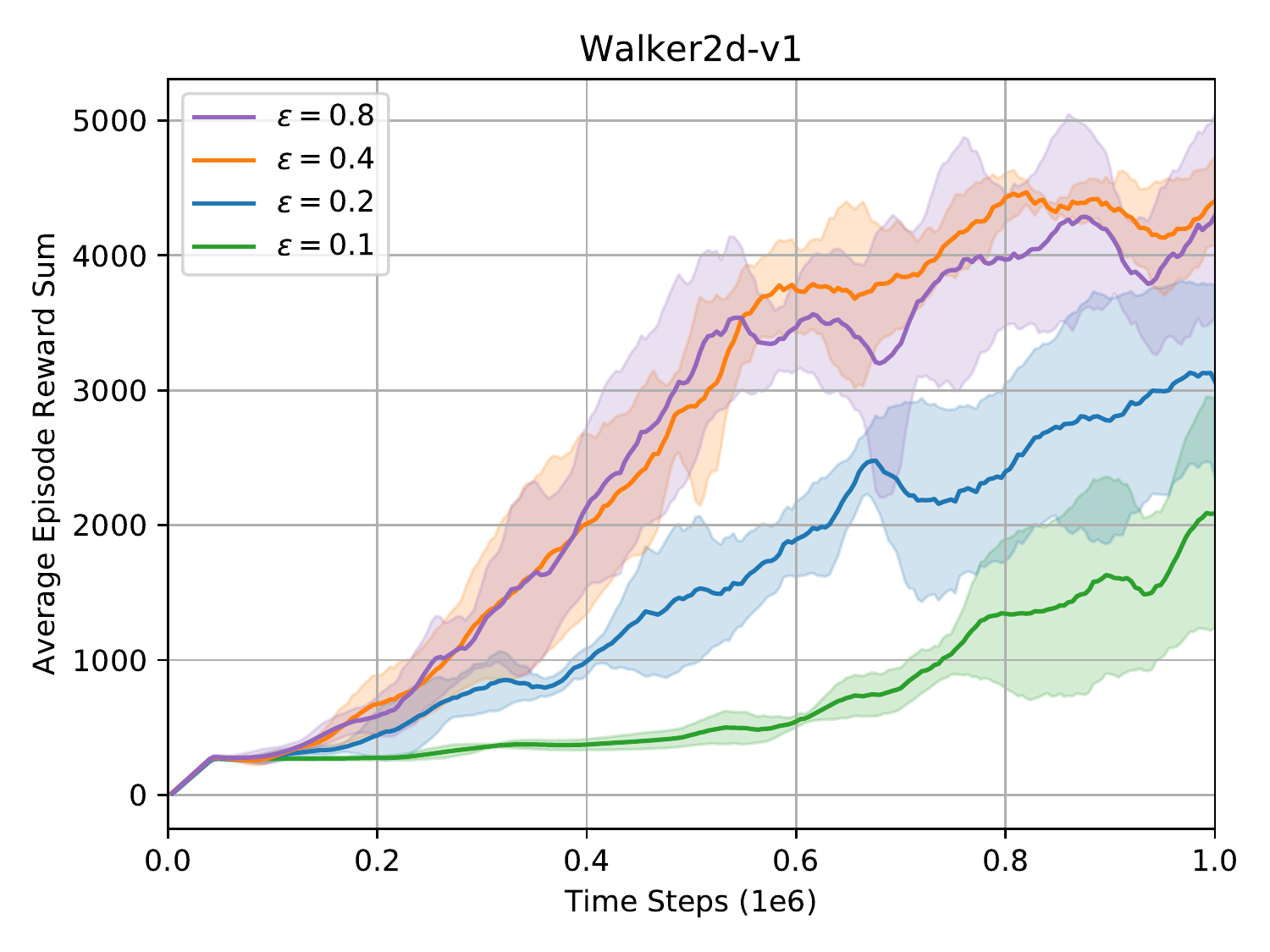}
		\caption{Impact of the clipping factor $\epsilon$}
		\label{fig:ablclip}
		\centering
		\includegraphics[width=0.24\textwidth]{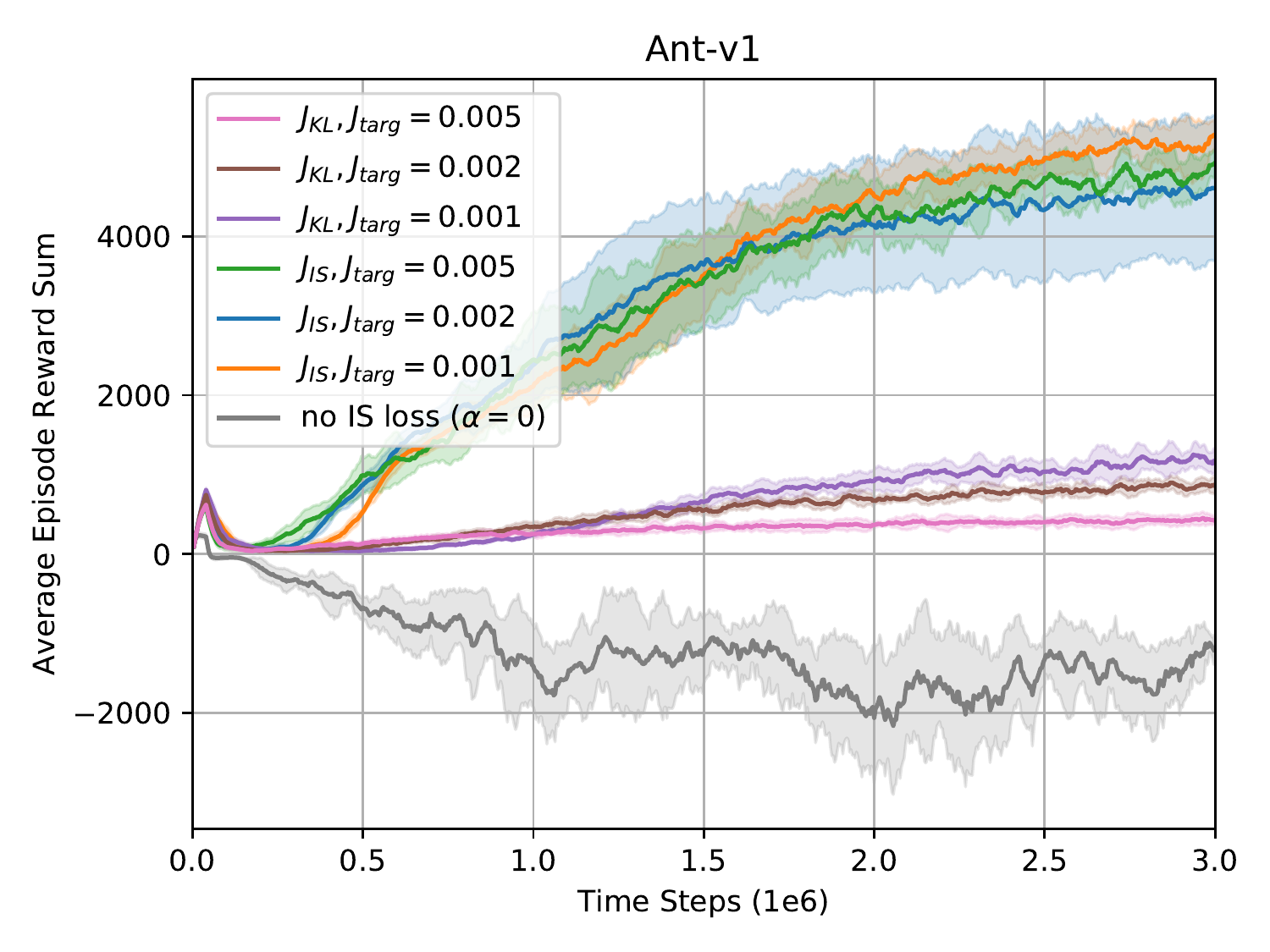}
		\includegraphics[width=0.24\textwidth]{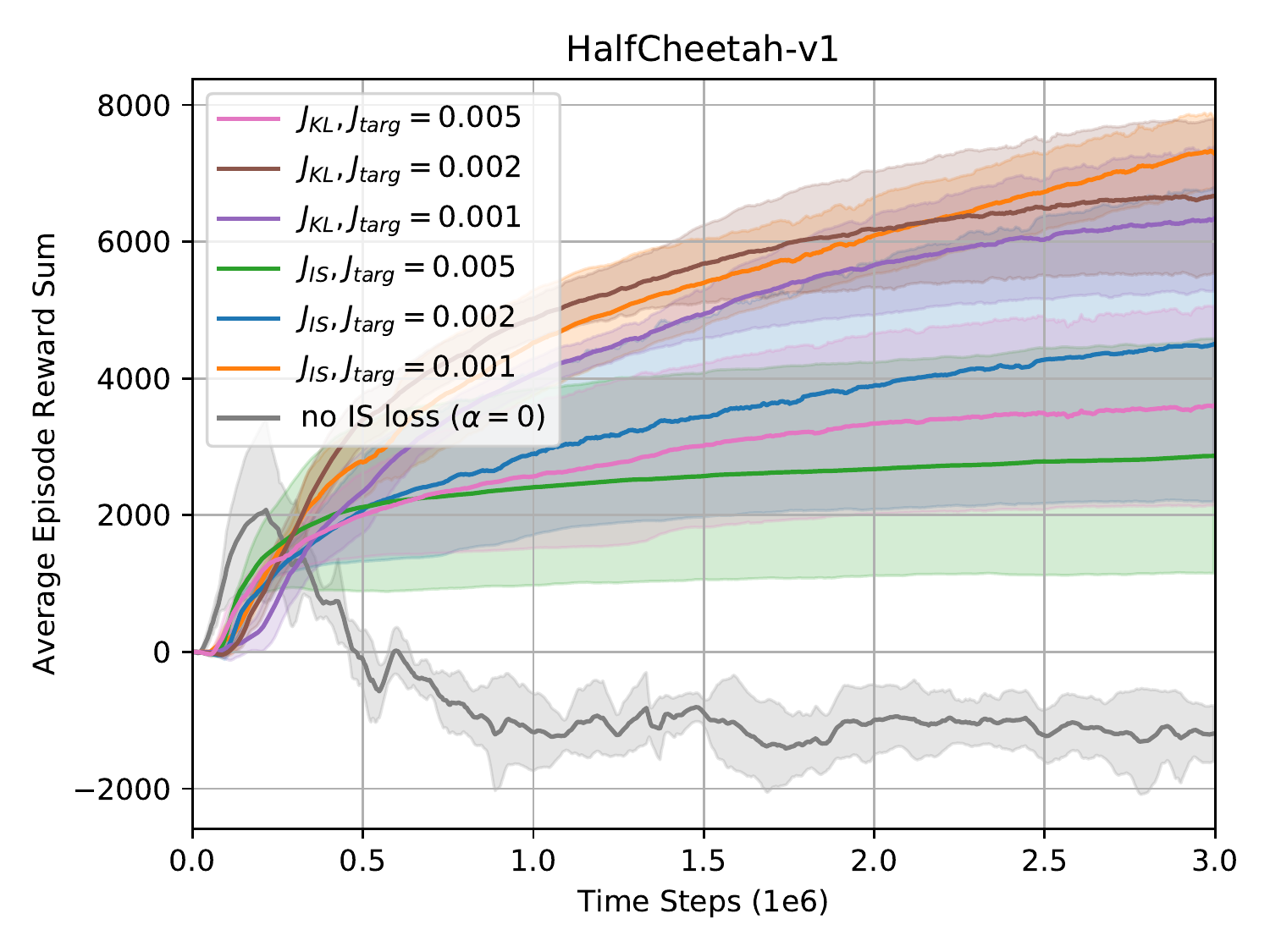}
		\includegraphics[width=0.24\textwidth]{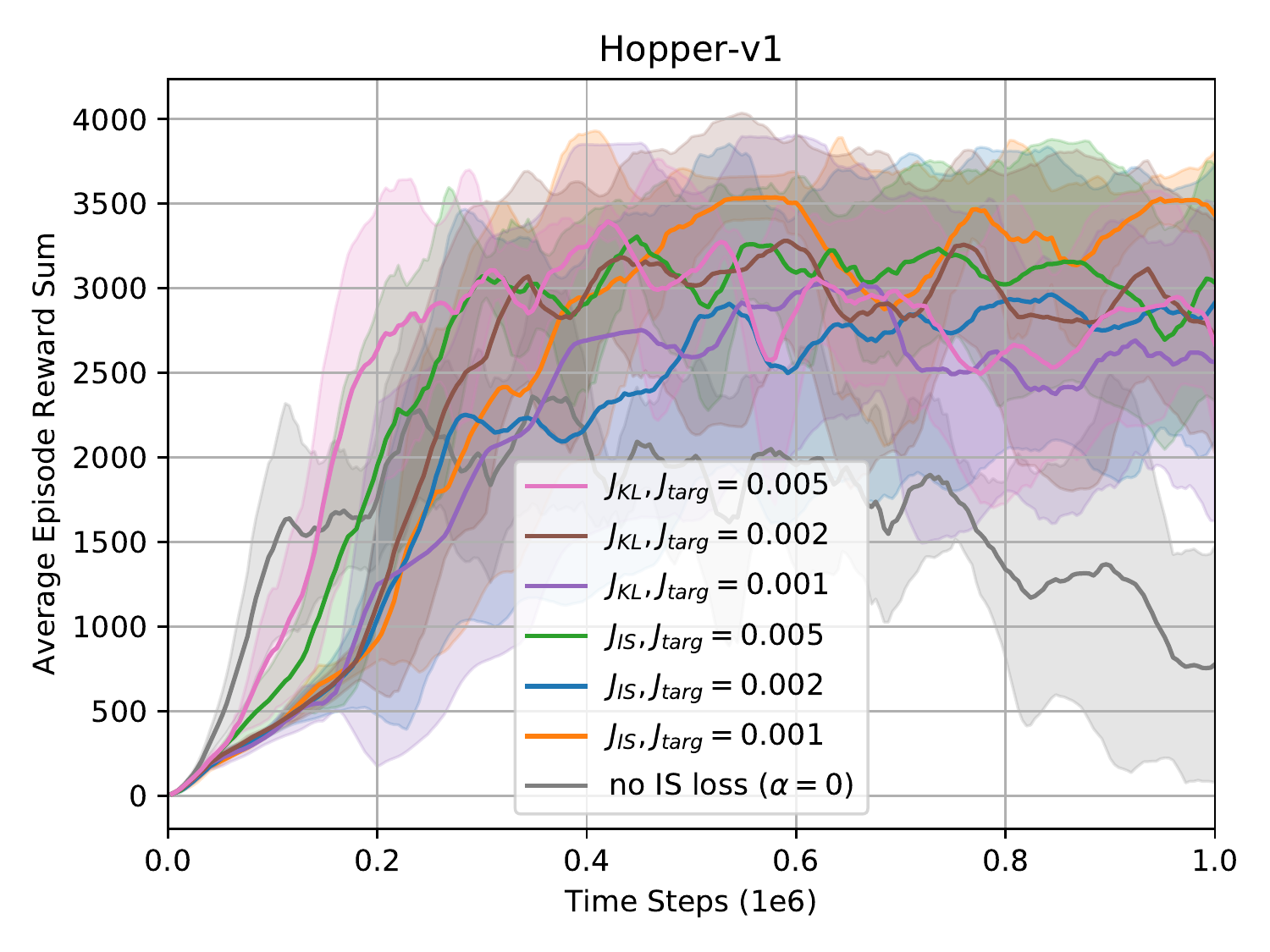}
		\includegraphics[width=0.24\textwidth]{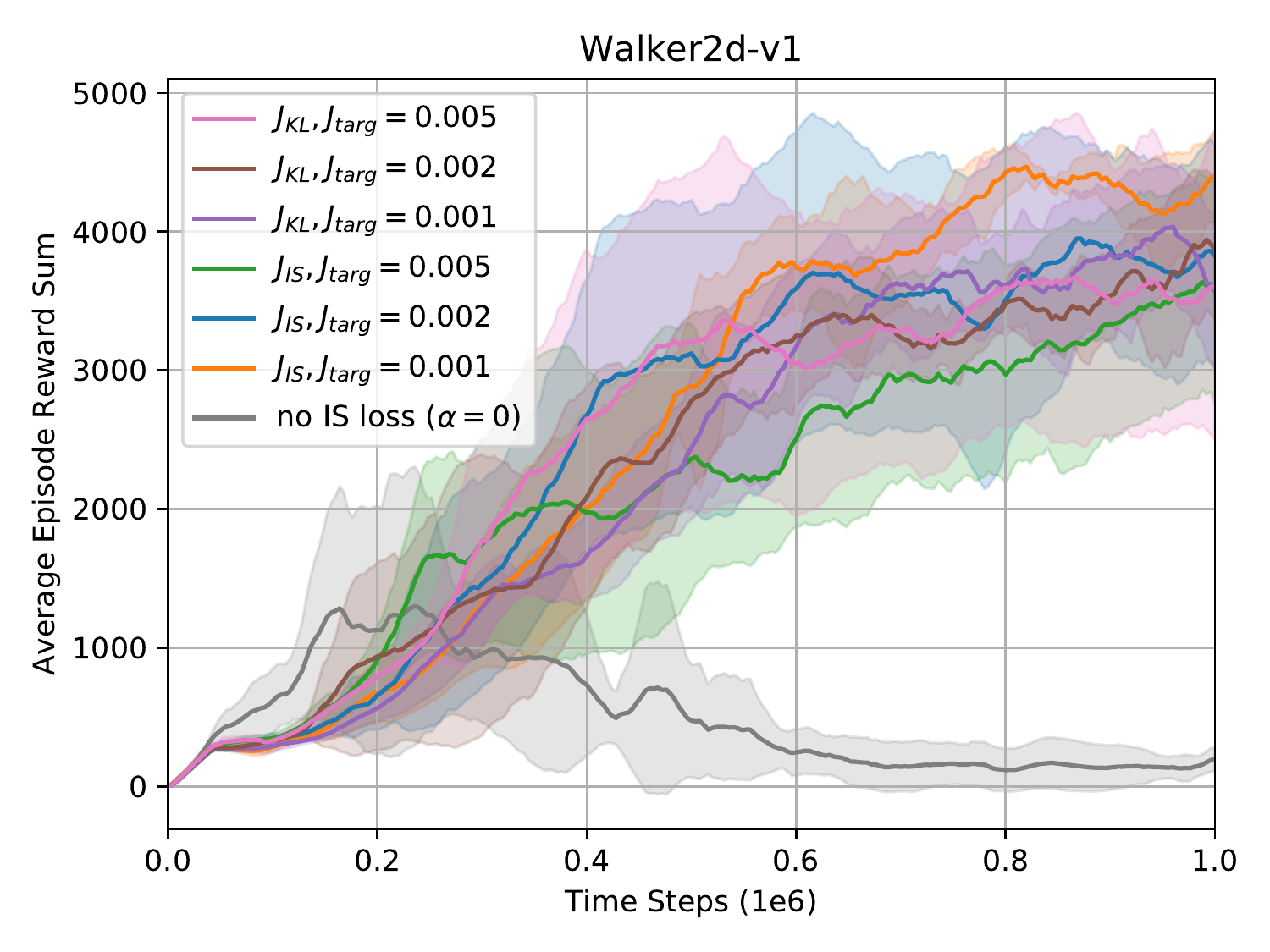}
		\caption{Impact of the IS loss term $J_{IS}$ target factor $J_{targ}$}
		\label{fig:ablIStarg}
		\centering
		\includegraphics[width=0.24\textwidth]{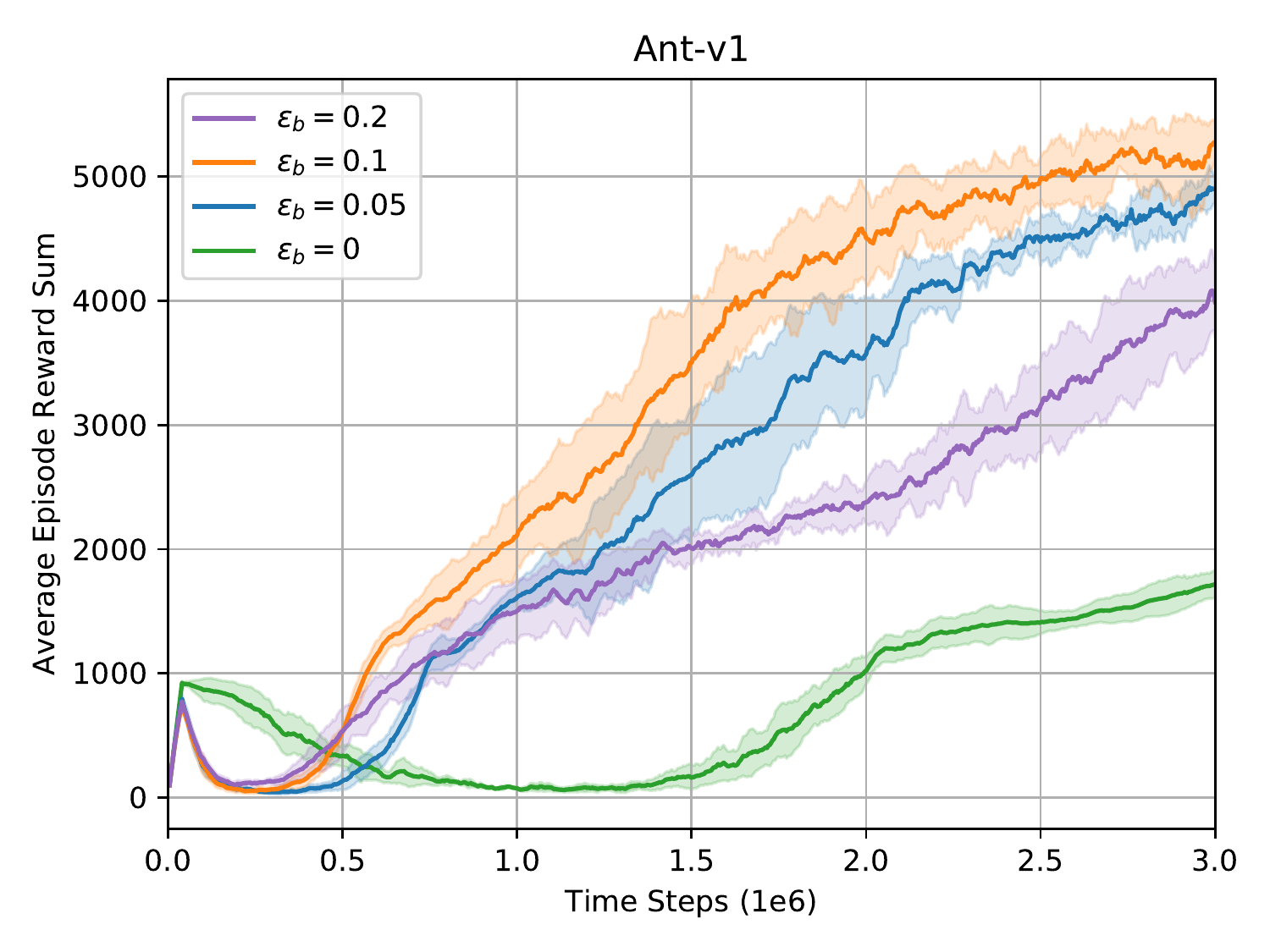}
		\includegraphics[width=0.24\textwidth]{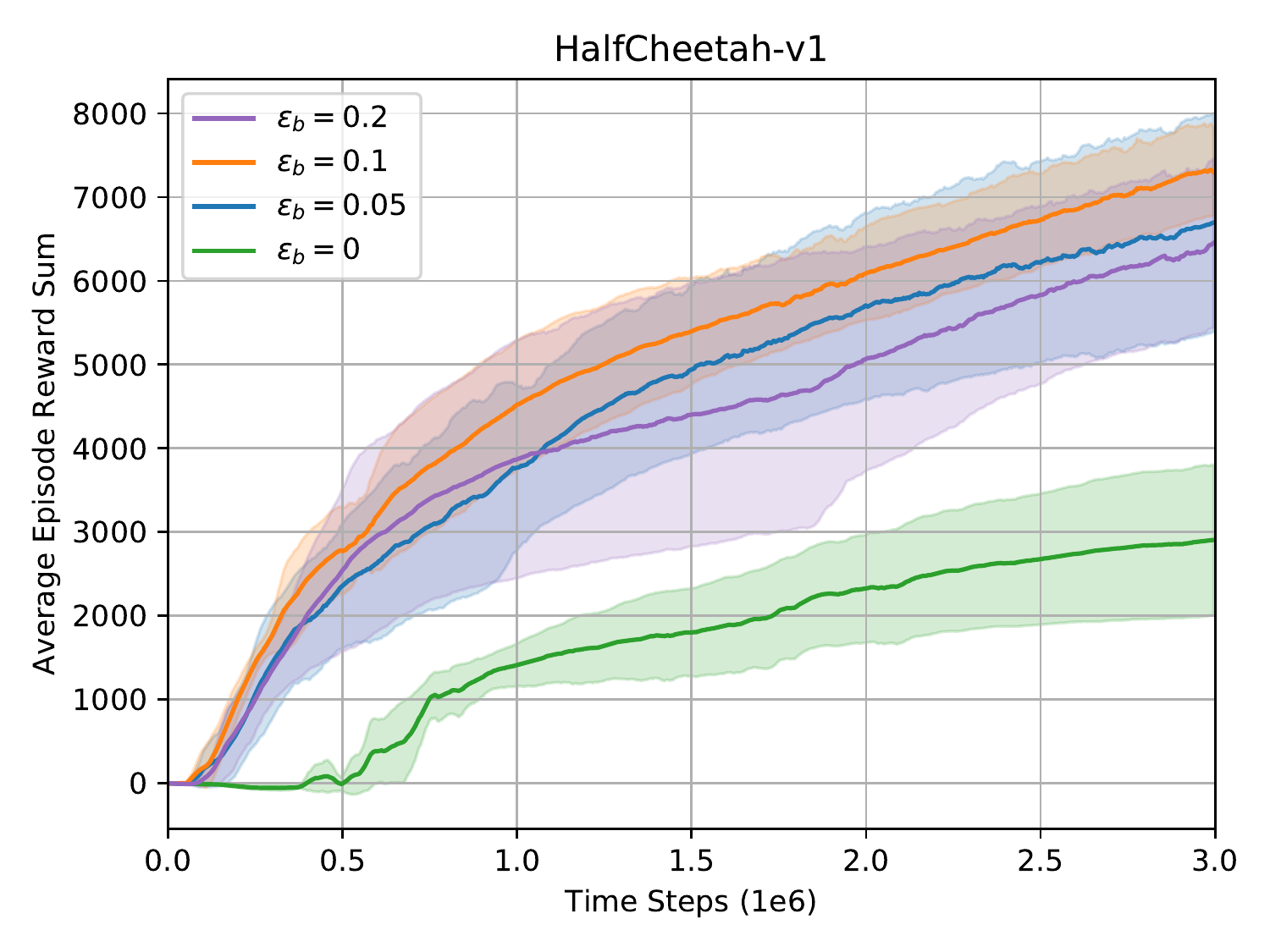}
		\includegraphics[width=0.24\textwidth]{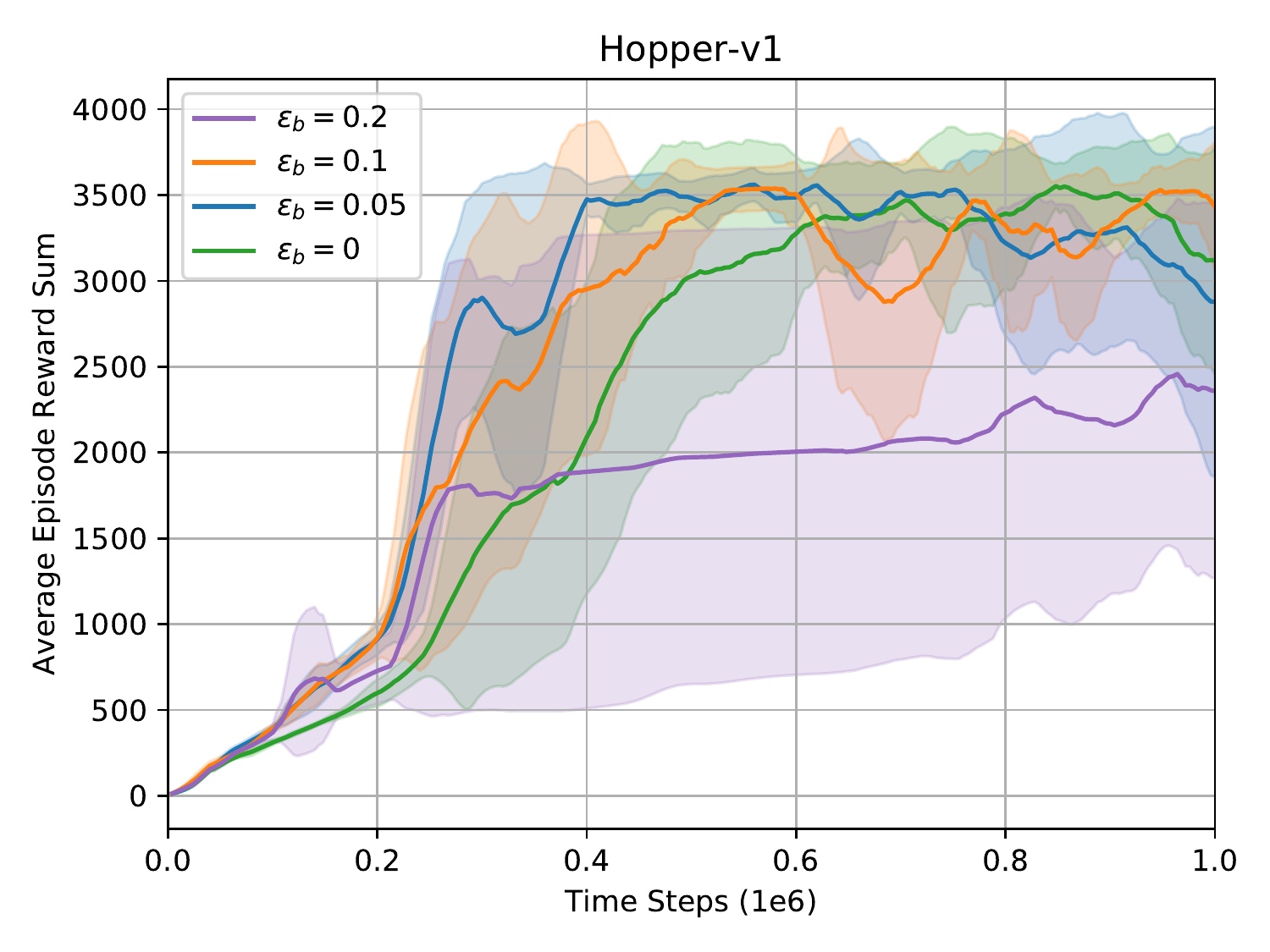}
		\includegraphics[width=0.24\textwidth]{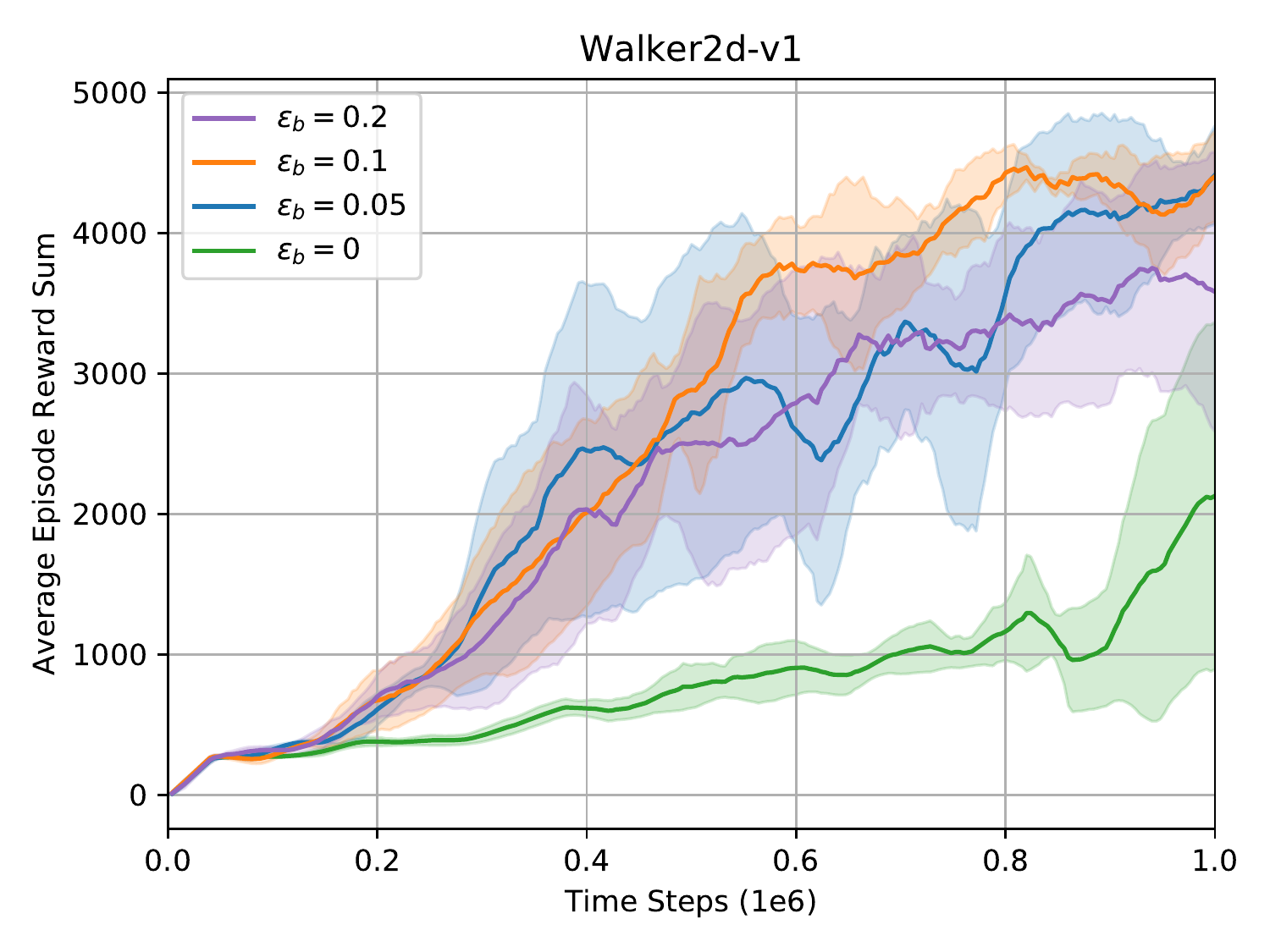}
		\caption{Impact of the batch inclusion parameter $\epsilon_b$}
		\label{fig:ablbatchlimit}
	\end{figure}

	\newpage
	\section{Learning Curves of DISC and Other State-of-the-Art RL Algorithms}
	\begin{figure}[!h]
		\centering
		\includegraphics[width=0.33\textwidth]{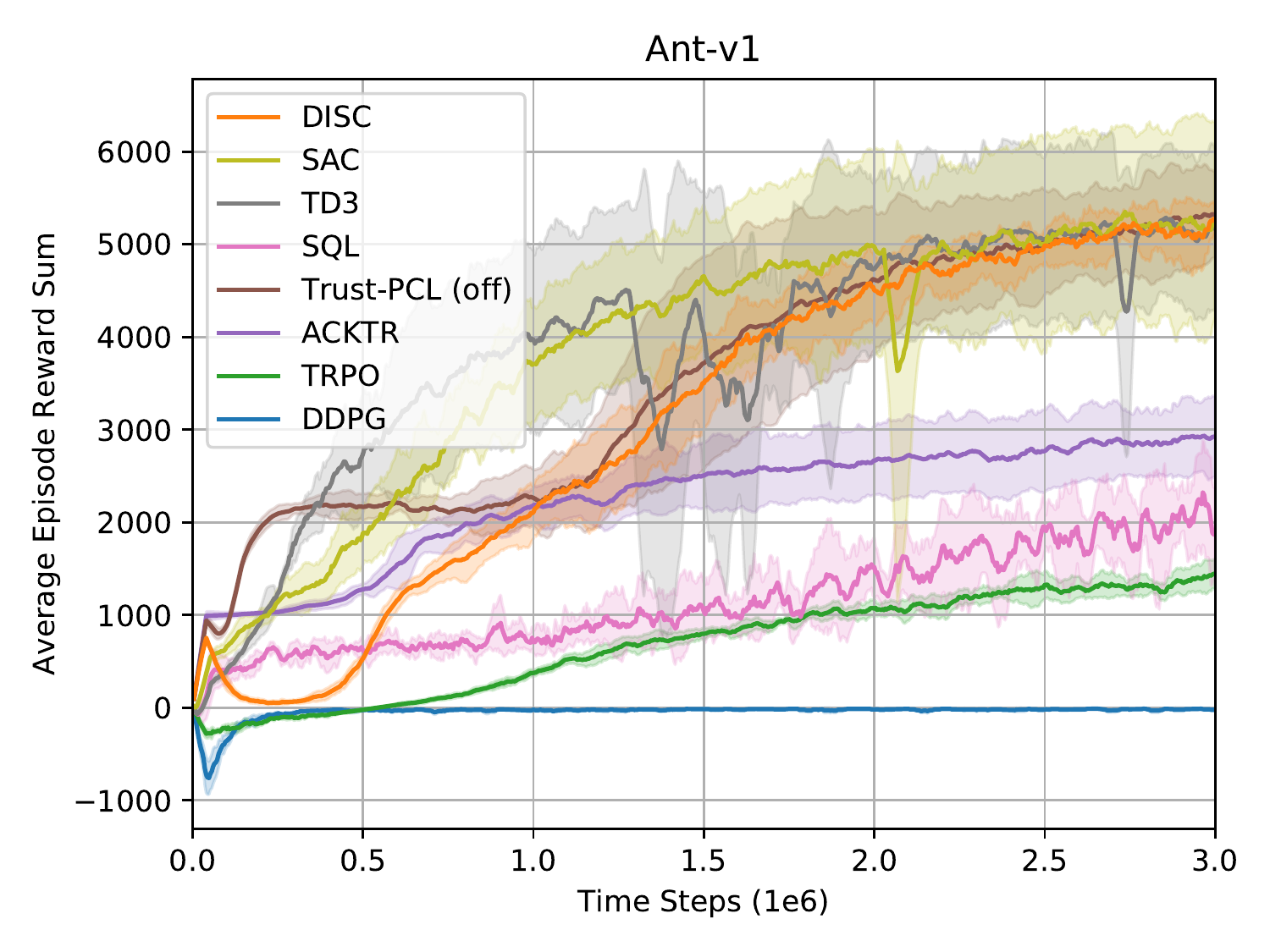}
		\includegraphics[width=0.33\textwidth]{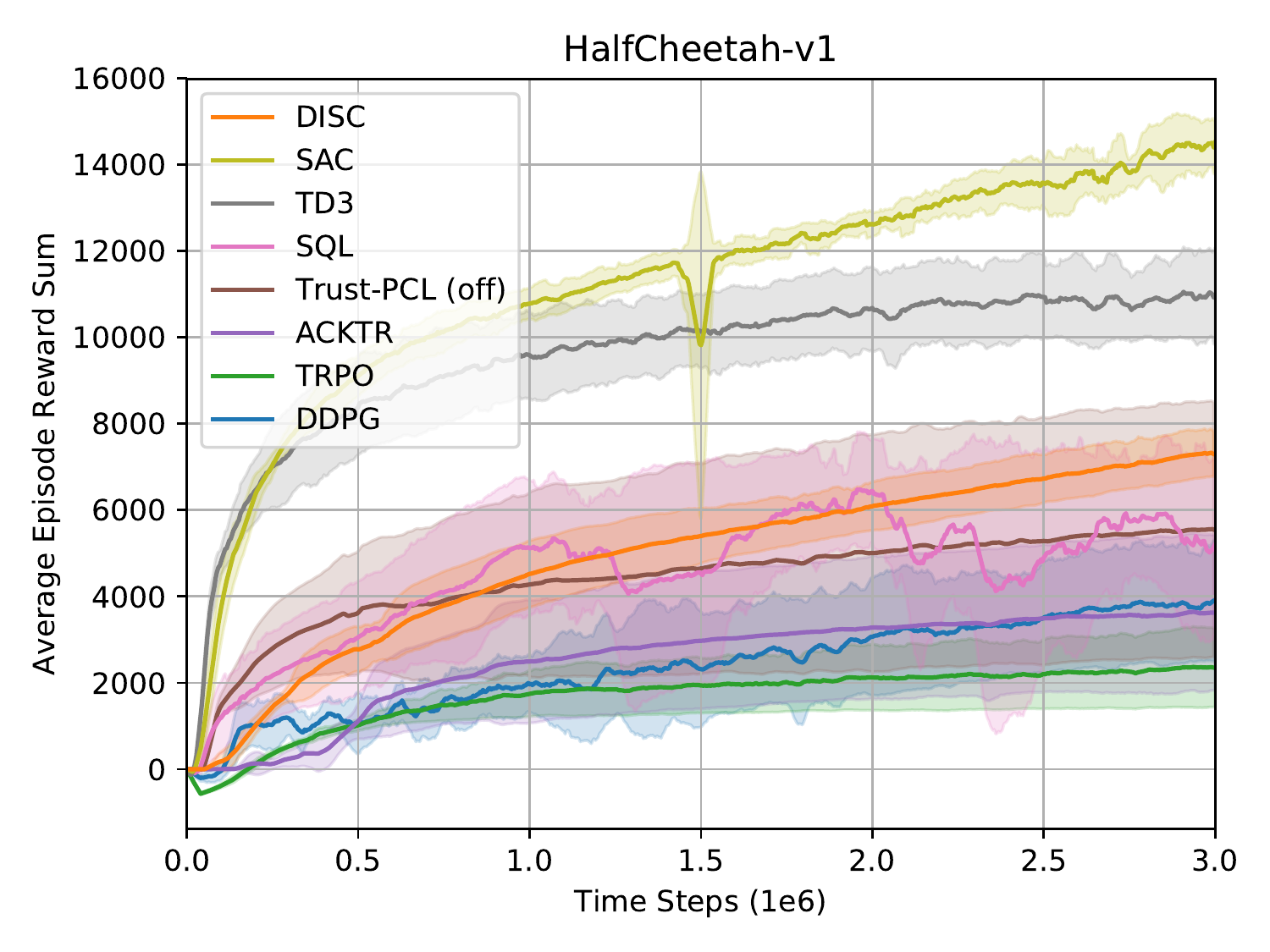}
		\includegraphics[width=0.33\textwidth]{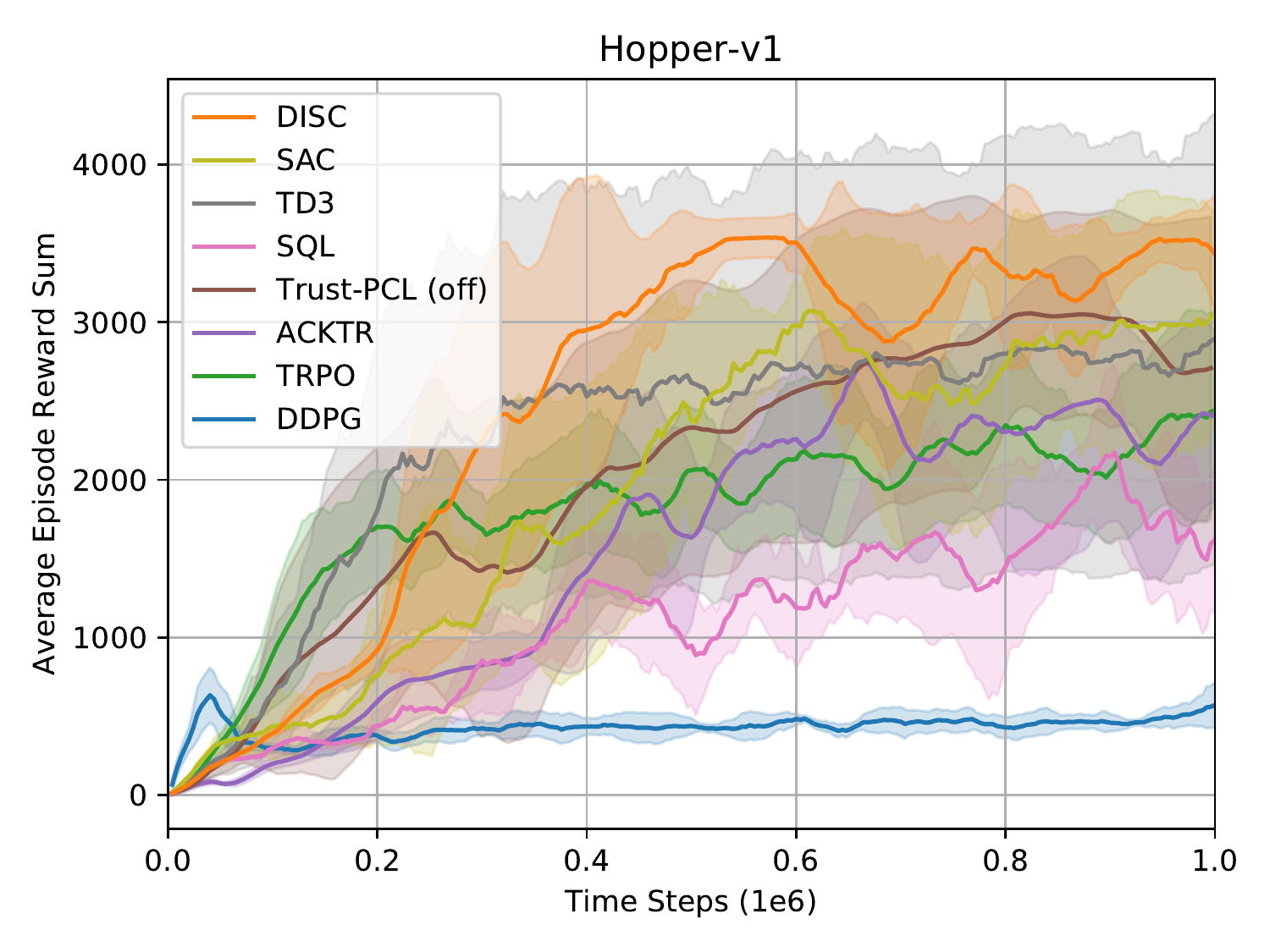}
		\includegraphics[width=0.33\textwidth]{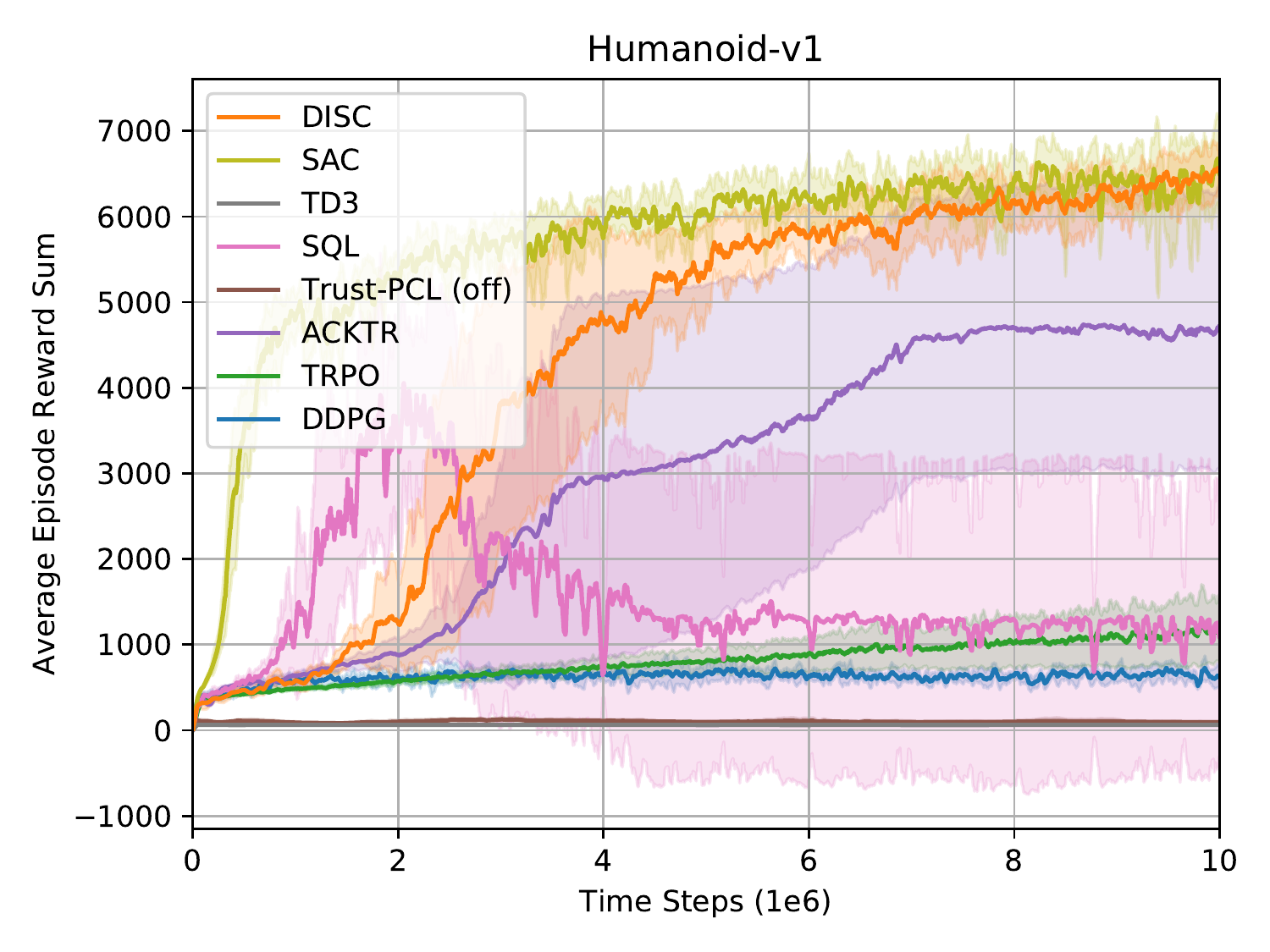}
		\includegraphics[width=0.33\textwidth]{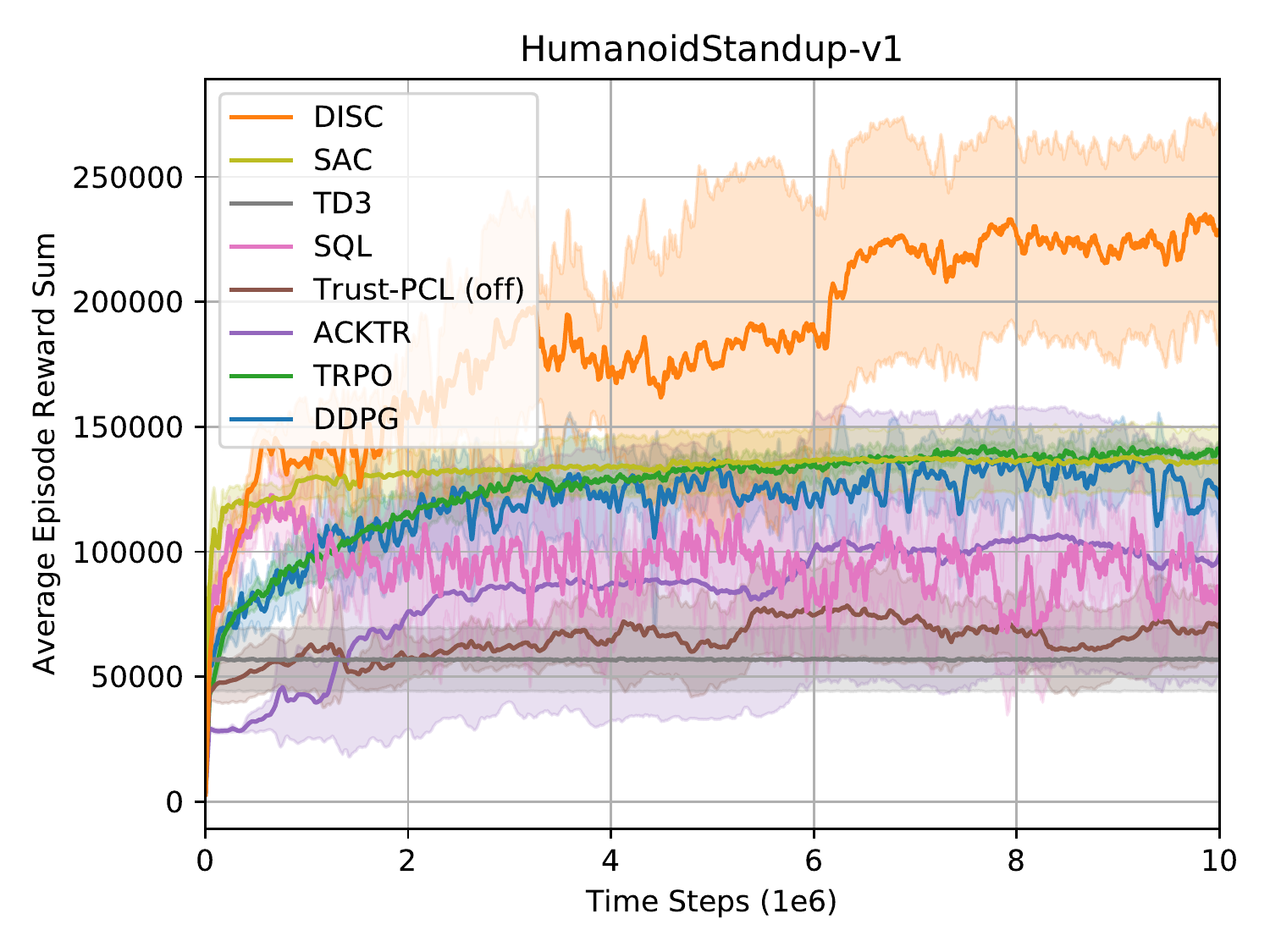}
		\includegraphics[width=0.33\textwidth]{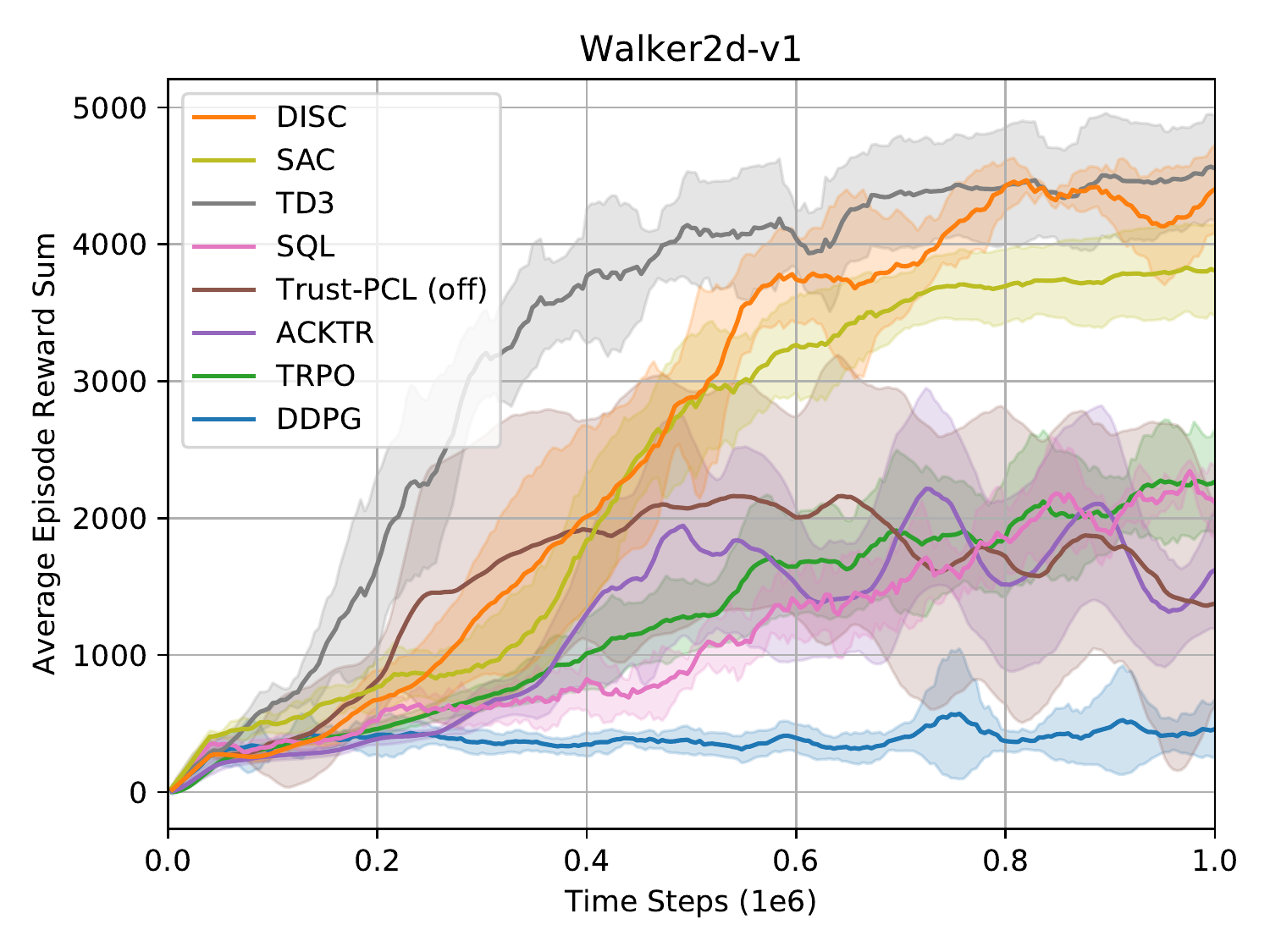}
		\caption{Learning Curves of DISC and Other State-of-the-Art RL Algorithms on Mujoco tasks}
		\label{fig:perfcomp}
	\end{figure}
\end{document}
